\documentclass{article}

\usepackage{arxiv}
\AtBeginDocument{\newgeometry{textwidth=6.5in,textheight=9.6in,top=1in,headheight=14pt,headsep=25pt,footskip=30pt}}

\usepackage[utf8]{inputenc}
\usepackage[T1]{fontenc}
\usepackage{tikz}
\usetikzlibrary{arrows.meta, positioning}
\usepackage{amsmath,amssymb,amsfonts,amsthm}
\usepackage{graphicx}
\usepackage{textcomp}
\usepackage{xcolor}
\usepackage{booktabs}
\usepackage{multirow}
\usepackage{tabularx}
\usepackage[most]{tcolorbox}
\usepackage{makecell}
\usepackage{placeins}
\usepackage{adjustbox}
\usepackage{float}
\usepackage{pdflscape}
\usepackage{cite}
\usepackage{url}
\usepackage[hidelinks]{hyperref}
\usepackage{authblk}

\newtheorem{proposition}{Proposition}

\newcommand{\examplebox}[2]{%
  \begin{tcolorbox}[colframe=black!70!white,colback=blue!2!white,arc=4pt,
    boxrule=0.5pt,fontupper=\small,top=2pt,left=2pt,right=2pt,bottom=2pt,
    title=#1]
  #2
  \end{tcolorbox}
}

\newif\ifanonymous
\anonymousfalse

\renewcommand{\headeright}{Preprint}
\renewcommand{\undertitle}{Preprint}
\renewcommand{\shorttitle}{Small Open-Weight Language Models for Crowd Decision-Making}

\title{Small Open-Weight Language Models for Crowd Decision-Making from Textual Opinions: Reliability of Linguistic Aggregation and Output-Level Fusion}

\author[1,2]{David Herrera-Poyatos\thanks{Corresponding author: \texttt{divadhp@ugr.es}}}
\author[1,2]{Cristina Zuheros}
\author[1,2]{Carlos Pel\'aez-Gonz\'alez}
\author[1,3]{Andr\'es Herrera-Poyatos}
\author[1,4]{Rosana Montes}
\author[1,2,5]{Francisco Herrera}
\affil[1]{Andalusian Research Institute in Data Science and Computational Intelligence (DaSCI), University of Granada, Spain}
\affil[2]{Department of Computer Science and Artificial Intelligence, University of Granada, 18071 Granada, Spain}
\affil[3]{Department of Algebra, University of Granada, Spain}
\affil[4]{Department of Software Engineering, University of Granada, Spain}
\affil[5]{ADIA Lab, Abu Dhabi, United Arab Emirates}

\hypersetup{pdftitle={Small Open-Weight Language Models for Crowd Decision-Making from Textual Opinions: Reliability of Linguistic Aggregation and Output-Level Fusion}, pdfauthor={David Herrera-Poyatos, Cristina Zuheros, Carlos Pel\'aez-Gonz\'alez, Andr\'es Herrera-Poyatos, Rosana Montes, Francisco Herrera}, pdfkeywords={2-tuple linguistic representation, crowd decision-making, ranking uncertainty, small open-weight language models}}

\date{\small This work has been submitted to the IEEE for possible publication. Copyright may be transferred without notice, after which this version may no longer be accessible.}

\begin{document}

\maketitle

\begin{abstract}
Small open-weight language models, quantized to run on one consumer graphics card, can support crowd decision-making on local infrastructure. We evaluate whether their collective comparisons retain resolution, remain stable and distinguish alternatives, through a pipeline of repeated judgments, linguistic aggregation and fixed-panel output-level fusion. Five local models and a hosted reference assess 474 reviews of four London restaurants; three local models assess 600 reviews of twelve Madrid restaurants. Continuous arithmetic aggregation, expressed through the 2-tuple linguistic representation, leaves about half the residual standard deviation of rounded operators, whereas rounding at every stage leaves at most two distinct collective values. Redrawing the participants changes the winner far more often than redrawing the model runs: with ten runs per review, the inference variance of the winner's lead is at most 4.3\% of its participant variance, and an illustrative 10\% threshold is met with one to five runs. Simultaneous pairwise intervals that quantify ranking uncertainty identify no unique winner, and neither do the reviewers' own star ratings; in London, every observed lead is smaller than the differences these intervals can resolve. Linear fusion averages its members' collective evaluations, so it cannot separate alternatives more than its best member; it yields highly repeatable winners without depending on one model, but its additional calls bring limited gains in winner persistence. Local crowd decision-making with these models is therefore feasible if it keeps assessments continuous, sizes inference to the participant uncertainty, and reports margins and candidate sets alongside point rankings.
\end{abstract}


\section{Introduction}
\label{sec:introduction}

Crowd decision-making (CDM) uses individual opinions to support collective comparisons among alternatives \cite{zuheros2022crowd}. When those opinions arrive as unrestricted text, a decision system must connect language interpretation with the numerical or linguistic assessments from which collective evaluations are constructed. Sentiment-based methodologies have established this connection \cite{zuheros2021sentiment}, and large language models (LLMs) have expanded the possibilities for processing opinions through natural-language instructions \cite{zuheros2024large}. The resulting engineering problem extends across the complete pipeline: the system must extract usable assessments, combine them without unnecessary information loss, and determine which comparisons the available opinions support.

Small open-weight language models make such pipelines deployable on local infrastructure, where institutions retain the opinions and control the weights, inference settings and processing rules, so that collective evaluations can be reproduced and inspected without a hosted service. Their practical value depends on the decision support obtained within the available hardware and inference budget, which requires connecting their response behavior with the collective outcomes produced downstream.

Repeated inference illustrates why this connection matters. Identical requests can produce different judgments, motivating repeated querying and aggregation \cite{herrera2025overview,reiss2023testing}, and combining several models further reduces dependence on one response \cite{jiang2023llmblender}. Yet a less variable assessment does not by itself establish an informative comparison: rounding may suppress distinctions between opinions, a highly repeatable winner may have only a small lead, and a ranking stable across model executions may change with a different sample of participants. The engineering question is which distinctions survive the pipeline, which perturbations change them, and what additional inference contributes to their support.

This paper develops and applies an integrated evaluation protocol for locally deployed small open-weight CDM systems. The protocol connects the stages at which judgments are aggregated with review-level resolution, collective margins, ranking stability and simultaneous candidate sets. It evaluates inference, participant and joint perturbations, and compares repeated single-model inference, fixed-panel output-level fusion and greedy decoding under explicit call and runtime budgets. Its components are standard; the contribution lies in their joint use for a configured decision pipeline and in the design rules they yield: keep the mean continuous until the final report, size the number of runs to the participant uncertainty, judge a fixed panel by the repeatability it adds rather than by separation, and report candidate sets with the point ranking.

We instantiate the protocol on two restaurant-review decision problems. Five local open-weight models and a hosted reference assess 474 English reviews of four London restaurants; three local models assess 600 reviews of twelve Madrid restaurants. Here, \emph{small} denotes the studied deployment envelope: quantized open-weight models of 7.5B to 36B total parameters running on one consumer GPU. The contributions address three connected questions:

\begin{itemize}
    \item \textbf{Capabilities at the decision level.} We characterize individual configurations through output validity, agreement with an external rating reference, margins, ranking stability and candidate sets, using matched participant-resampling streams to separate inference repeatability from crowd sensitivity.

    \item \textbf{Information preservation across aggregation stages.} We compare continuous averaging, which the 2-tuple representation expresses linguistically without changing its value, with rounded mean, rounded median and majority operators applied at specified stages, separating rounding from changes of estimator.

    \item \textbf{The decision-level value of additional inference.} We compare fixed-panel fusion with repeated individual inference and greedy decoding at equal calls and estimated runtime, and show that linear fusion averages its members' collective evaluations, so that it can add repeatability and independence from any single model but not separation.
\end{itemize}

The appendices, whose sections, tables and figures carry the prefix S, gives the prompts, decoding settings, model provenance, timings and complete results. Sections~\ref{sec:related} to~\ref{sec:conclusion} cover related work, the research questions, the protocol, the findings, their implications and the conclusions.

\section{Related Work}
\label{sec:related}

\subsection{Crowd Decision-Making from Textual Opinions}

Zuheros et al.~\cite{zuheros2021sentiment} developed a sentiment-based multi-person multi-criteria methodology for evaluating alternatives from unrestricted reviews. Subsequent CDM work incorporated sparse representation to exploit crowd opinions \cite{zuheros2022crowd}, while explainable CDM connected collective outcomes with expert opinions and subgroup descriptions \cite{zuheros2023explainable}. The closest prior study is that of Herrera-Poyatos et al.~\cite{zuheros2024large}, which elicited five-term linguistic assessments of the same TripR-2020Large reviews from a hosted LLM, aggregated them with the 2-tuple weighted average among other prompt-design scenarios, and repeated queries in a sensitivity analysis. Its 2-tuple scenario already placed the four restaurants within the term \emph{Positive}, in the order that most systems reproduce here (Section~\ref{sec:results-rq1}). The present evaluation differs in four respects: it uses small open-weight models served locally, repeats every query and treats the runs as a sample for the decision, fuses several models at the output level, and evaluates the resulting decision under run and participant resampling instead of reporting a single ranking. That study also reports the rankings of the four restaurants obtained by sentiment-based CDM methods \cite{zuheros2022crowd,zuheros2023explainable}, which serve below as an external reference that needs no reimplementation.

The reassessment of sentiment analysis with LLMs \cite{zhang2024reality} shows why the language-processing component needs empirical scrutiny, but review-level performance leaves open how errors, variability and aggregation choices propagate into comparisons between alternatives, which is the focus here. With 132 and 599 participants, both problems are also large-scale group decisions \cite{ding2020largescale}, a setting in which consensus models examine how far opinions converge before a choice \cite{palomares2014consensus} and LLMs have recently been used to group participants \cite{gonzalezquesada2025modelling}. The participant sensitivity examined here is the decision-level face of limited consensus relative to the differences between alternatives, although no consensus measure is computed.

\subsection{Small Open-Weight Models and Inference Variability}

Openly available language models differ in architecture, parameter scale and release conditions \cite{kukreja2024literature}; access to weights enables local deployment of a selected checkpoint, so reproducibility depends on documenting the complete deployed configuration. Herrera-Poyatos et al. \cite{herrera2025overview} review uncertainty and variability in LLM-based sentiment analysis, including sources associated with inference and prompt design. Reiss~\cite{reiss2023testing} finds that pooling repeated outputs can improve the reliability of text annotation while still requiring independent validation, self-consistency \cite{wang2023selfconsistency} aggregates sampled reasoning paths for answer selection, and statistical treatments of evaluation variance separate the sampling of items from that of model responses \cite{miller2024adding}. Their evaluation objects, however, differ from a collective ranking derived from many participants' opinions.

A related study \cite{herrerapoyatos2026ordinalization} examines how elicitation formats shape the variability of individual sentiment judgments and builds collective rankings from individual runs. Two of its London collections, Gemma 4 E4B and GPT-5.2, are reused here; the analyses of aggregation stages, participant resampling, simultaneous separation and inference budgets are new here.

\subsection{Linguistic Aggregation and Output-Level Fusion}

The 2-tuple linguistic model \cite{herrera2000} pairs a linguistic term with a symbolic translation, allowing a numerical aggregate to be expressed without discarding its displacement from the nearest term \cite{2tuplasBook}. In the translation, manipulation and retranslation scheme of computing with words \cite{yager2004retranslation}, the symbolic model with approximation rounds to the nearest index after every aggregation step \cite{delgado1993aggregation}, and its information loss motivated the 2-tuple model \cite{herrera2000}. The rounded operators evaluated here apply the symbolic model with approximation at chosen stages of a crowd pipeline. The question is therefore not whether the 2-tuple model avoids this loss, which is established, but what the loss does to a crowd decision built from repeated LLM judgments, and at which stage retranslation can occur without changing that decision. Other representations keep more of a set of assessments: linguistic distribution assessments retain proportions over terms \cite{zhang2014distribution}, hesitant fuzzy linguistic term sets retain the terms between which an assessor hesitates \cite{rodriguez2012hesitant}, and personalized individual semantics give each assessor its own numerical meaning of the terms \cite{li2017personalized}, which could absorb model-specific biases.

Output-level fusion combines model responses at inference time without changing parameters, unlike model merging \cite{yang2408model}. LLM-Blender \cite{jiang2023llmblender} uses learned pairwise ranking and generative fusion; sampling-and-voting shows that more agents can improve benchmark performance \cite{li2024moreagents}, while compound inference systems need not improve monotonically with the number of calls \cite{chen2024morecalls}. The diversity of ensemble members, often invoked to justify combining them, shows only a weak and unreliable relation to ensemble accuracy \cite{kuncheva2003measures}, and noisy annotators without ground truth are classically combined by estimating each annotator's error rates \cite{dawid1979maximum}; the fixed panels studied here weight their members equally.

These reference-based evaluations do not establish how fusion affects a collective comparison whose human opinion sample is itself uncertain. Ranking uncertainty has its own methodological foundations. Subset selection and multiple comparisons with the best identify which alternatives may be best \cite{gupta1965some,hsu1996multiple}, stochastic multicriteria acceptability analysis examines rank distributions under uncertain inputs \cite{lahdelma2001smaa2}, and inference for ranks constructs confidence sets for ranks and for the identity of the best alternative \cite{mogstad2024inference}. The bootstrap of empirical ranks, in contrast, can fail when alternatives are tied or nearly tied \cite{hall2009using}, as they are in the problems studied here.

\section{System Formulation and Research Questions}
\label{sec:problem}

Let $X=\{x_1,\ldots,x_m\}$ be a set of alternatives and $\mathcal{U}$ a crowd of participants who provide textual opinions about them. Each alternative $x_j$ receives a set of reviews $\mathcal{R}_j$, and a participant may contribute more than one review. A language model maps each review to an assessment under a specified elicitation and parsing protocol; the assessments from repeated queries and, with several models, from the panel members are combined through an aggregation hierarchy into a collective evaluation $v_j$ of each alternative, which induces a ranking.

The decision object is the identity of the best alternative, reported as a point winner together with a candidate set of the alternatives that the uncertainty analysis does not exclude from being best; the complete ranking is a secondary output. The estimand of $v_j$ is the expected assessment, under the configured pipeline and its stochastic inference, of a review of $x_j$ by a participant drawn from a population resembling the sampled one. In London that population is represented by a selected panel whose participants contribute several reviews each on average; in Madrid nearly every participant contributes a single review, so participant resampling amounts to review resampling.

The evaluation distinguishes four properties: \emph{resolution}, the numerical distinctions retained after aggregation; \emph{stability}, the persistence of ranking features, including the winner, under a specified perturbation; \emph{sensitivity}, the changes associated with repeated inference, participant composition and modeling choices; and \emph{separation}, whether differences between alternatives can be distinguished under the stated uncertainty procedure. They must be read jointly: a stable ranking may contain very small margins, and retained resolution does not establish that the opinions support a unique winner. The reviewers' own star ratings provide an external anchor for the evaluations extracted from their texts; they are not a ground-truth collective ranking, and none is available, so conclusions are restricted to the studied data, configurations and perturbations. The analysis addresses three research questions:

\begin{itemize}
    \item \textbf{RQ1 (capabilities and limitations).} What capabilities and limitations do locally deployed small open-weight language models exhibit in converting textual opinions into collective evaluations and rankings?

    \item \textbf{RQ2 (linguistic aggregation).} How do linguistic aggregation operators and their placement within the decision pipeline affect evaluation resolution, ranking stability, and sensitivity to crowd composition?

    \item \textbf{RQ3 (fixed-panel fusion and inference budget).} Given that a linear fixed panel cannot separate alternatives more than its best member, how much winner repeatability and independence from any single model does it add, under comparable call and runtime budgets, to repeated single-model inference and greedy decoding?
\end{itemize}

RQ1 uses a common reference configuration, with GPT-5.2 as a hosted anchor; RQ2 changes the operators and the stages at which they act; and RQ3 holds the panels fixed.

\section{Decision Pipeline and Experimental Protocol}
\label{sec:method}

The protocol follows each assessment from the model response to the collective comparison. The reference configuration uses five-label elicitation (LABEL-5, Section~\ref{sec:configs}), continuous arithmetic aggregation, ten runs per model and review, and equal weighting of reviews; operator, participant-weighting and inference-budget comparisons modify identified components of it.

\subsection{Pipeline and Aggregation Hierarchies}

\begin{figure}[t]
    \centering
    \includegraphics[width=\columnwidth]{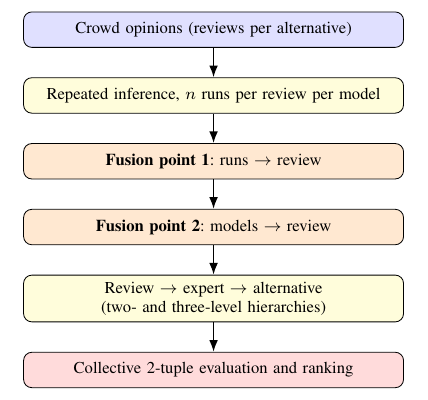}
    \caption{Decision pipeline with repeated-run aggregation, fixed-panel output-level fusion and collective pooling. Collective evaluations determine point rankings and margins; resampling analyses assess stability and construct simultaneous candidate sets.}
    \label{fig:pipeline}
\end{figure}

In the pipeline of Fig.~\ref{fig:pipeline}, repeated responses are aggregated within each model and review, a fixed panel fuses its members' review-level assessments, and the result is pooled into alternative-level evaluations, rankings and margins, whose stability and separation are then assessed by resampling.

Let $\mathcal{A}_{M,r}$ contain the accepted run indices for model $M$ and review $r$, and let $\beta_{M,r}^{(a)}\in[0,g]$, with $g+1$ linguistic terms (Section~\ref{sec:representation}), denote an accepted assessment on the linguistic term-index domain. For an aggregation operator $\Phi$, the model-specific review assessment is
\begin{equation}
\bar{\beta}_{M,r}
=
\Phi\left(
 \left\{\beta_{M,r}^{(a)}:
 a\in\mathcal{A}_{M,r}\right\}
\right).
\end{equation}
Rejected outputs are removed before this calculation, so the number of accepted judgments may differ between reviews.

For a fixed panel $\mathcal{M}$, the fused assessment is
\begin{equation}
\beta_r
=
\Phi\left(
 \left\{\bar{\beta}_{M,r}:M\in\mathcal{M}\right\}
\right).
\end{equation}
For an individual system, $\beta_r=\bar{\beta}_{M,r}$. Each model contributes one assessment to panel fusion; the panel does not pool all accepted raw responses into a single collection. Under continuous averaging, this preserves equal model weights even when accepted-run counts differ.

Two collective pooling hierarchies are examined. Let $\mathcal{R}_{u,j}$ contain the reviews written by participant $u$ about alternative $x_j$, $\mathcal{R}_j$ contain all reviews of that alternative, and $\mathcal{U}_j$ contain its represented participants. The reference, two-level (review-weighted) hierarchy assigns equal weight to reviews:
\begin{equation}
v_j^{(2)}
=
\frac{1}{|\mathcal{R}_j|}
\sum_{r\in\mathcal{R}_j}\beta_r.
\end{equation}
The three-level (participant-weighted) hierarchy first averages within each participant--alternative combination and then assigns equal weight to participants:
\begin{equation}
v_j^{(3)}
=
\frac{1}{|\mathcal{U}_j|}
\sum_{u\in\mathcal{U}_j}
\left(
 \frac{1}{|\mathcal{R}_{u,j}|}
 \sum_{r\in\mathcal{R}_{u,j}}\beta_r
\right).
\end{equation}
The distinction concerns repeated contributions: a participant with several reviews of an alternative receives additional weight under the two-level hierarchy, whereas those reviews are averaged before participant pooling under the three-level hierarchy.

With continuous averaging, linear fusion adds no separation of its own. Let $v_j^{M}$ denote the collective evaluation produced by model $M$ alone and $v_j^{\mathcal{M}}$ that of the fixed panel.

\begin{proposition}\label{prop:fusion}
Let $\Phi$ be the arithmetic mean at the run-to-review and model-to-review stages, let collective pooling follow the two-level hierarchy, and let every member of $\mathcal{M}$ assess every review. Then, in the full data and in every resampling replica, $v_j^{\mathcal{M}}=|\mathcal{M}|^{-1}\sum_{M\in\mathcal{M}}v_j^{M}$. Hence every fused difference is the mean of the members' differences, and $v_i^{\mathcal{M}}-v_j^{\mathcal{M}}\le\max_{M\in\mathcal{M}}(v_i^{M}-v_j^{M})$.
\end{proposition}
\begin{proof}
Pooling weights each review by the multiplicity of its author in the replica, or by one in the full data, and these weights are the same for every member; the average over members and the average over reviews can therefore be exchanged.
\end{proof}

In both corpora, the fused evaluations computed by the pipeline equal the member means exactly. Fusion therefore never gives a pair a larger lead than its best member does, and it shrinks the margin of any pair that members order differently. Under run resampling, in contrast, the members' runs are redrawn independently, so the inference variance of a fused difference is the average of the members' variances divided by $|\mathcal{M}|$: averaging makes the fused winner more repeatable without widening its lead. Medians, ordered weighted averages or reliability weights across members escape this bound but are not evaluated here.

In the principal operator comparison, $\Phi$ acts at the run-to-review and model-to-review stages; subsequent collective pooling uses arithmetic means. A separate all-stage comparison also replaces the final review-to-alternative mean with $\Phi$, separating the discretization of review assessments from that of the collective evaluations (Table~\ref{tab:stages}).

\subsection{Linguistic Representation and Operators}
\label{sec:representation}

Let $S=\{s_0,\ldots,s_g\}$ be an ordered linguistic term set. The linguistic 2-tuple representation \cite{herrera2000} expresses $\beta\in[0,g]$ as
\begin{equation}
\Delta(\beta)=(s_i,\alpha),
\qquad
i=\lfloor\beta+0.5\rfloor,
\qquad
\alpha=\beta-i,
\end{equation}
with inverse
\begin{equation}
\Delta^{-1}(s_i,\alpha)=i+\alpha=\beta.
\end{equation}
Half-integer values are rounded upward, giving $\alpha\in[-0.5,0.5)$; calculations retain $\beta$ continuously.

For a nonempty collection of accepted inputs $Z=\{z_1,\ldots,z_q\}$ with mean $\bar z=q^{-1}\sum_{\ell=1}^{q}z_\ell$, define $R(z)=\lfloor z+0.5\rfloor$. The four evaluated operators are
\begin{gather}
\Phi_{\mathrm{two\_tuple}}(Z)=\bar z,
\qquad
\Phi_{\mathrm{mean\_round}}(Z)=R(\bar z),
\\
\Phi_{\mathrm{median\_round}}(Z)=R\left(\operatorname{median}(Z)\right),
\\
\Phi_{\mathrm{majority}}(Z)=\min\,\operatorname*{arg\,max}_{t}\textstyle\sum_{\ell=1}^{q}\mathbf{1}(z_\ell=t).
\end{gather}
For an even number of inputs, the median is the arithmetic mean of the two central ordered values. Majority ties are resolved toward the lower term index, whereas $R$ rounds half-integers upward; an alternative tie rule is examined in Table~\ref{tab:cdm_tiebreak}. For the rounded operators, the first aggregation stage produces term indices, which remain valid inputs to subsequent applications of the operator.

The comparison between \texttt{two\_tuple} and \texttt{mean\_round} isolates rounding of the same arithmetic estimator; \texttt{median\_round} and \texttt{majority} also change the estimator, so differences involving them cannot be attributed to information loss alone.

The reference representation assigns equal spacing to adjacent linguistic terms. Sensitivity analyses alter interior-term positions while preserving their order and endpoints, as unbalanced term sets \cite{herrera2008unbalanced} and numerical scales for the 2-tuple model \cite{dong2009numerical} do in a principled way. For LABEL-5, the compressed and stretched positions are $(0,1.6,2,2.4,4)$ and $(0,0.5,2,3.5,4)$, respectively. These two fixed placements, rather than scales estimated from data, examine dependence on the numerical embedding of the ordered vocabulary; the LABEL-7 placements are in the appendices.

\subsection{Decision Problems and Inference Configurations}
\label{sec:configs}

\begin{table*}[t]
\centering
\caption{Experimental coverage and deployment configuration. Local models run with Q4\_K\_M quantization on one NVIDIA GeForce RTX 5090 (32\,GB); the main elicitation is LABEL-5 with ten runs per review and model, with SCORE and LABEL-7 as additional formats; sampled collections request $T=1.0$ and are compared with local greedy decoding. Model sizes are runtime-reported total parameter counts. Ornith 35B and Qwen 3.6 35B are mixture-of-experts models with the same runtime architecture, and Qwen 3.6 35B activates about 3B parameters per token. The London Gemma 4 E4B digest was captured after its April collection; the Madrid one is a re-published checkpoint with a different digest (Tables~\ref{tab:provenance} and~\ref{tab:uniovi_provenance}). The hosted reference is excluded from fusion.}
\label{tab:experimental_configuration}
\small
\begin{tabular}{@{}>{\raggedright\arraybackslash}p{0.20\textwidth}>{\raggedright\arraybackslash}p{0.36\textwidth}>{\raggedright\arraybackslash}p{0.36\textwidth}@{}}
\toprule
\textbf{Property} & \textbf{London} & \textbf{Madrid} \\
\midrule
Corpus & TripR-2020Large & TripAdvisor Restaurant Reviews, version 2.0.0 \\
Alternatives / reviews / participants & 4 / 474 / 132 & 12 / 600 / 599 \\
Review coverage & 95--148 reviews per alternative & 50 reviews per alternative \\
Local models & Gemma 4 E4B (8.0B); Ministral 3 8B (8.9B); Gemma 4 31B (31.3B); Ornith 35B (34.7B); Qwen 3.6 35B (36.0B) & Gemma 4 E4B (7.5B); Ministral 3 8B (8.9B); Qwen 3.6 35B (36.0B) \\
Hosted reference / fixed fusion & GPT-5.2 / all five local models & None / all three local models \\
Serving runtime / collection & Ollama 0.31.1 (July collection) / April and July 2026 & Ollama 0.40.4 / 14 October 2026 \\
\bottomrule
\end{tabular}
\end{table*}

\paragraph{Decision problems} Table~\ref{tab:experimental_configuration} summarizes both problems. The London problem uses TripR-2020Large \cite{zuheros2022crowd}, comprising 474 English TripAdvisor reviews from 132 participants. The four alternatives are The Ivy, The Wolseley, J.~Sheekey and Oxo Tower, with 148, 134, 97 and 95 reviews, respectively.

The Madrid problem uses version 2.0.0 of the TripAdvisor Restaurant Reviews dataset \cite{uniovi2025tripadvisor}. A selection protocol fixed before inference identified, within the most populated price band, the twelve restaurants with the most eligible English reviews (at least 20 tokens and a valid rating) and randomly sampled fifty reviews per restaurant, a selection that favors popular restaurants within one price band; the protocol and its seed are recorded in the released data manifest. The resulting sample contains 600 reviews from 599 participants.

London evaluates five open-weight models, Gemma~4 E4B, Ministral~3 8B, Gemma~4 31B, Ornith~35B and Qwen~3.6 35B \cite{gemma2026gemma4,liu2026ministral3,deepreinforce2026ornith,qwen2026qwen36}, served through Ollama~\cite{ollama}, with GPT-5.2~\cite{openai2025gpt52} as a hosted reference excluded from fusion. Madrid evaluates Gemma~4 E4B, Ministral~3 8B and Qwen~3.6 35B, so the fixed panels contain five local members in London and three in Madrid, unchanged throughout the fusion and budget comparisons.

\paragraph{Elicitation and decoding} The principal decision analyses use LABEL-5, which asks for one of five labels from \emph{very negative} to \emph{very positive}, with ten requested responses per review and model. Its prompt reads ``Classify the sentiment of the following text as very\_positive, positive, neutral, negative or very\_negative, the answer must be a single label and one word:'', followed by the review. Both problems additionally include SCORE, a number in $[-1,1]$, and LABEL-7, which adds \emph{extremely negative} and \emph{extremely positive}, each with ten responses per review and model (Table~\ref{tab:prompts}). SCORE supports output-validity assessment; LABEL-7 additionally supports representation sensitivity.

Accepted outputs are normalized to $y\in[-1,1]$ and mapped to the term-index domain through
\begin{equation}
\beta=\frac{g(y+1)}{2}.
\end{equation}
Labels are mapped to equally spaced points of $[-1,1]$, so that under LABEL-5, with $g=4$, the term $s_i$ maps to $\beta=i$.

The main sampled collection requests temperature $T=1.0$ and a maximum output length of 2,048 tokens; other decoding parameters retain model-specific defaults. At $T=1.0$, the mean of repeated runs estimates the expected label under each model's predictive distribution, as truncated by those defaults, and this is the quantity that additional runs approximate. The request overrides the default temperatures of the Ministral 3 8B and Ornith 35B model files, 0.15 and 0.6 (Table~\ref{tab:decoding}), so their run-to-run spread partly reflects that choice. For GPT-5.2, the records establish the requested temperature but not how the provider applied it. Local greedy-decoding collections use $T=0$; both problems include ten greedy responses per review and local model (Tables~\ref{tab:cdm_temperature} and~\ref{tab:uniovi_cdm_temperature}), retained as observed executions rather than assumed to be identical. Reasoning traces were disabled where this setting was recorded; it is unknown for the initial London Gemma~4 E4B and GPT-5.2 collections. Tables~\ref{tab:prompts}--\ref{tab:uniovi_provenance} provide prompts, decoding records, output validity and model provenance.

\paragraph{Parsing and rejected outputs} After presentation formatting is removed, an ordinal response must identify exactly one term of the requested vocabulary, with longer entries matched first so that \emph{very positive} is not parsed as \emph{positive}; numerical responses outside $[-1,1]$ are rejected. The one exception is a label joined to a word outside the vocabulary, such as \texttt{exceptionally\_positive}, which is read as that label; three sampled replies are resolved this way, and one of them sets Ministral 3 8B's narrowest London gap at $0.0008$ rather than $0.00003$.

Rejected outputs are logged and excluded from aggregation; accepted-run counts are retained for each review and model. A review--model combination with no accepted response is omitted: the collective evaluation of that model averages the reviews it did assess, and fixed-panel fusion averages, with equal weights and without imputation, the members available for each review. With all ten sampled runs every review has at least two accepted runs from every model, so Proposition~\ref{prop:fusion} applies; omissions arise in the greedy collections, where Ministral 3 8B has no accepted response for eight London reviews and one Madrid review, and in small run subsets of the budget analysis. A separate listwise sensitivity analysis removes reviews with any rejected run.

\paragraph{External rating reference} Reviewers' star ratings are withheld from prompts and aggregation. For external agreement, a rating of $h$ stars is encoded as $\beta^\star=h-1$, matching the LABEL-5 range and placing three stars on the central term; the signed differences reported below depend on this alignment of the two midpoints. Agreement is assessed through review-level correlations and signed differences. The ratings provide an auxiliary reference: stars evaluate the visit, whereas the extracted assessment concerns its textual description.

\subsection{Resolution, Margins and Resampling Diagnostics}
\label{sec:diagnostics}

\paragraph{Variability and retained resolution} For each configuration, we report between-review standard deviation $\mathrm{SD}_b$, within-review raw-run spread $\mathrm{SD}_{\mathrm{in}}$, and residual aggregate spread $\mathrm{SD}_{\mathrm{res}}$ under run resampling. The latter two are pooled as the square root of the mean per-review variance, which for $\mathrm{SD}_{\mathrm{in}}$ uses divisor $n$ over the $n$ accepted runs. Residual variability is estimated using 5,000 replicas. For continuous averaging it has a closed form: resampling $n$ accepted runs leaves a spread of the run-level standard deviation divided by $\sqrt{n}$, so with ten runs $\mathrm{SD}_{\mathrm{res}}\approx\mathrm{SD}_{\mathrm{in}}/\sqrt{10}$. The comparison with rounded operators therefore measures the spread that rounding adds to this floor.

The number of distinct values is reported separately at the review and alternative levels, because continuous pooling can recover collective resolution above discretized review assessments.

\paragraph{Observed margins and ranking persistence} Collective evaluations are ranked in descending order. The winner's lead is the difference between the largest and second-largest evaluations; the smallest adjacent gap is the minimum difference between consecutive evaluations in that order. Both are expressed in $\beta$ units. Ties yield a zero gap.

Resampling summaries include mean Kendall correlation with the full-data ranking, exact-ranking reproduction and Top-1 persistence. For a full-data winner $w$, Top-1 persistence is the proportion of replicas satisfying
\begin{equation}
v_w^{*}\geq\max_{j\neq w}v_j^{*}.
\end{equation}
A tie involving the original winner therefore counts as retention, which favors operators that tie alternatives; Top-1 is compared across operators only where every operator keeps the alternatives distinct, as in London. We call it run Top-1 under run resampling and crowd Top-1 under participant resampling. It measures persistence of the system's own selection and does not estimate decision accuracy.

Ranking agreement is Kendall's $\tau_b$ between rank vectors, computed with \texttt{scipy.stats.kendalltau} \cite{virtanen2020scipy}; the coefficient is undefined only when one ranking ties every alternative, as under all-stage rounding, and is then counted as zero. Collective evaluations are compared after rounding to nine decimals, tied alternatives share the minimum rank, and exact-ranking reproduction compares these rank vectors, tied groups included. When the full data have co-winners, the reference winner is the first in a fixed order of the alternatives; among the ten-run rankings this arises only for Ministral 3 8B under majority voting in Madrid.

\paragraph{Perturbation designs} Three resampling conditions distinguish complementary sources of sensitivity. \emph{Run resampling} draws accepted responses with replacement within each review--model combination, preserving its accepted-response count and holding the participants fixed. \emph{Participant resampling} draws participants with replacement, each bringing all of their reviews across alternatives, so that a participant drawn twice contributes two copies of the cluster; model assessments remain fixed and collective pooling is recomputed. \emph{Joint resampling} varies both. Ranking-stability analyses use 20,000 replicas. Within each corpus, identical participant draws are used across systems and operators, permitting comparisons under matched changes in crowd composition. The three conditions are sensitivity analyses; their results are not additive variance components.

Participant resampling draws as many participants as each corpus contains, 132 in London and 599 in Madrid, with replacement; no replica left an alternative without observations: every alternative received at least 57 sampled participants in London and 20 in Madrid. Random numbers come from NumPy's PCG64 generator \cite{harris2020numpy}, with seed 42 for run, participant and within-restaurant resampling and seed 43 for the run component of joint resampling. Repeating the participant analyses with four further seeds and with 60{,}000 replicas leaves every London candidate set unchanged and changes in Madrid only alternatives at the boundary of a set (Section~\ref{sec:results-rq1}).

Madrid additionally resamples reviewers within each restaurant, keeping fifty per alternative, as a sensitivity analysis aligned with its balanced sampling design. Wilson intervals \cite{wilson1927probable} accompanying resampling percentages describe Monte Carlo precision given the observed data, not population-level uncertainty. For differences in crowd Top-1 between configurations, the latter is assessed by a nested bootstrap \cite{davison1997bootstrap}: each of the first 2,000 participant replicas receives 1,000 inner redraws of its participants, every configuration's Top-1 is recomputed against its own winner in that replica, and the spread of the paired difference across replicas gives its standard error and a normal 95\% interval (Table~\ref{tab:paired_comparisons}).

\subsection{Simultaneous Comparisons and Candidate Sets}

Candidate sets assess which alternatives remain compatible with being best under simultaneous pairwise inference. For a given system and pooling hierarchy, let $\widehat d_{ij}=\widehat v_i-\widehat v_j$ be the observed difference, $d_{ij,b}^{*}$ its value in participant replica $b$, and $\widehat{\mathrm{se}}_{ij}$ the standard deviation of these replicated differences.

For each replica, the standardized maximum deviation over all unordered pairs is
\begin{equation}
Q_b^{*}
=
\max_{\substack{i<j\\
\widehat{\mathrm{se}}_{ij}>0}}
\frac{
 |d_{ij,b}^{*}-\widehat d_{ij}|
}{
 \widehat{\mathrm{se}}_{ij}
}.
\end{equation}
If $c_{0.95}$ is its empirical 95th percentile, the simultaneous interval is
\begin{equation}
I_{ij}
=
\left[
\widehat d_{ij}
-c_{0.95}\widehat{\mathrm{se}}_{ij},
\widehat d_{ij}
+c_{0.95}\widehat{\mathrm{se}}_{ij}
\right].
\end{equation}
Pairs with zero replicated variance receive the point interval $[\widehat d_{ij},\widehat d_{ij}]$. Reverse comparisons use $I_{ji}=-I_{ij}$. The multiplicity adjustment covers all pairs within the evaluated system and configuration; it does not provide simultaneous coverage across every model and experimental comparison.

The candidate set is
\begin{equation}
\mathcal{C}
=
\left\{
x_j:
\nexists\,i\neq j
\text{ such that }\inf I_{ij}>0
\right\}.
\end{equation}

An alternative is excluded only when another has a simultaneously positive difference against it, following the confidence-set perspective on ranking inference \cite{mogstad2024inference}.

Coverage is approximate and relies on treating the observed participant clusters as independent draws from the population of Section~\ref{sec:problem}; it does not extend automatically to all customers or platform users. Two-sided simultaneous intervals over all pairs, computed in one step, provide a conservative basis for identifying the best alternative: one-sided stepwise procedures, such as the confidence set for the best of \cite{mogstad2024inference}, can exclude more alternatives, so the reported sets are at least as large as theirs. Pairwise percentile intervals are reported in the appendices as sensitivity analyses.

\subsection{Fixed-Panel Fusion and Inference Budgets}
\label{sec:budget}

Inference expenditure is evaluated on the collected responses: for each $k\in\{1,2,3,5,10\}$, every $k$-subset of the ten run positions is applied to all reviews and, for a panel, to all members, as a common subset of inference rounds.

A single system requires $k$ calls per review; a panel of $|\mathcal{M}|$ members requires $|\mathcal{M}|k$ calls. For each subset, the pipeline is recomputed and its winner, lead, crowd Top-1 against its own winner and agreement with the ten-run winner are evaluated; a subset that ties the ten-run winner for first place is credited with that winner, which occurs only in Madrid (Table~\ref{tab:uniovi_cdm_budget}). Subsets are summarized by the median and the 5th and 95th percentiles.

To relate the two sources of uncertainty, let $\mathrm{var}_{\mathrm{part}}$ be the variance of a system's lead between its full-data winner and runner-up over the participant replicas, with the model outputs fixed, and $\mathrm{var}_{\mathrm{inf}}(k)$ its variance with the participants fixed when every review receives $k$ runs drawn with replacement from its accepted runs, computed in closed form. Because the lead is a fixed weighted average of review means, $\mathrm{var}_{\mathrm{inf}}(k)=\mathrm{var}_{\mathrm{inf}}(1)/k$ by construction. Following the separation of item and response sampling in \cite{miller2024adding}, we report the ratio $\mathrm{var}_{\mathrm{inf}}(k)/\mathrm{var}_{\mathrm{part}}$ rather than a share of a total variance. Both variances shrink roughly in proportion to the number of reviews, so the ratio depends little on crowd size. As an illustrative sizing rule, $k^{*}$ is the smallest $k$ with $\mathrm{var}_{\mathrm{inf}}(k)\le\theta\,\mathrm{var}_{\mathrm{part}}$, that is, $k\ge 10\,\mathrm{var}_{\mathrm{inf}}(10)/(\theta\,\mathrm{var}_{\mathrm{part}})$, for a stated threshold $\theta$; we use $\theta=10\%$ and, as an alternative, $5\%$.

Runtime expenditure is estimated from per-request median latencies:
\begin{equation}
\widehat t_M(k)=k\,\widetilde t_M,
\qquad
\widehat t_{\mathcal{M}}(k)
=
k\sum_{M\in\mathcal{M}}\widetilde t_M.
\end{equation}
The panel expression is accumulated sequential request time: the five London members do not fit together on the 32~GB card and run one after another. Model loading, concurrent requests, prompt caching and parallel sampling with a shared prompt are not modeled, so the estimates describe sequential single-request latency, are pessimistic for repeated runs of one model and are not deployment costs. They are compared only within each corpus: the same Ministral 3 8B and Qwen 3.6 35B weights have median latencies of 0.13 and 0.19\,s in London and 0.027 and 0.071\,s in the Madrid re-measurement, made one request at a time with a later runtime because the collection kept four requests in flight (Tables~\ref{tab:latency} and~\ref{tab:uniovi_latency}). Single-call greedy deployment takes each of the ten greedy executions in turn, with sampled-collection latency as a cost proxy (Tables~\ref{tab:cdm_budget} and~\ref{tab:uniovi_cdm_budget}). Individual members highlighted as the strongest benchmarks are identified retrospectively; they delimit the measured benefit of the predefined panel, not a prospective selection rule.

\section{Results}
\label{sec:results}

\subsection{Capabilities and Limits of Small Open-Weight Models (RQ1)}
\label{sec:results-rq1}

\paragraph{Usable assessments and external agreement} Table~\ref{tab:individual_capabilities} summarizes the individual models under the reference configuration. Under LABEL-5, all London replies from Gemma 4 E4B, Gemma 4 31B and Ornith 35B are accepted; rejections are concentrated in Ministral 3 8B, which loses 3.0\% of its replies in London and 1.3\% in Madrid, where Gemma 4 E4B and Qwen 3.6 35B reject none (Tables~\ref{tab:cdm_validity} and~\ref{tab:uniovi_cdm_validity}). The London review-level aggregates correlate with the reviewers' own star ratings, with Pearson correlations ranging from $0.752$ to $0.816$ across the local models. Gemma 4 31B, Ornith 35B and Qwen 3.6 35B correlate more strongly than the hosted reference ($r=0.782$), and paired participant-resampling intervals for these differences exclude zero, narrowly for Gemma 4 31B (Table~\ref{tab:stars_paired}). Agreement also depends on the diagnostic: the proportion of review aggregates rounding to the author's star category ranges from 37.8\% for Ministral 3 8B to 65.6\% for Ornith 35B. All systems assign lower evaluations than the stars on average: the mean signed difference is $-0.639$ in $\beta$ units for Ministral 3 8B, sampled well above its default temperature, against $-0.158$ to $-0.045$ for the other local models. In Madrid the correlations are $0.780$--$0.871$ and the mean signed differences $-0.615$ to $-0.006$ (Tables~\ref{tab:cdm_stars} and~\ref{tab:uniovi_cdm_stars}).

\begin{table*}[t]
\centering
\caption{Individual-model capabilities under the reference configuration (LABEL-5, ten runs, continuous aggregation, two-level hierarchy). Lead is the full-data gap between the first two alternatives, in $\beta$ units; Run and Crowd Top-1 are retention rates of each system's own winner, counting ties as retention; Candidates is the simultaneous candidate-set size; $r$ is the review-level Pearson correlation with the stars. The fusions use ten runs per member (50 and 30 calls per review); Stars are the reviewers' ratings aggregated and resampled in the same way. Percentages are descriptive resampling estimates, not decision accuracies.}
\label{tab:individual_capabilities}
\small
\begin{tabular}{llrrrrr}
\toprule
\textbf{Corpus} &
\textbf{Model} &
\textbf{Lead} &
\textbf{Run Top-1} &
\textbf{Crowd Top-1} &
\textbf{Candidates} &
\textbf{$r$} \\
\midrule
London & Gemma 4 E4B
& 0.0171 & 85.8\% & 47.4\% & 4/4 & 0.764 \\
& Ministral 3 8B
& 0.0236 & 90.3\% & 53.8\% & 4/4 & 0.752 \\
& Gemma 4 31B
& 0.0773 & 100.0\% & 71.1\% & 4/4 & 0.798 \\
& Ornith 35B
& 0.0649 & 100.0\% & 70.7\% & 4/4 & 0.816 \\
& Qwen 3.6 35B
& 0.0355 & 99.9\% & 61.3\% & 4/4 & 0.814 \\
& GPT-5.2 (hosted reference)
& 0.1336 & 100.0\% & 81.2\% & 4/4 & 0.782 \\
& Fixed five-model fusion
& 0.0369 & 100.0\% & 60.8\% & 4/4 & 0.814 \\
& Stars (reference)
& 0.0104 & -- & 54.5\% & 2/4 & -- \\
\midrule
Madrid & Gemma 4 E4B
& 0.0340 & 98.7\% & 55.9\% & 5/12 & 0.815 \\
& Ministral 3 8B
& 0.0301 & 88.9\% & 56.8\% & 8/12 & 0.780 \\
& Qwen 3.6 35B
& 0.1280 & 100.0\% & 90.1\% & 6/12 & 0.871 \\
& Fixed three-model fusion
& 0.0580 & 100.0\% & 63.8\% & 6/12 & 0.853 \\
& Stars (reference)
& 0.1200 & -- & 86.4\% & 7/12 & -- \\
\bottomrule
\end{tabular}
\end{table*}

\paragraph{Collective evaluations and margins} In London, every individual system places all four alternatives within the linguistic term \emph{Positive}, so their ordering is carried entirely by the symbolic translation $\alpha$ (Tables~\ref{tab:cdm_alternatives_label5_two_level} and~\ref{tab:uniovi_cdm_alternatives_label5_two_level} give the complete evaluations of both problems). This replicates the 2-tuple scenario of~\cite{zuheros2024large}, which placed the four restaurants within \emph{Positive} in the order J.~Sheekey, The Ivy, The Wolseley, Oxo Tower; Ministral 3 8B, Ornith 35B, Qwen 3.6 35B, GPT-5.2 and the fusion reproduce that order. Four of the five local models rank J.~Sheekey first; Gemma 4 E4B ranks The Ivy first. The local models' winner leads range from $0.0171$ to $0.0773$ in $\beta$ units, and Ministral 3 8B separates The Wolseley and Oxo Tower by only $0.0008$.

The star ranking places The Ivy ahead of J.~Sheekey by $0.0104$; Gemma 4 E4B reproduces it completely, while the other models differ in at least one pair; with so small a margin, disagreement with the stars' point ranking need not indicate a decision error. Published CDM methods also split on these two restaurants: as reported in~\cite{zuheros2024large}, the sentiment-based methods of~\cite{zuheros2022crowd} and~\cite{zuheros2023explainable}, labeled CDM-SA and ECDM-SDAM there, rank The Ivy first, by $0.02$ and $0.002$ on their $[0,1]$ scales, whereas the 2-tuple scenario of that study ranks J.~Sheekey first by $0.027$. The fused lead found here, $0.0369$ in $\beta$ or $0.009$ on $[0,1]$, lies between the first two leads and below the third. In Madrid, Gemma 4 E4B and Qwen 3.6 35B rank Zenith Brunch \& Cocktails--Valverde first, while Ministral 3 8B selects Los Montes de Galicia (leads of $0.0301$--$0.1280$).

\paragraph{Inference stability and participant sensitivity} The local London models retain their full-data winner in 85.8--100.0\% of run replicas, and Gemma 4 31B, Ornith 35B and Qwen 3.6 35B reach at least 99.9\%: their winner is highly repeatable given the observed reviews, although run resampling cannot produce labels absent from the ten observed runs (Section~\ref{sec:limitations}). Redrawing participants reduces crowd Top-1 to 47.4--71.1\%; GPT-5.2 shows the same contrast, with 100.0\% and 81.2\%. Differences in crowd Top-1 between systems are themselves uncertain: their nested-bootstrap standard errors are 6.3--23.0 points in London and 11.8--27.5 in Madrid, and no such difference has a 95\% interval that excludes zero (Table~\ref{tab:paired_comparisons}). On the same redrawn crowds, GPT-5.2's lead exceeds those of Qwen 3.6 35B and of the fusion (paired intervals $[0.008, 0.184]$ and $[0.017, 0.173]$). Its J.~Sheekey$-$The Ivy gap also exceeds that of Gemma 4 E4B, which ranks the pair in the opposite order ($[0.043, 0.253]$); the intervals against Ministral 3 8B, Gemma 4 31B and Ornith 35B include zero.

In Madrid, Gemma 4 E4B, a re-published checkpoint there, retains its winner in 98.7\% of run replicas, against 85.8\% for the London checkpoint, and in 55.9\% of participant replicas; Ministral 3 8B reaches 88.9\% and 56.8\%, and Qwen 3.6 35B 100.0\% and 90.1\%. Joint run-and-participant Top-1 differs from crowd Top-1 by at most 1.2 percentage points. Winner persistence also differs from stability of the complete order: Qwen 3.6 35B reproduces its complete Madrid ranking in only 0.1\% of participant replicas (Tables~\ref{tab:cdm_joint} and~\ref{tab:uniovi_cdm_joint}; Fig.~\ref{fig:london_sensitivity} shows the London intervals).

\paragraph{Supported distinctions between alternatives} The simultaneous intervals retain all four London alternatives as candidates for every local model and for GPT-5.2, so neither near-perfect run stability nor the hosted reference's larger lead identifies a unique winner. This is a matter of resolving power: a difference is separated only if it exceeds $c_{0.95}\widehat{\mathrm{se}}_{ij}$; with $c_{0.95}$ of 2.56--2.57 and pairwise standard errors of 0.097--0.165, that threshold is about 0.25--0.42 in $\beta$ for the London systems, whereas their winners' leads are 0.017--0.134 (Table~\ref{tab:cdm_simultaneous_pairs}). With 132 participants and near-tied alternatives, the London problem cannot separate a winner with any of the evaluated systems. Even the unadjusted pairwise percentile interval between each system's winner and runner-up contains zero in both problems (Tables~\ref{tab:cdm_candidates} and~\ref{tab:uniovi_cdm_candidates}).

The reviewers' own stars, aggregated and resampled in the same way, do not identify a unique winner either: they keep The Ivy and J.~Sheekey in London and seven of the twelve restaurants in Madrid, and hold their own winner on 54.5\% and 86.4\% of redrawn crowds. The absence of a unique winner therefore belongs to the opinions, not only to the models, although the stars separate more than any system in London and keep more candidates than every system except Ministral 3 8B in Madrid. The three London intervals that exclude The Wolseley and Oxo Tower for the stars have lower limits of $0.014$, $0.025$ and $0.083$ (Table~\ref{tab:cdm_simultaneous_pairs}), and on the same redrawn crowds the stars' gaps exceed the fusion's with a paired interval clear of zero only for The Ivy against Oxo Tower ($0.178$, $[0.030, 0.330]$; Table~\ref{tab:stars_paired}).

The Madrid models narrow the twelve alternatives to five candidates for Gemma 4 E4B, eight for Ministral 3 8B and six for Qwen 3.6 35B, unchanged under within-restaurant resampling, but none singles out one alternative (Tables~\ref{tab:uniovi_cdm_candidates} and~\ref{tab:uniovi_cdm_simultaneous_pairs}). Membership at the boundary is subject to Monte Carlo error. Ministral 3 8B keeps a ninth alternative with two of four other seeds and with 60{,}000 replicas, because the lower limit of the interval between its second and ninth alternatives lies within that error of zero; the stars' set shrinks from seven to six alternatives with one seed and under within-restaurant resampling. The different ordering of models by Top-1 and by set size shows why these diagnostics should be read jointly.

\paragraph{Answer to RQ1} Under the studied configurations, small open-weight models provide usable assessments that can be aggregated locally into collective evaluations and rankings, with substantial review-level agreement with the stars ($r$ of 0.752--0.816 in London and 0.780--0.871 in Madrid) and, for several configurations, highly repeatable winners. Participant sensitivity exceeds inference sensitivity, and simultaneous intervals identify no unique winner, for the systems as for the stars; in London, every observed lead is below what 132 participants allow these intervals to resolve. The models thus supply assessments from which the protocol can report collective comparisons with explicit margins and candidate sets; the evidence does not establish the accuracy of a forced single-winner decision.

\subsection{Effects of Linguistic Aggregation (RQ2)}
\label{sec:results-rq2}

\paragraph{Resolution and residual variability} We next vary the aggregation operator and the stages at which it acts, using the same model judgments (Tables~\ref{tab:cdm_stability} and~\ref{tab:cdm_discriminance} give the London comparisons). In London, continuous aggregation produces 21--65 distinct review-level assessments across the local models, whereas each rounded operator produces five. With ten accepted runs the mean can take only 41 values, so larger counts, such as Ministral 3 8B's 65, come from reviews with rejected runs. Distinct counts do not establish semantic accuracy; the ratio of between-review to residual spread more directly measures how well an operator separates reviews, and it is 6.7--33.5 for continuous averaging against 3.2--15.2 for the rounded operators across the local models (Table~\ref{tab:cdm_discriminance}).

Continuous averaging also leaves less residual variability under run resampling. For the local models, rounding the mean increases $\mathrm{SD}_{\mathrm{res}}$ from $0.033$--$0.138$ to $0.073$--$0.280$. Across all three rounded operators and the six London systems, including the hosted reference, residual spread is 1.85--2.27 times that of continuous averaging: a continuous aggregate near a category boundary moves slightly between replicas, whereas its rounded counterpart jumps by a whole term when the boundary is crossed. In Madrid, continuous aggregation leaves residual spread of $0.075$--$0.121$, against $0.148$--$0.250$ for the rounded operators, factors of 1.94--2.13 computed from the unrounded spreads (Table~\ref{tab:uniovi_cdm_discriminance}).

\paragraph{Consequences for ranking stability} Lower residual spread does not order the operators uniformly by ranking stability. For Qwen 3.6 35B in London, continuous aggregation retains its winner in 99.9\% of run replicas, against 95.3--97.5\% under the rounded operators, and Gemma 4 31B and Ornith 35B retain theirs in at least 99.7\% under every operator. For Ministral 3 8B, however, \texttt{median\_round} retains the winner slightly more often than continuous averaging (91.1\% against 90.3\%), and majority voting reproduces the complete ranking more often (58.0\% against 46.8\%), because the continuous ranking contains a fragile $0.0008$ gap between its bottom alternatives. Under participant resampling, review-level rounding can also raise crowd Top-1, from 53.8\% to 71.0\% for Ministral 3 8B with \texttt{median\_round} and from 60.8\% to 74.2\% for the fusion with \texttt{majority} (Table~\ref{tab:cdm_crowd}), without any gain in resolution.

Operator choice can also change the point decision. For Gemma 4 E4B in London, continuous averaging ranks The Ivy first, whereas \texttt{mean\_round} and \texttt{median\_round} rank J.~Sheekey first, as does majority voting when its ties are broken upward (Table~\ref{tab:cdm_tiebreak}). This change therefore reflects the direction of the rounding and tie conventions as well as the estimator, and since stability is measured against each operator's own ranking, it describes persistence of the answer that operator produces.

\paragraph{The stage of rounding determines the remaining resolution} When rounding is restricted to the review-level stages, continuous pooling still gives all four London alternatives distinct collective values under every operator; in Madrid, rounded operators yield as few as ten distinct values for twelve alternatives, whereas continuous aggregation keeps twelve for each model (Table~\ref{tab:uniovi_cdm_scope}). Rounding at every stage has a more pronounced effect. In London, \texttt{mean\_round} and \texttt{majority} assign the same collective value to all four alternatives for every system. Under \texttt{median\_round}, three systems also collapse to one value and the other three retain only two distinct values. In Madrid, all-stage rounding leaves only one or two distinct collective values for each of the three models. This collapse follows from the data rather than from the models: a final-stage operator returns one of the $g+1$ terms, so alternatives whose aggregates fall on the same term must tie, and every London collective evaluation lies within \emph{Positive}.

Thus, in London the non-zero margins obtained with review-level rounding depend on the continuous averaging performed above that stage; in Madrid, review-level rounding already ties at least two alternatives in eight of the nine operator--model combinations (Tables~\ref{tab:cdm_scope} and~\ref{tab:uniovi_cdm_scope}). An operator's name alone does not specify the behavior of the decision pipeline.

\paragraph{Representation and participant sensitivity} Retaining resolution does not remove sensitivity to the sampled participants or to the pooling assumptions. In London, participant-weighted aggregation changes Ministral 3 8B's winner and reorders the bottom alternatives for Gemma 4 E4B and Gemma 4 31B, whose affected pairs are separated by only $0.0009$--$0.0059$; Ornith 35B and Qwen 3.6 35B keep their complete order (Table~\ref{tab:cdm_alternatives_label5_three_level}). In Madrid the two hierarchies coincide, because no participant contributes several reviews of the same alternative.

Term placement introduces another sensitivity. Under stretched LABEL-5 spacing, Gemma 4 E4B changes its London winner, and Ministral 3 8B and Gemma 4 31B reorder their bottom alternatives (Table~\ref{tab:cdm_mapping}). In Madrid, Gemma 4 E4B and Ministral 3 8B each change their winner in two of the four displaced-placement conditions, while Qwen 3.6 35B retains its winner throughout (Table~\ref{tab:uniovi_cdm_mapping}). Continuous aggregation preserves the distinctions a representation implies, not their independence from it.

\paragraph{Answer to RQ2} Continuous arithmetic aggregation retains finer review-level resolution and leaves approximately half the residual standard deviation of the rounded operators on both corpora. This is the loss that motivated the 2-tuple model, measured here on the review-level inputs to the decision. Its downstream effect depends on the aggregation stage and on the margins between alternatives. Rounding at every stage removes the distinctions required for ranking whenever the collective evaluations share a term, while review-level rounding can alter winners or raise crowd Top-1 without adding information. Continuous aggregation therefore preserves the information available for comparison without adding rounding noise or a rounding direction; whether that information separates the alternatives is assessed as in RQ1.

\subsection{Fixed-Panel Fusion and Inference Budget (RQ3)}
\label{sec:results-rq3}

\paragraph{Fixed-panel fusion against individual members} We compare the fixed five-member London panel and three-member Madrid panel with their members, first with ten runs per model and then at specified call or runtime budgets. By Proposition~\ref{prop:fusion}, the fused lead is the mean of the members' leads on the same pair, so the question is what fusion adds in repeatability and in independence from single members. With ten runs per member, the London fusion retains its winner in 100.0\% of run replicas and reproduces its complete ranking in 99.0\%, as expected from averaging the run noise of five members (Table~\ref{tab:cdm_fusion_vs_best}). Its review-level agreement with the stars ($r=0.814$) is close to that of the best member ($0.816$ for Ornith 35B) and above that of the worst ($0.752$). Its lead of $0.0369$ is the mean of the members' gaps between J.~Sheekey and The Ivy, which range from $-0.0171$ for Gemma 4 E4B, which ranks the pair in reverse, to $0.0773$ (Table~\ref{tab:cdm_simultaneous_pairs}). Its crowd Top-1 of 60.8\% accordingly lies within the members' range of 47.4--71.1\%, below those of Gemma 4 31B (71.1\%) and Ornith 35B (70.7\%).

These differences do not establish inferiority of fusion. On the same redrawn crowds, the lead of Gemma 4 31B exceeds the fused lead by an amount whose 95\% percentile interval is $[-0.020,0.099]$, and that of Ornith 35B by $[-0.019,0.079]$; the nested-bootstrap intervals of their crowd Top-1 advantages, 10.4 and 9.9 points ($[-7.7, 28.5]$ and $[-8.3, 28.1]$), also include zero. The data do not distinguish the fixed fusion from its strongest observed members in either direction.

In Madrid, the fusion again reaches 100.0\% run Top-1, with 98.1\% exact-ranking reproduction. Its crowd Top-1 of 63.8\% lies above Gemma 4 E4B and Ministral 3 8B and below Qwen 3.6 35B's 90.1\%, with nested-bootstrap intervals that include zero (Table~\ref{tab:paired_comparisons}), and both the fusion and Qwen 3.6 35B keep six candidates. Qwen 3.6 35B also agrees more closely with the stars than the fusion ($r$ of 0.871 against 0.853; paired difference $+0.017$, $[0.007, 0.028]$; Table~\ref{tab:stars_paired}), whereas the fusion's collective ranking is closer to the star ranking than any member's ($\tau_b$ of 0.95 against 0.89; Table~\ref{tab:uniovi_cdm_stars}).

\paragraph{Repetition and budget allocation} In London (Table~\ref{tab:fusion_budget}), increasing the runs per panel member from one to ten raises the call count from five to fifty, while median crowd Top-1 changes from 60.5\% to 60.8\%. At five calls per review, five runs of Gemma 4 31B reach 71.1\%, against 60.5\% for one run of each member, and the same percentages hold at ten calls. Equal calls are not equal cost: at five calls, the estimated runtime is 1.58\,s per review for Gemma 4 31B and 1.10\,s for the fusion, and with ten runs per member the fusion needs 10.98\,s, against 3.17\,s for Gemma 4 31B and 1.61\,s for Ornith 35B.

In Madrid, fusion crowd Top-1 rises from 57.7\% with one run per member to 63.8\% with ten, a difference of 6.1 points whose nested-bootstrap interval, $[-5.8, 18.0]$, includes zero. One sampled run of Qwen 3.6 35B reaches a median of 87.3\%, compared with 63.8\% for thirty calls to the panel, at estimated costs of 0.07\,s and 1.32\,s per review. At the equal budget of three calls, Qwen 3.6 35B reaches 88.9\% and the fusion 57.7\%, with estimated costs of 0.21\,s and 0.13\,s. None of these budget differences in crowd Top-1 has a nested-bootstrap 95\% interval that excludes zero (Table~\ref{tab:paired_comparisons}).

\begin{table*}[t]
\centering
\caption{Selected inference-budget comparisons under LABEL-5, continuous aggregation and the two-level hierarchy. $k$ is the number of sampled runs per model. Crowd Top-1 is the median over run subsets of the retention rate of each subset's own winner, with the 5th and 95th percentiles over subsets (a single subset at $k=10$); because the $k$-subsets of ten fixed runs overlap, these percentiles understate the variability of a new $k$-run collection. Same winner is the share of subsets that select the ten-run winner. Cost is estimated from median request latencies, summed over members for fusion, and is comparable only within each corpus.}
\label{tab:fusion_budget}
\small
\begin{tabular}{llrrrrrr}
\toprule
\textbf{Corpus} &
\textbf{Configuration} &
\textbf{$k$} &
\textbf{Calls} &
\textbf{Crowd Top-1} &
\textbf{5th--95th} &
\textbf{Same winner} &
\textbf{Cost (s)} \\
\midrule
London & Gemma 4 31B
& 5 & 5 & 71.1\% & 69.7--72.5 & 100\% & 1.58 \\
& Fixed fusion
& 1 & 5 & 60.5\% & 57.2--65.6 & 100\% & 1.10 \\
& Gemma 4 31B
& 10 & 10 & 71.1\% & -- & 100\% & 3.17 \\
& Fixed fusion
& 2 & 10 & 60.5\% & 57.4--65.1 & 100\% & 2.20 \\
& Fixed fusion
& 10 & 50 & 60.8\% & -- & 100\% & 10.98 \\
\midrule
Madrid & Qwen 3.6 35B
& 1 & 1 & 87.3\% & 68.7--94.6 & 100\% & 0.07 \\
& Qwen 3.6 35B
& 3 & 3 & 88.9\% & 79.1--94.2 & 100\% & 0.21 \\
& Fixed fusion
& 1 & 3 & 57.7\% & 52.6--67.6 & 100\% & 0.13 \\
& Qwen 3.6 35B
& 10 & 10 & 90.1\% & -- & 100\% & 0.71 \\
& Fixed fusion
& 10 & 30 & 63.8\% & -- & 100\% & 1.32 \\
\bottomrule
\end{tabular}
\end{table*}

\paragraph{Agreement on the selected winner} Repetition and fusion stabilize which alternative is selected without materially increasing its crowd Top-1. In London, a single sampled run agrees with the ten-run winner in 7 of 10 subsets for Gemma 4 E4B, 5 of 10 for Ministral 3 8B and 8 of 10 for Qwen 3.6 35B, whereas both fusions select their ten-run winner in every subset at every budget.

\paragraph{Greedy decoding as a deployment alternative} Greedy decoding greatly reduces within-review variation, with problem-dependent consequences. In London, Gemma 4 31B, Ornith 35B, Qwen 3.6 35B and the fusion retain the sampled collection's complete ranking. For Gemma 4 E4B and Ministral 3 8B, the ten-run greedy aggregate ranks first a different alternative from the sampled one; taken one at a time, every greedy run of Gemma 4 E4B and nine of ten of Ministral 3 8B do so (Table~\ref{tab:cdm_budget}), and for Gemma 4 E4B the contrast also spans two collection campaigns. In Madrid, every single-run greedy subset retains its system's sampled winner; greedy crowd Top-1 is 69.2\% for Gemma 4 E4B and 66.5\% for Ministral 3 8B, above their sampled 55.9\% and 56.8\%, and 79.4\% for Qwen 3.6 35B and 58.6\% for the fusion, below their sampled 90.1\% and 63.8\%. With ten runs per member, the fused review-level aggregate has 175 distinct values under sampled inference and 31 under greedy decoding in the paired London comparison; the corresponding Madrid counts are 127 and 32 (Tables~\ref{tab:cdm_temperature} and~\ref{tab:uniovi_cdm_temperature}). Greedy decoding therefore provides a lower-call alternative whose effects on the winner, margins and resolution must be assessed for the configuration.

\paragraph{Robustness of the fixed fusion} In London, the fused ranking is invariant to the tested term placements, pooling hierarchies and treatment of rejected outputs, whereas among its members only Ornith 35B and Qwen 3.6 35B show the same invariance (Tables~\ref{tab:cdm_alternatives_label5_three_level}--\ref{tab:cdm_missing}). In Madrid, listwise exclusion of reviews with rejected outputs changes the fused winner from Zenith Brunch \& Cocktails--Valverde to Los Montes de Galicia, although each member retains its own winner; the fused analysis drops 43 reviews that Gemma 4 E4B and Qwen 3.6 35B keep, so it is not computed on the same reviews as theirs. Compressed LABEL-7 spacing changes the fused winner in the same direction, so robustness to these conventions does not transfer uniformly across problems (Tables~\ref{tab:uniovi_cdm_mapping} and~\ref{tab:uniovi_cdm_missing}).

\paragraph{Why additional inference moves crowd Top-1 so little} With the observed participants held fixed, the inference variance of the winner's lead at ten runs per review is 0.1--3.1\% of its participant variance in London and 0.8--4.3\% in Madrid (Table~\ref{tab:variance_lead}). Since this variance falls as $1/k$ by construction, the illustrative rule of Section~\ref{sec:budget} asks for one to four runs per review in London and one to five in Madrid with $\theta=10\%$, and for one to seven and two to nine with $\theta=5\%$. The largest requirements belong to Ministral 3 8B, which is sampled well above its default temperature. The rule bounds the contribution of inference to the uncertainty of the lead, which the candidate-set report already carries; it does not make a forced winner repeatable. Near $k^{*}$, the ten-run winner is selected by 76\% of the two-run subsets of Gemma 4 E4B ($k^{*}=2$) and by 72\% and 79\% of the three- and five-run subsets of Ministral 3 8B ($k^{*}=4$; Table~\ref{tab:cdm_budget}). Crowd Top-1, which holds the model outputs fixed, rises with $\mathrm{lead}/\mathrm{se}$, approximately as the standard normal distribution function $F_{\mathcal{N}}(\mathrm{lead}/\mathrm{se})$ corrected for the rivals other than the runner-up (Table~\ref{tab:variance_lead}). Because the participant standard error $\mathrm{se}$ varies little across configurations (0.08--0.13), anything that widens the observed lead raises crowd Top-1, whether or not the wider lead reflects information in the reviews. Review-level majority voting and greedy decoding widen the London fusion's lead from $0.037$ to $0.086$ and $0.061$ and raise its crowd Top-1 from 60.8\% to 74.2\% and 68.2\%, while reducing its distinct review-level values from 171 (ten sampled runs per member) to 5 and to 23 (one greedy run per member; Table~\ref{tab:cdm_temperature}). Crowd Top-1 therefore measures persistence of a selection, not its quality, and is read here only together with resolution, margins and candidate sets.

\paragraph{Answer to RQ3} Linear fixed-panel fusion cannot separate alternatives more than its best member, because its evaluations are the means of its members' (Proposition~\ref{prop:fusion}). What it adds is repeatability and independence from any single model: the fixed panels give highly repeatable winners, select their ten-run winner from one run per member, and lie between the best and the worst member in agreement with the stars, closer to the best, without requiring the best member to be known in advance. Because the inference variance of the lead is small relative to its participant variance, calls beyond a few runs per review bring limited or problem-dependent gains in crowd Top-1. The retrospectively strongest members reach higher point values of crowd Top-1 at comparable budgets, as the proposition and the approximation by $F_{\mathcal{N}}$ lead one to expect, by margins the data cannot separate from sampling variability; greedy decoding offers further savings with configuration-dependent effects.

\section{Discussion and Limitations}
\label{sec:discussion}

\subsection{Implications for Local Decision-System Design}

The findings support a conditional role for small open-weight models in local crowd decision-making: the studied configurations turn textual opinions into collective assessments on one consumer GPU, but the unit of assessment is the configured pipeline. The three research questions give distinct requirements, namely usable assessments, aggregation that keeps the distinctions available, and inference whose contribution to the decision is measured; meeting one does not meet the others. Because the inference variance of the lead falls below a tenth of its participant variance within one to five runs per review (Section~\ref{sec:results-rq3}), further calls address only a small part of the sensitivity. In London, moreover, every observed lead lies well below the difference that simultaneous intervals can resolve with 132 participants, so opinion coverage, not inference, limits separation; this study does not estimate how many more opinions would be needed. A fixed panel should be judged by what it delivers at its measured cost: panel size alone does not justify fusion, and linear fusion buys repeatability and independence from one model rather than separation.

Local deployment makes reproducibility and audit possible only when checkpoint identifiers, quantization, prompts, parsing rules and decoding settings are recorded with the data and aggregation configuration; access to weights provides the basis for that record, not the validity of the extracted judgments.

\subsection{Reporting Decisions Under Uncertainty}

A complete order is readily obtained from point estimates, including differences too small to survive plausible perturbations. We therefore recommend reporting collective evaluations and their margins together with the perturbation behind each stability estimate: run Top-1 describes repeatability given the observed opinions, crowd Top-1 describes sensitivity to the sampled crowd, and agreement with an external rating reference is a further check whose meaning depends on how the two constructs relate. When simultaneous intervals leave several alternatives as candidates, the system can present that set with the point ranking and the comparisons that support each exclusion. Membership means only that an alternative has not been excluded from being best; it does not establish equivalence, and failing to separate a pair does not make their difference negligible. Explaining which passages or criteria produced an assessment would require representations beyond the single-polarity pipeline studied here.

\subsection{Limitations}
\label{sec:limitations}

The empirical scope comprises two English-language restaurant-review corpora with closely rated alternatives. Madrid broadens the decision problem but also changes participant coverage, serving runtime and the model panel, and nearly all its participants contribute one review, so the comparison examines whether patterns carry over; it does not isolate the effect of alternative count or establish generalization to other domains. No ground-truth collective decision is available, and star ratings are a related external annotation rather than a measure of decision accuracy; pretraining data may also contain these reviews and their ratings, which could inflate agreement with the stars. Each review is reduced to one polarity, although the London corpus is multi-criteria; this isolates the aggregation pipeline from criterion weights, which could reorder near-tied alternatives.

Model comparisons are descriptive. Training, architecture, parameter count and decoding configuration vary jointly, and Ornith 35B and Qwen 3.6 35B share an architecture, which limits the diversity of the London panel. Each elicitation format uses one prompt wording with a fixed label order, local models use one quantization, and the common temperature does not create equivalent sampling conditions (Section~\ref{sec:configs}). Some initial reasoning-trace settings and checkpoint provenance were not recorded, notably for the London Gemma 4 E4B, the only model that ranks The Ivy first there, and timing estimates combine collection-time measurements with identified re-measurements. The study therefore supports conclusions about the documented configurations rather than causal claims about model size or deployment efficiency.

The uncertainty analysis assumes independent participant clusters drawn from a population resembling the observed reviewers, which platform self-selection limits, and bootstrap simultaneous coverage is approximate. Resampling does not detect shared systematic errors in model judgments. Run resampling redraws the observed runs, so it cannot reflect labels that a model would produce only rarely, and with ten runs it understates inference variability for reviews whose runs all agree. Greedy decoding is not fully deterministic either: Gemma 4 E4B and Ministral 3 8B give ten identical greedy responses for 97--99\% of reviews (Tables~\ref{tab:cdm_temperature} and~\ref{tab:uniovi_cdm_temperature}). Rejections are not random: the Ministral 3 8B reviews with rejected runs average $+0.50$ against $+0.28$ for the rest in London, and $+0.51$ against $+0.45$ in Madrid, on the normalized scale (Tables~\ref{tab:cdm_missing} and~\ref{tab:uniovi_cdm_missing}), so discarding those runs may bias their review means. Equal term spacing, rejected-output treatment and participant weighting are further modeling assumptions; the sensitivity analyses examine selected alternatives to them.

Finally, budget comparisons reuse the collected runs, and the strongest individual benchmarks are identified retrospectively; they do not establish performance on unseen runs or validate a prospective model-selection policy. Label probabilities, where the runtime exposes them, would estimate the expected assessment of a review in one call without sampling noise; this alternative is not evaluated here, and for models that reason before answering it would condition on one reasoning trace.

\section{Conclusion}
\label{sec:conclusion}

Across two restaurant-review problems, locally deployed small open-weight models produced usable assessments and rankings whose resolution, stability and separation could be examined explicitly. Continuous arithmetic aggregation, expressible as linguistic 2-tuples, preserved intermediate assessments, while rounding at successive stages could remove most distinctions between alternatives. Repeated inference and fixed-panel fusion made winners highly repeatable, but participant sensitivity dominated: with ten runs per review, the inference variance of the winner's lead was at most a few percent of its participant variance, one to five runs brought it within a tenth, and further calls brought little. Linear fusion averaged its members' evaluations and showed no distinguishable advantage over its retrospectively strongest members; its contribution lay in repeatability and in not depending on one model. No system identified a unique winner under simultaneous intervals, and neither did the reviewers' own stars. Local decision support with these models should therefore keep assessments continuous, size inference to the participant uncertainty, and report margins and candidate sets alongside point rankings. Future work should validate prospective panel selection under budget constraints, extend evaluation across domains and languages, and examine criterion-level assessments.

\section*{Data and Code Availability}
\ifanonymous
The code, the per-run model outputs and the scripts that regenerate every table and figure will be released in a public repository upon publication.
\else
The code, the per-run model outputs and the scripts that regenerate every table and figure are available at \url{https://github.com/ari-dasci/S-SOWLM-CDM}.
\fi

\bibliographystyle{IEEEtran}
\bibliography{bibliography-IEEE}

\clearpage
\appendix
\setcounter{section}{0}\renewcommand{\thesection}{S\arabic{section}}
\setcounter{table}{0}\renewcommand{\thetable}{S\arabic{table}}
\setcounter{figure}{0}\renewcommand{\thefigure}{S\arabic{figure}}
\section*{Appendices}
\addcontentsline{toc}{section}{Appendices}
These appendices document the experimental configurations and detailed analyses supporting the main text. Sections S1--S3 provide a worked example, aggregation stages, elicitation and parsing protocols, model provenance and deployment costs. Sections S4--S8 report collective evaluations, external agreement, operator comparisons, resampling diagnostics, simultaneous candidate sets and sensitivity analyses. Section S9 documents fixed-panel inference budgets and greedy-decoding comparisons, and Section S10 provides additional graphical diagnostics. Section S11 gives the Madrid counterparts of the validity, latency, star-agreement, operator and sensitivity tables, and Section S12 the paired comparisons and the ratio of inference to participant variance cited in the main text. The fixed panels contain five local models in London and three in Madrid; the hosted reference is excluded from fusion. The London problem uses TripR-2020Large and the Madrid problem version 2.0.0 of the TripAdvisor Restaurant Reviews dataset; these corpora, the models, the serving runtime and the statistical methods are cited in the reference list of the main text. Throughout, Lead is the gap between a system's winner and its runner-up, and Min gap the smallest gap between adjacently ranked alternatives, both in $\beta$ units.

The materials support RQ1 (individual capabilities and participant sensitivity), RQ2 (operators, resolution and processing conventions), and RQ3 (fixed-panel inference budgets). Exhaustive model-subset searches, dependence diagnostics and expanded-panel results are excluded: they concern panel composition rather than the fixed-panel evaluation.

The two London run-resampling tables use separate Monte Carlo streams. For Gemma 4 E4B, 86.1\% in the operator comparison and 85.8\% in the joint-sensitivity analysis reflect that difference, not different full-data configurations. Wilson intervals quantify Monte Carlo precision. Simultaneous intervals instead approximate population coverage under independent participant-cluster sampling. Equal calls do not imply equal runtime or computation.

\section{Worked example and aggregation stages}

\begin{table}[H]
\centering
\begin{adjustbox}{max totalsize={\textwidth}{0.95\textheight},center}
\begin{minipage}{\textwidth}
\examplebox{Worked example: \texttt{review-300}}{%
\textit{``We went to the Ivy for dinner on the last night before our friends flew back to the USA. We had made a reservation for 7:15 pm and the restaurant was very busy with a mixture of earlier dinners and people about to leave to go on to the theatre. We started with some predinner drinks which unfortunately took some time for our server to get right and when they came the mixers had already been poured into the glasses [\ldots{} 284 more words]''}

\smallskip\noindent Author's rating: 4 stars. Alternative: The Ivy. Run labels are given as term indices, from 0 (\emph{Very Negative}) to 4 (\emph{Very Positive}).

\smallskip\noindent{\footnotesize\begin{tabular}{@{}lll@{}}
System & Ten runs & 2-tuple \\
\hline
Gemma 4 E4B & 4 4 4 4 3 3 4 4 4 3 & (Very Positive, $-$0.30) \\
Ministral 3 8B & 1 1 1 1 1 1 1 1 2 1 & (Negative, +0.10) \\
Gemma 4 31B & 3 3 3 3 3 3 3 3 3 3 & (Positive, +0.00) \\
Ornith 35B & 3 3 3 3 3 4 4 3 3 3 & (Positive, +0.20) \\
Qwen 3.6 35B & 3 3 3 3 3 3 3 3 3 3 & (Positive, +0.00) \\
GPT-5.2 & 3 3 3 3 3 3 3 3 3 3 & (Positive, +0.00) \\
Fusion (open-weight) & -- & (Positive, $-$0.20) \\
\end{tabular}} }
\caption{The review on which the systems disagree most under LABEL-5 (largest range of the review-level $\beta$ across systems, chosen by \texttt{utils/worked\_example.py}), its ten runs per system, and the 2-tuples they aggregate to.}
\label{tab:cdm_worked_example}
\end{minipage}
\end{adjustbox}
\end{table}

\begin{table}[H]
\centering
\begin{adjustbox}{max totalsize={\textwidth}{0.95\textheight},center}
\begin{minipage}{\textwidth}
\centering
\caption{Where each operator acts. $\Phi$ is the operator; \emph{mean} is the arithmetic mean on the continuous $\beta$ domain. The reported pipeline applies $\Phi$ only at the run-to-review and model-to-review stages, which is why a rounded operator can still yield non-zero margins between alternatives: the term-valued review evaluations are averaged, not voted on, at the levels above (in Madrid some alternatives still tie; Table~\ref{tab:uniovi_cdm_scope}). The rows follow the three-level hierarchy; the two-level one (the main text) averages the reviews of an alternative in a single step. The \texttt{all} contrast of the main text, computed on the two-level hierarchy, replaces that single mean with $\Phi$.}
\label{tab:stages}
\small
\adjustbox{max width=\linewidth}{%
\begin{tabular}{llc}
\toprule
\textbf{Stage} & \textbf{Reported pipeline} & \textbf{Domain of output} \\
\midrule
Runs $\rightarrow$ review & $\Phi$ & term if $\Phi$ rounds \\
Models $\rightarrow$ review & $\Phi$ & term if $\Phi$ rounds \\
Reviews $\rightarrow$ participant & mean & continuous $\beta$ \\
Participants $\rightarrow$ alternative & mean & continuous $\beta$ \\
\bottomrule
\end{tabular}%
}
\end{minipage}
\end{adjustbox}
\end{table}

\section{Elicitation, decoding and output validity}

\paragraph{Parsing policy} A reply is first stripped of formatting that carries no meaning: markdown emphasis, a leading ``Sentiment:'', backslash escapes and surrounding punctuation. If what remains, after case, space and hyphen normalization, is a label of the scheme's vocabulary, it is accepted. Otherwise every vocabulary label is searched for in the reply on word boundaries, longest label first, each match consumed before shorter labels are searched (so ``very positive'' is one label, not two). A reply containing exactly one distinct label is accepted as that label; a reply containing none is rejected as \texttt{no\_label}; a reply containing two or more distinct labels (``neutral to very\_positive'') is rejected as \texttt{ambiguous}. Rejected replies are listed with their reason in \texttt{invalid\_outputs.csv}. The SCORE parser accepts a single number in $[-1, 1]$ and rejects anything else.

\begin{table}[H]
\centering
\begin{adjustbox}{max totalsize={\textwidth}{0.95\textheight},center}
\begin{minipage}{\textwidth}
\centering
\caption{Prompts, verbatim from \texttt{config/experiments.yaml}. The system prompt is empty for every scheme; in every user prompt the instruction is followed by a line break and a space, and then by \texttt{\{review\}}, which is replaced by the review text. The $T = 0$ collection uses the LABEL-5 prompt unchanged. Every call to a locally served model sets the temperature shown, caps the reply at 2{,}048 tokens and disables the reasoning trace (\texttt{think:~false}), except where Table~\ref{tab:decoding} marks the trace setting as unrecorded.}
\label{tab:prompts}
\small
\begin{tabular}{lp{0.70\textwidth}r}
\toprule
\textbf{Scheme} & \textbf{User prompt} & \textbf{$T$} \\
\midrule
  SCORE & \texttt{Classify the sentiment of the following text using a score between -1 and 1, where -1 represents a completely negative sentiment, 0 a neutral sentiment and 1 a completely positive sentiment. The response must be only the score:  \{review\}} & 1.0 \\
  LABEL-7 & \texttt{Classify the sentiment of the following text as extremely\_positive, very\_positive, positive, neutral, negative, very\_negative or extremely\_negative, the answer must be a single label:  \{review\}} & 1.0 \\
  LABEL-5 & \texttt{Classify the sentiment of the following text as very\_positive, positive, neutral, negative or very\_negative, the answer must be a single label and one word:  \{review\}} & 1.0 \\
  LABEL-5, $T = 0$ & \texttt{Classify the sentiment of the following text as very\_positive, positive, neutral, negative or very\_negative, the answer must be a single label and one word:  \{review\}} & 0.0 \\
\bottomrule
\end{tabular}
\end{minipage}
\end{adjustbox}
\end{table}

\begin{table}[H]
\centering
\begin{adjustbox}{max totalsize={\textwidth}{0.95\textheight},center}
\begin{minipage}{\textwidth}
\centering
\caption{London problem (TripR-2020Large, 4 alternatives). Decoding parameters. \emph{Request} is the reasoning-trace setting the runner sent with every call; the temperature and the 2{,}048-token cap it also sent are those of Table~\ref{tab:prompts}. Digest is the one of Table~\ref{tab:provenance}; for Gemma 4 E4B it was captured after the collection and does not establish which weights were used. \emph{Model defaults} are the Modelfile parameters the runtime applied to everything else, as \texttt{ollama show --parameters} reports them; they belong to the manifest, so they are given only where that digest is still the one installed, and are otherwise unrecorded. A request temperature overrides the Modelfile's. The hosted reference used its provider's defaults. The Madrid collection sent the same requests, with the trace off for all three systems (Table~\ref{tab:uniovi_provenance}); for Gemma 4 E4B it also set the Ollama option \texttt{draft\_num\_predict} to 0, disabling the speculative-decoding draft model that the re-published \texttt{gemma4:e4b} tag (digest \texttt{dc35e8d9c606}) ships and that the April London collection did not have.}
\label{tab:decoding}
\small
\begin{tabular}{lllp{0.42\textwidth}}
\toprule
\textbf{System} & \textbf{Digest} & \textbf{Request} & \textbf{Model defaults} \\
\midrule
  Gemma 4 E4B & \texttt{c6eb396dbd59} & think unrecorded & \texttt{unrecorded} \\
  Ministral 3 8B & \texttt{1922accd5827} & think off & \texttt{temperature=0.15} \\
  Gemma 4 31B & \texttt{6316f0629137} & think off & \texttt{temperature=1; top\_k=64; top\_p=0.95} \\
  Ornith 35B & \texttt{5a470e0f652c} & think off & \texttt{temperature=0.6; top\_k=20; top\_p=0.95} \\
  Qwen 3.6 35B & \texttt{07d35212591f} & think off & \texttt{temperature=1; top\_k=20; top\_p=0.95; min\_p=0; presence\_penalty=1.5; repeat\_penalty=1} \\
  GPT-5.2 & -- & think unrecorded & provider default \\
\bottomrule
\end{tabular}
\end{minipage}
\end{adjustbox}
\end{table}

\begin{table}[H]
\centering
\begin{adjustbox}{max totalsize={\textwidth}{0.95\textheight},center}
\begin{minipage}{\textwidth}
\centering
\caption{London problem (TripR-2020Large, 4 alternatives). Output validity per system under the stated parsing policy: replies collected over the three schemes, replies rejected, and the rejected share over the three schemes (\%) and under LABEL-5 alone. Rejected replies leave the requested vocabulary or name more than one of its labels; they are excluded from every reported metric.}
\label{tab:cdm_validity}
\small
\begin{tabular}{lrrrr}
\toprule
\textbf{Model} & \textbf{Collected} & \textbf{Rejected} & \textbf{\%} & \textbf{LABEL-5 \%} \\
\midrule
  Gemma 4 E4B & 14{,}220 & 0 & 0.00\% & 0.00\% \\
  Ministral 3 8B & 14{,}220 & 155 & 1.09\% & 3.04\% \\
  Gemma 4 31B & 14{,}220 & 0 & 0.00\% & 0.00\% \\
  Ornith 35B & 14{,}220 & 0 & 0.00\% & 0.00\% \\
  Qwen 3.6 35B & 14{,}220 & 4 & 0.03\% & 0.06\% \\
  GPT-5.2 & 14{,}220 & 0 & 0.00\% & 0.00\% \\
\bottomrule
\end{tabular}
\end{minipage}
\end{adjustbox}
\end{table}

\section{Model provenance and deployment costs}

\begin{table}[H]
\centering
\begin{adjustbox}{max totalsize={\textwidth}{0.95\textheight},center}
\begin{minipage}{\textwidth}
\centering
\caption{London problem (TripR-2020Large, 4 alternatives). Provenance of the served models. Tag is the identifier passed to the runtime and Digest the manifest hash the runtime resolved that tag to on 2026-07-31, when this table was captured; the tag is mutable and the digest identifies the weights. Gemma 4 E4B was collected in April 2026, so for it the digest does not establish which weights were used; the other four systems were collected on 2026-07-30, the day before the capture. Architecture, parameters, context length (Ctx, in tokens) and quantization are as the runtime reports them, not as the model card advertises them. Trace is the reasoning-trace setting recorded in the run records. GPT-5.2, served through the OpenAI API, carries none of these and is omitted; it was requested under the dated model identifier \texttt{gpt-5.2-2025-12-11} and collected on 2026-04-07 and 2026-04-08.}
\label{tab:provenance}
\small
\adjustbox{max width=\linewidth}{%
\begin{tabular}{llllrrll}
\toprule
\textbf{Model} & \textbf{Tag} & \textbf{Digest} & \textbf{Architecture} & \textbf{Params} & \textbf{Ctx} & \textbf{Quant.} & \textbf{Trace} \\
\midrule
  Gemma 4 E4B & \texttt{gemma4:e4b} & \texttt{c6eb396dbd59} & \texttt{gemma4} & 8.0B & 131072 & Q4\_K\_M & unrecorded \\
  Ministral 3 8B & \texttt{ministral-3:8b} & \texttt{1922accd5827} & \texttt{mistral3} & 8.9B & 262144 & Q4\_K\_M & False \\
  Gemma 4 31B & \texttt{gemma4:31b} & \texttt{6316f0629137} & \texttt{gemma4} & 31.3B & 262144 & Q4\_K\_M & False \\
  Ornith 35B & \texttt{ornith:35b} & \texttt{5a470e0f652c} & \texttt{qwen35moe} & 34.7B & 262144 & Q4\_K\_M & False \\
  Qwen 3.6 35B & \texttt{qwen3.6:35b} & \texttt{07d35212591f} & \texttt{qwen35moe} & 36.0B & 262144 & Q4\_K\_M & False \\
\bottomrule
\end{tabular}%
}
\\[2pt]\footnotesize Serving stack: Ollama 0.31.1, NVIDIA GeForce RTX 5090, NVIDIA driver 610.43.03.
\end{minipage}
\end{adjustbox}
\end{table}

\begin{table}[H]
\centering
\begin{adjustbox}{max totalsize={\textwidth}{0.95\textheight},center}
\begin{minipage}{\textwidth}
\centering
\caption{Madrid problem (TripAdvisor Restaurant Reviews v2.0.0, 12 alternatives). Provenance of the served models. Tag is the identifier passed to the runtime and Digest the manifest hash the runtime resolved that tag to on 2026-10-14, the day of the collection, when this table was captured; the tag is mutable and the digest identifies the weights. Architecture, parameters, context length (Ctx, in tokens) and quantization are as the runtime reports them, not as the model card advertises them. Trace is the reasoning-trace setting recorded in the run records. No hosted system was run on this problem.}
\label{tab:uniovi_provenance}
\small
\adjustbox{max width=\linewidth}{%
\begin{tabular}{llllrrll}
\toprule
\textbf{Model} & \textbf{Tag} & \textbf{Digest} & \textbf{Architecture} & \textbf{Params} & \textbf{Ctx} & \textbf{Quant.} & \textbf{Trace} \\
\midrule
  Gemma 4 E4B & \texttt{gemma4:e4b} & \texttt{dc35e8d9c606} & \texttt{gemma4} & 7.5B & 131072 & Q4\_K\_M & False \\
  Ministral 3 8B & \texttt{ministral-3:8b} & \texttt{1922accd5827} & \texttt{mistral3} & 8.9B & 262144 & Q4\_K\_M & False \\
  Qwen 3.6 35B & \texttt{qwen3.6:35b} & \texttt{07d35212591f} & \texttt{qwen35moe} & 36.0B & 262144 & Q4\_K\_M & False \\
\bottomrule
\end{tabular}%
}
\\[2pt]\footnotesize Serving stack: Ollama 0.40.4, NVIDIA GeForce RTX 5090, NVIDIA driver 615.78.08.
\end{minipage}
\end{adjustbox}
\end{table}

\begin{table}[H]
\centering
\begin{adjustbox}{max totalsize={\textwidth}{0.95\textheight},center}
\begin{minipage}{\textwidth}
\centering
\caption{London problem (TripR-2020Large, 4 alternatives). Wall-clock latency per request on a single consumer GPU, measured around the backend call over every collected run. $n$ is the number of requests; median and the 95th percentile are reported alongside the mean because a model that emits a reasoning trace on a minority of reviews has a mean far above its typical request. Source distinguishes the two ways a row can be obtained: \emph{instrumented} means every request of the collection was timed as it was made; \emph{remeasured} means the collection predates that instrumentation and the figure comes from replaying the first 60 reviews of the corpus under all three schemes through the same experiment configuration on the same hardware (180 requests, of which the first 3 per scheme are discarded as warm-up, leaving 171). Systems with neither are omitted rather than estimated.}
\label{tab:latency}
\adjustbox{max width=\linewidth}{%
\begin{tabular}{lrrrrl}
\toprule
\textbf{Model} & \textbf{$n$} & \textbf{Mean (s)} & \textbf{Median (s)} & \textbf{p95 (s)} & \textbf{Source} \\
\midrule
  Gemma 4 E4B & 171 & 0.35 & 0.30 & 0.50 & remeasured \\
  Ministral 3 8B & 14{,}220 & 0.14 & 0.13 & 0.15 & instrumented \\
  Gemma 4 31B & 14{,}220 & 0.32 & 0.32 & 0.38 & instrumented \\
  Ornith 35B & 14{,}220 & 0.17 & 0.16 & 0.21 & instrumented \\
  Qwen 3.6 35B & 14{,}220 & 0.19 & 0.19 & 0.24 & instrumented \\
\bottomrule
\end{tabular}%
}
\end{minipage}
\end{adjustbox}
\end{table}

\section{Collective evaluations and external agreement}

\begin{landscape}
\begin{table}[H]
\centering
\begin{adjustbox}{max totalsize={\linewidth}{0.95\textheight},center}
\begin{minipage}{\linewidth}
\centering
\caption{London problem (TripR-2020Large, 4 alternatives). Collective evaluation of each alternative under LABEL-5, aggregated with the 2-tuple operator over the two-level hierarchy. Each cell is $(s_i, \alpha)$; the rank within each column is given in brackets. The fused system aggregates the open-weight models only.}
\label{tab:cdm_alternatives_label5_two_level}
\adjustbox{max width=\linewidth}{%
\begin{tabular}{lccccccc}
\toprule
\textbf{Alternative} & \textbf{Gemma 4 E4B} & \textbf{Ministral 3 8B} & \textbf{Gemma 4 31B} & \textbf{Ornith 35B} & \textbf{Qwen 3.6 35B} & \textbf{GPT-5.2} & \textbf{Fusion (open-weight)} \\
\midrule
  J. Sheekey & (Positive, +0.15) [2] & (Positive, $-$0.31) [1] & (Positive, +0.36) [1] & (Positive, +0.30) [1] & (Positive, +0.34) [1] & (Positive, +0.27) [1] & (Positive, +0.17) [1] \\
  The Ivy & (Positive, +0.16) [1] & (Positive, $-$0.34) [2] & (Positive, +0.28) [2] & (Positive, +0.24) [2] & (Positive, +0.31) [2] & (Positive, +0.14) [2] & (Positive, +0.13) [2] \\
  The Wolseley & (Positive, +0.03) [3] & (Positive, $-$0.45) [3] & (Positive, +0.06) [4] & (Positive, +0.07) [3] & (Positive, +0.10) [3] & (Positive, +0.05) [3] & (Positive, $-$0.04) [3] \\
  Oxo Tower & (Positive, +0.02) [4] & (Positive, $-$0.45) [4] & (Positive, +0.09) [3] & (Positive, +0.03) [4] & (Positive, +0.05) [4] & (Positive, $-$0.03) [4] & (Positive, $-$0.05) [4] \\
\bottomrule
\end{tabular}%
}
\end{minipage}
\end{adjustbox}
\end{table}
\end{landscape}

\begin{table}[H]
\centering
\begin{adjustbox}{max totalsize={\textwidth}{0.95\textheight},center}
\begin{minipage}{\textwidth}
\centering
\caption{Madrid problem (TripAdvisor Restaurant Reviews v2.0.0, 12 alternatives). Collective evaluation of each alternative under LABEL-5, aggregated with the 2-tuple operator over the two-level hierarchy. Each cell is $(s_i, \alpha)$; the rank within each column is given in brackets. The fused system aggregates the open-weight models only. The three-level hierarchy gives identical values: no reviewer contributes more than one review of the same alternative (the one reviewer with two reviews in the sample wrote them about different restaurants).}
\label{tab:uniovi_cdm_alternatives_label5_two_level}
\adjustbox{max width=\linewidth}{%
\begin{tabular}{lcccc}
\toprule
\textbf{Alternative} & \textbf{Gemma 4 E4B} & \textbf{Ministral 3 8B} & \textbf{Qwen 3.6 35B} & \textbf{Fusion (open-weight)} \\
\midrule
  Zenith Brunch \& Cocktails--Valverde & (Very Positive, $-$0.29) [1] & (Positive, +0.23) [2] & (Very Positive, $-$0.06) [1] & (Very Positive, $-$0.37) [1] \\
  Los Montes de Galicia & (Very Positive, $-$0.32) [2] & (Positive, +0.26) [1] & (Very Positive, $-$0.23) [3] & (Very Positive, $-$0.43) [2] \\
  Taberna Mas Al Sur & (Very Positive, $-$0.44) [3] & (Positive, +0.14) [3] & (Very Positive, $-$0.19) [2] & (Very Positive, $-$0.50) [3] \\
  Meson Restaurante La Mi Venta & (Very Positive, $-$0.50) [5] & (Positive, +0.10) [4] & (Very Positive, $-$0.26) [4] & (Positive, +0.45) [4] \\
  Inclan Brutal Bar & (Very Positive, $-$0.50) [4] & (Positive, +0.07) [5] & (Very Positive, $-$0.39) [6] & (Positive, +0.39) [5] \\
  Oven Mozzarella Fuencarral & (Positive, +0.28) [6] & (Positive, 0.00) [6] & (Very Positive, $-$0.36) [5] & (Positive, +0.31) [6] \\
  La Cabana Argentina & (Positive, +0.20) [8] & (Positive, $-$0.16) [7] & (Positive, +0.48) [7] & (Positive, +0.17) [7] \\
  Oven Mozzarella Gran Via 6 & (Positive, +0.23) [7] & (Positive, $-$0.23) [9] & (Positive, +0.43) [8] & (Positive, +0.15) [8] \\
  Rosi La Loca & (Positive, +0.14) [9] & (Positive, $-$0.19) [8] & (Positive, +0.31) [9] & (Positive, +0.09) [9] \\
  Mercado de San Miguel & (Positive, $-$0.05) [10] & (Positive, $-$0.35) [10] & (Positive, +0.27) [10] & (Positive, $-$0.04) [10] \\
  Restaurante Botin & (Positive, $-$0.07) [11] & (Positive, $-$0.39) [11] & (Positive, +0.25) [11] & (Positive, $-$0.07) [11] \\
  La Barraca & (Positive, $-$0.41) [12] & (Neutral, +0.32) [12] & (Positive, $-$0.13) [12] & (Positive, $-$0.41) [12] \\
\bottomrule
\end{tabular}%
}
\end{minipage}
\end{adjustbox}
\end{table}

\begin{table}[H]
\centering
\begin{adjustbox}{max totalsize={\textwidth}{0.95\textheight},center}
\begin{minipage}{\textwidth}
\centering
\caption{London problem (TripR-2020Large, 4 alternatives). Agreement with the 1--5 star rating the author of each review attached to it, over all 474 reviews under LABEL-5. $r$ and $\rho$ are the Pearson and Spearman correlations, over reviews, between the review-level aggregate and the stars, and MAD is the mean absolute difference between them. A rating of $h$ stars is $\beta^{\star} = h-1$, so Diff.\ (the mean signed difference from the stars, negative when the pipeline scores a review below its author) and MAD are in $\beta$ units. Same term is the share of reviews whose aggregate rounds onto the author's own term; $\tau$ compares each collective ranking with the star ranking given beneath. The stars rate a visit, the answer rates the text: related constructs, not one quantity measured twice.}
\label{tab:cdm_stars}
\adjustbox{max width=\linewidth}{%
\begin{tabular}{lccrrcr}
\toprule
\textbf{Model} & \textbf{$r$} & \textbf{$\rho$} & \textbf{Diff.} & \textbf{MAD} & \textbf{Same term} & \textbf{$\tau$} \\
\midrule
  Gemma 4 E4B & 0.764 & 0.640 & $-$0.158 & 0.536 & 54.0\% & +1.00 \\
  Ministral 3 8B & 0.752 & 0.690 & $-$0.639 & 0.723 & 37.8\% & +0.67 \\
  Gemma 4 31B & 0.798 & 0.709 & $-$0.055 & 0.396 & 63.7\% & +0.33 \\
  Ornith 35B & 0.816 & 0.722 & $-$0.089 & 0.385 & 65.6\% & +0.67 \\
  Qwen 3.6 35B & 0.814 & 0.735 & $-$0.045 & 0.405 & 61.6\% & +0.67 \\
  GPT-5.2 & 0.782 & 0.710 & $-$0.146 & 0.444 & 60.8\% & +0.67 \\
  Fusion (open-weight) & 0.814 & 0.728 & $-$0.197 & 0.460 & 65.2\% & +0.67 \\
\midrule
  \multicolumn{7}{l}{Star ranking: The Ivy $>$ J. Sheekey $>$ The Wolseley $>$ Oxo Tower \quad (margin 0.01045)} \\
\bottomrule
\end{tabular}%
}
\end{minipage}
\end{adjustbox}
\end{table}

\section{Operator effects and stage-wise rounding}

\begin{table}[H]
\centering
\begin{adjustbox}{max totalsize={\textwidth}{0.95\textheight},center}
\begin{minipage}{\textwidth}
\centering
\caption{London problem (TripR-2020Large, 4 alternatives). Stability of the collective ranking under resampling of each review's repeated runs, with the crowd held fixed: this is the decision's exposure to the models' own inference variability, and says nothing about its exposure to who happened to write the reviews (Table~\ref{tab:cdm_crowd}). $\tau$ is the mean Kendall correlation between a replica's ranking and the full-data ranking; Exact is the percentage of replicas reproducing that ranking outright; Top-1 the percentage in which no alternative strictly beats Winner, the alternative ranked first on the full data (a tie for first keeps it); Lead the $\beta$ lead of Winner over the runner-up and Min gap the smallest $\beta$ gap between any two adjacently ranked alternatives, both on the full data. Brackets give the 95\% Wilson interval over 20{,}000 replicas; an interval printed as [100.0, 100.0] has its lower limit above 99.95\%. Reproducibility under a resampling and the size of the margin are different properties and are reported separately.}
\label{tab:cdm_stability}
\small
\begin{tabular}{llccclrr}
\toprule
\textbf{Model} & \textbf{Operator} & \textbf{$\tau$} & \textbf{Exact} & \textbf{Top-1} & \textbf{Winner} & \textbf{Lead} & \textbf{Min gap} \\
\midrule
  \multirow{4}{*}{Gemma 4 E4B} & \texttt{two\_tuple} & 0.809 & 48.6\% \tiny[47.9, 49.3] & 86.1\% \tiny[85.6, 86.5] & The Ivy & 0.0171 & 0.0027 \\
   & \texttt{mean\_round} & 0.675 & 25.5\% \tiny[24.9, 26.1] & 42.5\% \tiny[41.8, 43.1] & J. Sheekey & 0.0106 & 0.0106 \\
   & \texttt{median\_round} & 0.670 & 24.8\% \tiny[24.2, 25.4] & 41.6\% \tiny[40.9, 42.3] & J. Sheekey & 0.0070 & 0.0070 \\
   & \texttt{majority} & 0.716 & 29.9\% \tiny[29.3, 30.6] & 73.0\% \tiny[72.4, 73.6] & The Ivy & 0.0079 & 0.0079 \\
\midrule
  \multirow{4}{*}{Ministral 3 8B} & \texttt{two\_tuple} & 0.807 & 46.8\% \tiny[46.1, 47.5] & 90.3\% \tiny[89.9, 90.7] & J. Sheekey & 0.0236 & 0.0008 \\
   & \texttt{mean\_round} & 0.795 & 45.7\% \tiny[45.0, 46.4] & 87.2\% \tiny[86.7, 87.7] & J. Sheekey & 0.0556 & 0.0167 \\
   & \texttt{median\_round} & 0.800 & 45.0\% \tiny[44.3, 45.7] & 91.1\% \tiny[90.7, 91.5] & J. Sheekey & 0.1000 & 0.0013 \\
   & \texttt{majority} & 0.841 & 58.0\% \tiny[57.3, 58.7] & 82.3\% \tiny[81.7, 82.8] & J. Sheekey & 0.0609 & 0.0263 \\
\midrule
  \multirow{4}{*}{Gemma 4 31B} & \texttt{two\_tuple} & 1.000 & 100.0\% \tiny[100.0, 100.0] & 100.0\% \tiny[100.0, 100.0] & J. Sheekey & 0.0773 & 0.0229 \\
   & \texttt{mean\_round} & 0.998 & 99.4\% \tiny[99.3, 99.5] & 100.0\% \tiny[100.0, 100.0] & J. Sheekey & 0.0770 & 0.0126 \\
   & \texttt{median\_round} & 0.998 & 99.4\% \tiny[99.3, 99.5] & 100.0\% \tiny[100.0, 100.0] & J. Sheekey & 0.0770 & 0.0126 \\
   & \texttt{majority} & 0.998 & 99.4\% \tiny[99.3, 99.5] & 100.0\% \tiny[100.0, 100.0] & J. Sheekey & 0.0735 & 0.0320 \\
\midrule
  \multirow{4}{*}{Ornith 35B} & \texttt{two\_tuple} & 1.000 & 99.9\% \tiny[99.8, 99.9] & 100.0\% \tiny[100.0, 100.0] & J. Sheekey & 0.0649 & 0.0387 \\
   & \texttt{mean\_round} & 0.987 & 96.1\% \tiny[95.8, 96.3] & 99.8\% \tiny[99.7, 99.8] & J. Sheekey & 0.0664 & 0.0593 \\
   & \texttt{median\_round} & 0.988 & 96.4\% \tiny[96.2, 96.7] & 99.7\% \tiny[99.6, 99.8] & J. Sheekey & 0.0596 & 0.0593 \\
   & \texttt{majority} & 0.972 & 91.7\% \tiny[91.3, 92.1] & 99.7\% \tiny[99.6, 99.8] & J. Sheekey & 0.0795 & 0.0312 \\
\midrule
  \multirow{4}{*}{Qwen 3.6 35B} & \texttt{two\_tuple} & 1.000 & 99.9\% \tiny[99.9, 99.9] & 99.9\% \tiny[99.9, 99.9] & J. Sheekey & 0.0355 & 0.0355 \\
   & \texttt{mean\_round} & 0.962 & 89.0\% \tiny[88.6, 89.4] & 95.4\% \tiny[95.1, 95.7] & J. Sheekey & 0.0504 & 0.0216 \\
   & \texttt{median\_round} & 0.962 & 88.8\% \tiny[88.4, 89.3] & 95.3\% \tiny[95.0, 95.6] & J. Sheekey & 0.0504 & 0.0216 \\
   & \texttt{majority} & 0.963 & 89.1\% \tiny[88.7, 89.5] & 97.5\% \tiny[97.3, 97.7] & J. Sheekey & 0.0294 & 0.0128 \\
\midrule
  \multirow{4}{*}{GPT-5.2} & \texttt{two\_tuple} & 1.000 & 100.0\% \tiny[100.0, 100.0] & 100.0\% \tiny[100.0, 100.0] & J. Sheekey & 0.1336 & 0.0804 \\
   & \texttt{mean\_round} & 1.000 & 100.0\% \tiny[100.0, 100.0] & 100.0\% \tiny[100.0, 100.0] & J. Sheekey & 0.1432 & 0.0764 \\
   & \texttt{median\_round} & 1.000 & 100.0\% \tiny[100.0, 100.0] & 100.0\% \tiny[100.0, 100.0] & J. Sheekey & 0.1432 & 0.0764 \\
   & \texttt{majority} & 1.000 & 100.0\% \tiny[100.0, 100.0] & 100.0\% \tiny[100.0, 100.0] & J. Sheekey & 0.1500 & 0.0764 \\
\bottomrule
\end{tabular}
\end{minipage}
\end{adjustbox}
\end{table}

\begin{table}[H]
\centering
\begin{adjustbox}{max totalsize={\textwidth}{0.95\textheight},center}
\begin{minipage}{\textwidth}
\centering
\caption{London problem (TripR-2020Large, 4 alternatives). Between-review spread against the noise each operator leaves, in $\beta$ units. $\mathit{SD}_b$ is the spread of the aggregated per-review evaluation across reviews; $\mathit{SD}_{\mathrm{in}}$ the spread of the raw runs within a review, identical across operators by construction; $\mathit{SD}_{\mathrm{res}}$ the spread of the operator's own output under resampling of the runs, the noise that survives aggregation. Both within-review spreads are the square root of the mean of the per-review variances. Ratio is $\mathit{SD}_b/\mathit{SD}_{\mathrm{res}}$ and Share $\mathit{SD}^2_b/(\mathit{SD}^2_b+\mathit{SD}^2_{\mathrm{res}})$, descriptive ratios, not an ICC. Distinct counts the different review-level aggregates and the different collective values of the alternatives (two-level hierarchy).}
\label{tab:cdm_discriminance}
\small
\begin{tabular}{llcccccrr}
\toprule
\textbf{Model} & \textbf{Operator} & \textbf{$\mathit{SD}_b$} & \textbf{$\mathit{SD}_{\mathrm{in}}$} & \textbf{$\mathit{SD}_{\mathrm{res}}$} & \textbf{Ratio} & \textbf{Share} & \textbf{Distinct (review)} & \textbf{Distinct (collective)} \\
\midrule
  \multirow{4}{*}{Gemma 4 E4B} & \texttt{two\_tuple} & 1.123 & 0.383 & 0.121 & 9.3 & 0.989 & 39 & 4 \\
   & \texttt{mean\_round} & 1.193 & 0.383 & 0.265 & 4.5 & 0.953 & 5 & 4 \\
   & \texttt{median\_round} & 1.201 & 0.383 & 0.264 & 4.5 & 0.954 & 5 & 4 \\
   & \texttt{majority} & 1.209 & 0.383 & 0.266 & 4.5 & 0.954 & 5 & 4 \\
\midrule
  \multirow{4}{*}{Ministral 3 8B} & \texttt{two\_tuple} & 0.929 & 0.422 & 0.138 & 6.7 & 0.978 & 65 & 4 \\
   & \texttt{mean\_round} & 0.964 & 0.422 & 0.280 & 3.4 & 0.922 & 5 & 4 \\
   & \texttt{median\_round} & 0.966 & 0.422 & 0.288 & 3.4 & 0.918 & 5 & 4 \\
   & \texttt{majority} & 0.955 & 0.422 & 0.296 & 3.2 & 0.912 & 5 & 4 \\
\midrule
  \multirow{4}{*}{Gemma 4 31B} & \texttt{two\_tuple} & 1.109 & 0.105 & 0.033 & 33.5 & 0.999 & 21 & 4 \\
   & \texttt{mean\_round} & 1.109 & 0.105 & 0.073 & 15.2 & 0.996 & 5 & 4 \\
   & \texttt{median\_round} & 1.109 & 0.105 & 0.073 & 15.2 & 0.996 & 5 & 4 \\
   & \texttt{majority} & 1.111 & 0.105 & 0.075 & 14.8 & 0.996 & 5 & 4 \\
\midrule
  \multirow{4}{*}{Ornith 35B} & \texttt{two\_tuple} & 1.013 & 0.295 & 0.093 & 10.8 & 0.992 & 37 & 4 \\
   & \texttt{mean\_round} & 1.049 & 0.295 & 0.190 & 5.5 & 0.968 & 5 & 4 \\
   & \texttt{median\_round} & 1.048 & 0.295 & 0.193 & 5.4 & 0.967 & 5 & 4 \\
   & \texttt{majority} & 1.057 & 0.295 & 0.198 & 5.3 & 0.966 & 5 & 4 \\
\midrule
  \multirow{4}{*}{Qwen 3.6 35B} & \texttt{two\_tuple} & 1.067 & 0.281 & 0.089 & 12.0 & 0.993 & 41 & 4 \\
   & \texttt{mean\_round} & 1.099 & 0.281 & 0.180 & 6.1 & 0.974 & 5 & 4 \\
   & \texttt{median\_round} & 1.099 & 0.281 & 0.180 & 6.1 & 0.974 & 5 & 4 \\
   & \texttt{majority} & 1.113 & 0.281 & 0.188 & 5.9 & 0.972 & 5 & 4 \\
\midrule
  \multirow{4}{*}{GPT-5.2} & \texttt{two\_tuple} & 1.143 & 0.146 & 0.046 & 24.8 & 0.998 & 25 & 4 \\
   & \texttt{mean\_round} & 1.150 & 0.146 & 0.085 & 13.5 & 0.995 & 5 & 4 \\
   & \texttt{median\_round} & 1.150 & 0.146 & 0.085 & 13.5 & 0.995 & 5 & 4 \\
   & \texttt{majority} & 1.153 & 0.146 & 0.085 & 13.6 & 0.995 & 5 & 4 \\
\bottomrule
\end{tabular}
\end{minipage}
\end{adjustbox}
\end{table}

\begin{table}[H]
\centering
\begin{adjustbox}{max totalsize={\textwidth}{0.95\textheight},center}
\begin{minipage}{\textwidth}
\centering
\caption{London problem (TripR-2020Large, 4 alternatives). Where the operator is allowed to act, under LABEL-5 and the two-level hierarchy. \texttt{review} is the reported pipeline: the operator aggregates the ten runs of a review, and the level above carries $\beta$ forward as a continuous value. \texttt{all} applies the same operator again where the reviews of an alternative are pooled, so that the collective evaluation itself lands on a term, which is what a study aggregating ``by majority vote'' throughout would be doing. Distinct is how many different collective values the four alternatives receive, and Min gap the smallest $\beta$ gap between adjacently ranked alternatives (-- where some alternatives tie). An operator acting at every level cannot express a difference smaller than one term, so alternatives come back tied: no rounded operator leaves more than two distinct values here.}
\label{tab:cdm_scope}
\small
\adjustbox{max width=\linewidth}{%
\begin{tabular}{llccr}
\toprule
\textbf{Model} & \textbf{Operator} & \textbf{Scope} & \textbf{Distinct} & \textbf{Min gap} \\
\midrule
  \multirow{8}{*}{Gemma 4 E4B} & \texttt{two\_tuple} & \texttt{review} & 4 & 0.0027 \\
   & \texttt{two\_tuple} & \texttt{all} & 4 & 0.0027 \\
   & \texttt{mean\_round} & \texttt{review} & 4 & 0.0106 \\
   & \texttt{mean\_round} & \texttt{all} & 1 & -- \\
   & \texttt{median\_round} & \texttt{review} & 4 & 0.0070 \\
   & \texttt{median\_round} & \texttt{all} & 1 & -- \\
   & \texttt{majority} & \texttt{review} & 4 & 0.0079 \\
   & \texttt{majority} & \texttt{all} & 1 & -- \\
\midrule
  \multirow{8}{*}{Ministral 3 8B} & \texttt{two\_tuple} & \texttt{review} & 4 & 0.0008 \\
   & \texttt{two\_tuple} & \texttt{all} & 4 & 0.0008 \\
   & \texttt{mean\_round} & \texttt{review} & 4 & 0.0167 \\
   & \texttt{mean\_round} & \texttt{all} & 1 & -- \\
   & \texttt{median\_round} & \texttt{review} & 4 & 0.0013 \\
   & \texttt{median\_round} & \texttt{all} & 1 & -- \\
   & \texttt{majority} & \texttt{review} & 4 & 0.0263 \\
   & \texttt{majority} & \texttt{all} & 1 & -- \\
\midrule
  \multirow{8}{*}{Gemma 4 31B} & \texttt{two\_tuple} & \texttt{review} & 4 & 0.0229 \\
   & \texttt{two\_tuple} & \texttt{all} & 4 & 0.0229 \\
   & \texttt{mean\_round} & \texttt{review} & 4 & 0.0126 \\
   & \texttt{mean\_round} & \texttt{all} & 1 & -- \\
   & \texttt{median\_round} & \texttt{review} & 4 & 0.0126 \\
   & \texttt{median\_round} & \texttt{all} & 2 & -- \\
   & \texttt{majority} & \texttt{review} & 4 & 0.0320 \\
   & \texttt{majority} & \texttt{all} & 1 & -- \\
\midrule
  \multirow{8}{*}{Ornith 35B} & \texttt{two\_tuple} & \texttt{review} & 4 & 0.0387 \\
   & \texttt{two\_tuple} & \texttt{all} & 4 & 0.0387 \\
   & \texttt{mean\_round} & \texttt{review} & 4 & 0.0593 \\
   & \texttt{mean\_round} & \texttt{all} & 1 & -- \\
   & \texttt{median\_round} & \texttt{review} & 4 & 0.0593 \\
   & \texttt{median\_round} & \texttt{all} & 2 & -- \\
   & \texttt{majority} & \texttt{review} & 4 & 0.0312 \\
   & \texttt{majority} & \texttt{all} & 1 & -- \\
\midrule
  \multirow{8}{*}{Qwen 3.6 35B} & \texttt{two\_tuple} & \texttt{review} & 4 & 0.0355 \\
   & \texttt{two\_tuple} & \texttt{all} & 4 & 0.0355 \\
   & \texttt{mean\_round} & \texttt{review} & 4 & 0.0216 \\
   & \texttt{mean\_round} & \texttt{all} & 1 & -- \\
   & \texttt{median\_round} & \texttt{review} & 4 & 0.0216 \\
   & \texttt{median\_round} & \texttt{all} & 1 & -- \\
   & \texttt{majority} & \texttt{review} & 4 & 0.0128 \\
   & \texttt{majority} & \texttt{all} & 1 & -- \\
\midrule
  \multirow{8}{*}{GPT-5.2} & \texttt{two\_tuple} & \texttt{review} & 4 & 0.0804 \\
   & \texttt{two\_tuple} & \texttt{all} & 4 & 0.0804 \\
   & \texttt{mean\_round} & \texttt{review} & 4 & 0.0764 \\
   & \texttt{mean\_round} & \texttt{all} & 1 & -- \\
   & \texttt{median\_round} & \texttt{review} & 4 & 0.0764 \\
   & \texttt{median\_round} & \texttt{all} & 2 & -- \\
   & \texttt{majority} & \texttt{review} & 4 & 0.0764 \\
   & \texttt{majority} & \texttt{all} & 1 & -- \\
\bottomrule
\end{tabular}%
}
\end{minipage}
\end{adjustbox}
\end{table}

\section{Run, participant and joint sensitivity}

\begin{table}[H]
\centering
\begin{adjustbox}{max totalsize={\textwidth}{0.95\textheight},center}
\begin{minipage}{\textwidth}
\centering
\caption{London problem (TripR-2020Large, 4 alternatives). Stability of the collective ranking (LABEL-5, 2-tuple operator, two-level hierarchy) under three resamplings computed by one procedure: \emph{runs} redraws each review's repeated runs with the crowd held fixed; \emph{crowd} redraws the participants as clusters, each bringing all of their reviews, with the runs held fixed; \emph{both} does the two in the same replica, on the same redrawn crowds. For the fused system each member's runs are redrawn independently. Top-1 is the share of replicas in which no alternative strictly beats the system's full-data winner; Exact the share reproducing its full-data ranking. Brackets are 95\% Wilson intervals, one decimal (a printed 100.0 lower limit is above 99.95\%), over 20{,}000 replicas. The \emph{runs} rows use a replica stream separate from that of Table~\ref{tab:cdm_stability}, so they can differ from it within Monte Carlo error (Gemma 4 E4B: Top-1 $85.8\%$ here against $86.1\%$ there). Which rivals the crowd does not separate from the winner is in Table~\ref{tab:cdm_candidates}.}
\label{tab:cdm_joint}
\small
\adjustbox{max width=\linewidth}{%
\begin{tabular}{llcc}
\toprule
\textbf{Model} & \textbf{Resampled} & \textbf{Top-1} & \textbf{Exact} \\
\midrule
  \multirow{3}{*}{Gemma 4 E4B} & runs & 85.8\% \tiny[85.3, 86.3] & 48.4\% \tiny[47.8, 49.1] \\
   & crowd & 47.4\% \tiny[46.7, 48.1] & 17.1\% \tiny[16.5, 17.6] \\
   & both & 47.0\% \tiny[46.3, 47.6] & 16.6\% \tiny[16.1, 17.1] \\
\midrule
  \multirow{3}{*}{Ministral 3 8B} & runs & 90.3\% \tiny[89.9, 90.7] & 46.8\% \tiny[46.1, 47.5] \\
   & crowd & 53.8\% \tiny[53.1, 54.4] & 22.4\% \tiny[21.8, 22.9] \\
   & both & 53.4\% \tiny[52.7, 54.1] & 22.1\% \tiny[21.5, 22.7] \\
\midrule
  \multirow{3}{*}{Gemma 4 31B} & runs & 100.0\% \tiny[100.0, 100.0] & 100.0\% \tiny[100.0, 100.0] \\
   & crowd & 71.1\% \tiny[70.5, 71.7] & 33.5\% \tiny[32.8, 34.2] \\
   & both & 71.2\% \tiny[70.6, 71.9] & 33.4\% \tiny[32.8, 34.1] \\
\midrule
  \multirow{3}{*}{Ornith 35B} & runs & 100.0\% \tiny[100.0, 100.0] & 99.9\% \tiny[99.8, 99.9] \\
   & crowd & 70.7\% \tiny[70.0, 71.3] & 38.2\% \tiny[37.5, 38.9] \\
   & both & 70.3\% \tiny[69.7, 71.0] & 37.8\% \tiny[37.1, 38.5] \\
\midrule
  \multirow{3}{*}{Qwen 3.6 35B} & runs & 99.9\% \tiny[99.9, 99.9] & 99.9\% \tiny[99.9, 99.9] \\
   & crowd & 61.3\% \tiny[60.6, 62.0] & 36.0\% \tiny[35.4, 36.7] \\
   & both & 61.2\% \tiny[60.6, 61.9] & 36.1\% \tiny[35.4, 36.7] \\
\midrule
  \multirow{3}{*}{GPT-5.2} & runs & 100.0\% \tiny[100.0, 100.0] & 100.0\% \tiny[100.0, 100.0] \\
   & crowd & 81.2\% \tiny[80.7, 81.7] & 41.0\% \tiny[40.3, 41.7] \\
   & both & 81.2\% \tiny[80.7, 81.7] & 40.8\% \tiny[40.1, 41.5] \\
\midrule
  \multirow{3}{*}{Fusion (open-weight)} & runs & 100.0\% \tiny[100.0, 100.0] & 99.0\% \tiny[98.8, 99.1] \\
   & crowd & 60.8\% \tiny[60.1, 61.4] & 29.2\% \tiny[28.6, 29.8] \\
   & both & 60.6\% \tiny[59.9, 61.3] & 29.0\% \tiny[28.4, 29.6] \\
\bottomrule
\end{tabular}%
}
\end{minipage}
\end{adjustbox}
\end{table}

\begin{table}[H]
\centering
\begin{adjustbox}{max totalsize={\textwidth}{0.95\textheight},center}
\begin{minipage}{\textwidth}
\centering
\caption{Madrid problem (TripAdvisor Restaurant Reviews v2.0.0, 12 alternatives). Stability of the collective ranking (LABEL-5, 2-tuple operator, two-level hierarchy) under three resamplings computed by one procedure: \emph{runs} redraws each review's repeated runs with the crowd held fixed; \emph{crowd} redraws the participants as clusters, each bringing all of their reviews, with the runs held fixed; \emph{both} does the two in the same replica, on the same redrawn crowds. For the fused system each member's runs are redrawn independently. Top-1 is the share of replicas in which no alternative strictly beats the system's full-data winner; Exact the share reproducing its full-data ranking. Brackets are 95\% Wilson intervals, one decimal (a printed 100.0 lower limit is above 99.95\%), over 20{,}000 replicas. Which rivals the crowd does not separate from the winner is in Table~\ref{tab:uniovi_cdm_candidates}.}
\label{tab:uniovi_cdm_joint}
\small
\adjustbox{max width=\linewidth}{%
\begin{tabular}{llcc}
\toprule
\textbf{Model} & \textbf{Resampled} & \textbf{Top-1} & \textbf{Exact} \\
\midrule
  \multirow{3}{*}{Gemma 4 E4B} & runs & 98.7\% \tiny[98.5, 98.8] & 54.4\% \tiny[53.7, 55.1] \\
   & crowd & 55.9\% \tiny[55.2, 56.6] & 0.3\% \tiny[0.2, 0.4] \\
   & both & 55.7\% \tiny[55.0, 56.4] & 0.3\% \tiny[0.2, 0.4] \\
\midrule
  \multirow{3}{*}{Ministral 3 8B} & runs & 88.9\% \tiny[88.5, 89.4] & 53.8\% \tiny[53.1, 54.5] \\
   & crowd & 56.8\% \tiny[56.1, 57.5] & 0.2\% \tiny[0.1, 0.2] \\
   & both & 55.6\% \tiny[54.9, 56.3] & 0.2\% \tiny[0.1, 0.2] \\
\midrule
  \multirow{3}{*}{Qwen 3.6 35B} & runs & 100.0\% \tiny[100.0, 100.0] & 83.5\% \tiny[83.0, 84.1] \\
   & crowd & 90.1\% \tiny[89.7, 90.5] & 0.1\% \tiny[0.1, 0.2] \\
   & both & 89.5\% \tiny[89.0, 89.9] & 0.1\% \tiny[0.1, 0.1] \\
\midrule
  \multirow{3}{*}{Fusion (open-weight)} & runs & 100.0\% \tiny[100.0, 100.0] & 98.1\% \tiny[97.9, 98.3] \\
   & crowd & 63.8\% \tiny[63.2, 64.5] & 0.3\% \tiny[0.2, 0.4] \\
   & both & 63.4\% \tiny[62.8, 64.1] & 0.3\% \tiny[0.2, 0.4] \\
\bottomrule
\end{tabular}%
}
\end{minipage}
\end{adjustbox}
\end{table}

\begin{table}[H]
\centering
\begin{adjustbox}{max totalsize={\textwidth}{0.95\textheight},center}
\begin{minipage}{\textwidth}
\centering
\caption{London problem (TripR-2020Large, 4 alternatives). Stability of the collective ranking (LABEL-5, two-level hierarchy) under resampling of the \emph{participants}, with each participant carrying all of their reviews and the model outputs held fixed. Reviews by one participant are not independent observations of the crowd, so they are drawn as a cluster. This is the uncertainty a decision maker actually faces (a different sample of reviewers might have ranked the restaurants differently), and it is a different quantity from Table~\ref{tab:cdm_stability}, not a robustness check on it. Every row is scored on the same redrawn crowds, those of the crowd rows of Table~\ref{tab:cdm_joint}; $\tau$, Top-1 and Exact are defined as in Table~\ref{tab:cdm_stability}, and the small brackets after Exact and Top-1 are 95\% Wilson intervals. Min gap gives the full-data smallest adjacent gap with the 2.5th and 97.5th percentiles across 20{,}000 replicas, measured between alternatives consecutive in the full-data ranking so that it goes negative when a replica inverts a pair; Inv.\ is the share of replicas in which it does. An interval reaching below zero means the surveyed crowd does not determine the complete ranking; which rivals it leaves undistinguished from the winner is in Table~\ref{tab:cdm_candidates}.}
\label{tab:cdm_crowd}
\small
\begin{tabular}{llccclr}
\toprule
\textbf{Model} & \textbf{Operator} & \textbf{$\tau$} & \textbf{Exact} & \textbf{Top-1} & \textbf{Min gap [95\%]} & \textbf{Inv.} \\
\midrule
  \multirow{4}{*}{Gemma 4 E4B} & \texttt{two\_tuple} & 0.451 & 17.1\% \tiny[16.5, 17.6] & 47.4\% \tiny[46.7, 48.1] & 0.0027 [$-$0.3015, 0.0649] & 83\% \\
   & \texttt{mean\_round} & 0.469 & 16.9\% \tiny[16.4, 17.4] & 46.6\% \tiny[45.9, 47.3] & 0.0106 [$-$0.3240, 0.0700] & 83\% \\
   & \texttt{median\_round} & 0.443 & 15.8\% \tiny[15.3, 16.3] & 44.8\% \tiny[44.2, 45.5] & 0.0070 [$-$0.3273, 0.0664] & 84\% \\
   & \texttt{majority} & 0.498 & 18.2\% \tiny[17.7, 18.7] & 46.3\% \tiny[45.6, 47.0] & 0.0079 [$-$0.3204, 0.0774] & 82\% \\
\midrule
  \multirow{4}{*}{Ministral 3 8B} & \texttt{two\_tuple} & 0.513 & 22.4\% \tiny[21.8, 22.9] & 53.8\% \tiny[53.1, 54.4] & 0.0008 [$-$0.2484, 0.0729] & 78\% \\
   & \texttt{mean\_round} & 0.443 & 20.8\% \tiny[20.2, 21.4] & 58.6\% \tiny[57.9, 59.3] & 0.0167 [$-$0.2475, 0.0690] & 79\% \\
   & \texttt{median\_round} & 0.473 & 16.8\% \tiny[16.3, 17.4] & 71.0\% \tiny[70.3, 71.6] & 0.0013 [$-$0.2617, 0.0575] & 83\% \\
   & \texttt{majority} & 0.451 & 20.6\% \tiny[20.1, 21.2] & 59.9\% \tiny[59.2, 60.6] & 0.0263 [$-$0.2364, 0.0668] & 79\% \\
\midrule
  \multirow{4}{*}{Gemma 4 31B} & \texttt{two\_tuple} & 0.699 & 33.5\% \tiny[32.8, 34.2] & 71.1\% \tiny[70.5, 71.7] & 0.0229 [$-$0.2970, 0.1139] & 66\% \\
   & \texttt{mean\_round} & 0.681 & 31.0\% \tiny[30.4, 31.7] & 70.8\% \tiny[70.2, 71.4] & 0.0126 [$-$0.3061, 0.1086] & 69\% \\
   & \texttt{median\_round} & 0.681 & 31.0\% \tiny[30.4, 31.7] & 70.8\% \tiny[70.2, 71.4] & 0.0126 [$-$0.3061, 0.1086] & 69\% \\
   & \texttt{majority} & 0.705 & 34.6\% \tiny[33.9, 35.2] & 70.3\% \tiny[69.6, 70.9] & 0.0320 [$-$0.2898, 0.1147] & 65\% \\
\midrule
  \multirow{4}{*}{Ornith 35B} & \texttt{two\_tuple} & 0.713 & 38.2\% \tiny[37.5, 38.9] & 70.7\% \tiny[70.0, 71.3] & 0.0387 [$-$0.2455, 0.1169] & 62\% \\
   & \texttt{mean\_round} & 0.720 & 40.4\% \tiny[39.7, 41.1] & 70.8\% \tiny[70.1, 71.4] & 0.0593 [$-$0.2342, 0.1203] & 60\% \\
   & \texttt{median\_round} & 0.718 & 39.6\% \tiny[39.0, 40.3] & 68.9\% \tiny[68.2, 69.5] & 0.0593 [$-$0.2356, 0.1206] & 60\% \\
   & \texttt{majority} & 0.724 & 38.8\% \tiny[38.2, 39.5] & 74.4\% \tiny[73.8, 75.0] & 0.0312 [$-$0.2628, 0.1225] & 61\% \\
\midrule
  \multirow{4}{*}{Qwen 3.6 35B} & \texttt{two\_tuple} & 0.711 & 36.0\% \tiny[35.4, 36.7] & 61.3\% \tiny[60.6, 62.0] & 0.0355 [$-$0.2569, 0.1243] & 64\% \\
   & \texttt{mean\_round} & 0.686 & 33.6\% \tiny[33.0, 34.3] & 65.0\% \tiny[64.4, 65.7] & 0.0216 [$-$0.2872, 0.1255] & 66\% \\
   & \texttt{median\_round} & 0.686 & 33.6\% \tiny[33.0, 34.3] & 65.0\% \tiny[64.4, 65.7] & 0.0216 [$-$0.2872, 0.1255] & 66\% \\
   & \texttt{majority} & 0.649 & 28.0\% \tiny[27.4, 28.7] & 58.2\% \tiny[57.5, 58.9] & 0.0128 [$-$0.2992, 0.1132] & 72\% \\
\midrule
  \multirow{4}{*}{GPT-5.2} & \texttt{two\_tuple} & 0.696 & 41.0\% \tiny[40.3, 41.7] & 81.2\% \tiny[80.7, 81.7] & 0.0804 [$-$0.2340, 0.1289] & 59\% \\
   & \texttt{mean\_round} & 0.697 & 41.0\% \tiny[40.3, 41.7] & 82.9\% \tiny[82.3, 83.4] & 0.0764 [$-$0.2385, 0.1321] & 59\% \\
   & \texttt{median\_round} & 0.697 & 41.0\% \tiny[40.3, 41.7] & 82.9\% \tiny[82.3, 83.4] & 0.0764 [$-$0.2385, 0.1321] & 59\% \\
   & \texttt{majority} & 0.692 & 40.3\% \tiny[39.6, 41.0] & 83.8\% \tiny[83.3, 84.3] & 0.0764 [$-$0.2399, 0.1310] & 60\% \\
\midrule
  \multirow{4}{*}{Fusion (open-weight)} & \texttt{two\_tuple} & 0.636 & 29.2\% \tiny[28.6, 29.8] & 60.8\% \tiny[60.1, 61.4] & 0.0145 [$-$0.2673, 0.1002] & 71\% \\
   & \texttt{mean\_round} & 0.671 & 33.5\% \tiny[32.8, 34.1] & 65.9\% \tiny[65.2, 66.5] & 0.0251 [$-$0.2736, 0.1180] & 66\% \\
   & \texttt{median\_round} & 0.700 & 36.3\% \tiny[35.6, 37.0] & 67.4\% \tiny[66.7, 68.0] & 0.0356 [$-$0.2653, 0.1246] & 64\% \\
   & \texttt{majority} & 0.737 & 41.8\% \tiny[41.1, 42.5] & 74.2\% \tiny[73.5, 74.8] & 0.0509 [$-$0.2473, 0.1351] & 58\% \\
\bottomrule
\end{tabular}
\end{minipage}
\end{adjustbox}
\end{table}

\section{Candidate sets and simultaneous comparisons}

For each pair, let $\hat d_{ij}$ be its observed difference and $s_{ij}$ the standard deviation of participant-bootstrap differences $d^*_{ij,b}$. The critical value is the empirical 95th percentile of $\max_{i<j:s_{ij}>0}|d^*_{ij,b}-\hat d_{ij}|/s_{ij}$. Intervals are $\hat d_{ij}\pm c s_{ij}$; zero-variance pairs receive point intervals. An alternative is retained unless a rival has an entirely positive interval against it. Coverage is approximate for a population of reviewers resembling the observed sample. Set membership does not establish equivalence. The Madrid table lists only positive intervals; the remaining pairs include zero.

\begin{landscape}
\begin{table}[H]
\centering
\begin{adjustbox}{max totalsize={\linewidth}{0.95\textheight},center}
\begin{minipage}{\linewidth}
\centering
\caption{London problem (TripR-2020Large, 4 alternatives). Candidate sets under resampling of the crowd (LABEL-5, 2-tuple operator, two-level hierarchy). Every difference between two alternatives of a system is recomputed on 20{,}000 redrawn crowds, the same crowds for every system. Set (simult.)\ is the reported candidate set, a confidence set for the best alternative: the alternatives no rival is simultaneously better than, from bootstrap max-$t$ intervals over all pairwise differences of the system (critical value $c$ = 2.55 to 2.57); Not excluded lists its members other than the winner. Coverage is approximate and simultaneous over all pairwise differences of a system, for the population of reviewers of which the collected participants are treated as an independent sample. Set (pairw.)\ is the pairwise sensitivity statement: the winner and every rival whose gap to it has its 2.5th percentile at or below zero, one pair at a time. Neither set shows its members to be equivalent to each other. Lead is the gap between the winner and the runner-up, with its 2.5th and 97.5th percentiles; Adjacent pairs separated applies the pairwise test to pairs consecutive in the full-data ranking and lists the ranks of those it separates. Stars is the reviewers' own ratings aggregated and resampled in the same way, with no run component.}
\label{tab:cdm_candidates}
\small
\adjustbox{max width=\linewidth}{%
\begin{tabular}{llccclcl}
\toprule
\textbf{System} & \textbf{Winner} & \textbf{Top-1} & \textbf{Set (pairw.)} & \textbf{Set (simult.)} & \textbf{Not excluded (simult.)} & \textbf{Lead [95\%]} & \textbf{Adjacent pairs separated} \\
\midrule
  Gemma 4 E4B & The Ivy & 47.4\% & 4 of 4 & 4 of 4 & J. Sheekey, The Wolseley, Oxo Tower & +0.0171 [$-$0.2272, +0.2694] & 0 of 3 \\
  Ministral 3 8B & J. Sheekey & 53.8\% & 4 of 4 & 4 of 4 & The Ivy, The Wolseley, Oxo Tower & +0.0236 [$-$0.1824, +0.2203] & 0 of 3 \\
  Gemma 4 31B & J. Sheekey & 71.1\% & 3 of 4 & 4 of 4 & The Ivy, Oxo Tower, The Wolseley & +0.0773 [$-$0.1709, +0.3212] & 0 of 3 \\
  Ornith 35B & J. Sheekey & 70.7\% & 4 of 4 & 4 of 4 & The Ivy, The Wolseley, Oxo Tower & +0.0649 [$-$0.1556, +0.2820] & 0 of 3 \\
  Qwen 3.6 35B & J. Sheekey & 61.3\% & 4 of 4 & 4 of 4 & The Ivy, The Wolseley, Oxo Tower & +0.0355 [$-$0.1945, +0.2612] & 1 of 3: 2--3 \\
  GPT-5.2 & J. Sheekey & 81.2\% & 4 of 4 & 4 of 4 & The Ivy, The Wolseley, Oxo Tower & +0.1336 [$-$0.1259, +0.3827] & 0 of 3 \\
  Fusion (open-weight) & J. Sheekey & 60.8\% & 4 of 4 & 4 of 4 & The Ivy, The Wolseley, Oxo Tower & +0.0369 [$-$0.1864, +0.2547] & 0 of 3 \\
  Stars (reference) & The Ivy & 54.5\% & 2 of 4 & 2 of 4 & J. Sheekey & +0.0104 [$-$0.1695, +0.2039] & 1 of 3: 2--3 \\
\bottomrule
\end{tabular}%
}
\end{minipage}
\end{adjustbox}
\end{table}
\end{landscape}

\begin{landscape}
\begin{table}[H]
\centering
\begin{adjustbox}{max totalsize={\linewidth}{0.95\textheight},center}
\begin{minipage}{\linewidth}
\centering
\caption{Madrid problem (TripAdvisor Restaurant Reviews v2.0.0, 12 alternatives). Candidate sets under resampling of the crowd (LABEL-5, 2-tuple operator, two-level hierarchy). Every difference between two alternatives of a system is recomputed on 20{,}000 redrawn crowds, the same crowds for every system. Set (simult.)\ is the reported candidate set, a confidence set for the best alternative: the alternatives no rival is simultaneously better than, from bootstrap max-$t$ intervals over all pairwise differences of the system (critical value $c$ = 3.25 to 3.27 under \emph{participants}; 3.24 to 3.26 under \emph{strat.}); Not excluded lists its members other than the winner. Coverage is approximate and simultaneous over all pairwise differences of a system, for the population of reviewers of which the collected participants are treated as an independent sample. Set (pairw.)\ is the pairwise sensitivity statement: the winner and every rival whose gap to it has its 2.5th percentile at or below zero, one pair at a time. Neither set shows its members to be equivalent to each other. Lead is the gap between the winner and the runner-up, with its 2.5th and 97.5th percentiles; Adjacent pairs separated applies the pairwise test to pairs consecutive in the full-data ranking and lists the ranks of those it separates. Alternatives are given by their rank in the system's own full-data ranking. \emph{participants} redraws participants across the corpus; \emph{strat.}\ redraws them within each alternative, keeping its number of reviewers. Stars is the reviewers' own ratings aggregated and resampled in the same way, with no run component; its ranks follow the order of first appearance within the two exact ties of mean stars (3.76 and 3.54), and its exclusion of La Cabana Argentina (lower limit $+0.033$) while Oven Mozzarella Gran Via 6 stays in (lower limit $-0.0002$) is a decision at the boundary.}
\label{tab:uniovi_cdm_candidates}
\small
\adjustbox{max width=\linewidth}{%
\begin{tabular}{lllccclcl}
\toprule
\textbf{System} & \textbf{Crowd} & \textbf{Winner} & \textbf{Top-1} & \textbf{Set (pairw.)} & \textbf{Set (simult.)} & \textbf{Not excluded (simult.)} & \textbf{Lead [95\%]} & \textbf{Adjacent pairs separated} \\
\midrule
  Gemma 4 E4B & participants & Zenith Brunch \& Cocktails--Valverde & 55.9\% & 4 of 12 & 5 of 12 & 2, 3, 4, 5 & +0.0340 [$-$0.1762, +0.2715] & 0 of 11 \\
   & strat. & Zenith Brunch \& Cocktails--Valverde & 56.7\% & 4 of 12 & 5 of 12 & 2, 3, 4, 5 & +0.0340 [$-$0.1700, +0.2680] & 0 of 11 \\
  Ministral 3 8B & participants & Los Montes de Galicia & 56.8\% & 6 of 12 & 8 of 12 & 2, 3, 4, 5, 6, 7, 8 & +0.0301 [$-$0.2116, +0.2535] & 0 of 11 \\
   & strat. & Los Montes de Galicia & 57.7\% & 6 of 12 & 8 of 12 & 2, 3, 4, 5, 6, 7, 8 & +0.0301 [$-$0.2084, +0.2505] & 0 of 11 \\
  Qwen 3.6 35B & participants & Zenith Brunch \& Cocktails--Valverde & 90.1\% & 3 of 12 & 6 of 12 & 2, 3, 4, 5, 6 & +0.1280 [$-$0.0136, +0.3073] & 0 of 11 \\
   & strat. & Zenith Brunch \& Cocktails--Valverde & 91.3\% & 3 of 12 & 6 of 12 & 2, 3, 4, 5, 6 & +0.1280 [$-$0.0100, +0.3040] & 0 of 11 \\
  Fusion (open-weight) & participants & Zenith Brunch \& Cocktails--Valverde & 63.8\% & 5 of 12 & 6 of 12 & 2, 3, 4, 5, 6 & +0.0580 [$-$0.1391, +0.2902] & 0 of 11 \\
   & strat. & Zenith Brunch \& Cocktails--Valverde & 64.3\% & 4 of 12 & 6 of 12 & 2, 3, 4, 5, 6 & +0.0580 [$-$0.1379, +0.2873] & 0 of 11 \\
  Stars (reference) & participants & Zenith Brunch \& Cocktails--Valverde & 86.4\% & 2 of 12 & 7 of 12 & 2, 3, 4, 5, 6, 8 & +0.1200 [$-$0.0638, +0.3455] & 1 of 11: 11--12 \\
   & strat. & Zenith Brunch \& Cocktails--Valverde & 89.7\% & 2 of 12 & 6 of 12 & 2, 3, 4, 5, 8 & +0.1200 [$-$0.0600, +0.3400] & 1 of 11: 11--12 \\
\bottomrule
\end{tabular}%
}
\end{minipage}
\end{adjustbox}
\end{table}
\end{landscape}

\begin{table}[H]
\centering
\begin{adjustbox}{max totalsize={\textwidth}{0.95\textheight},center}
\begin{minipage}{\textwidth}
\centering
\caption{London problem (TripR-2020Large, 4 alternatives). Simultaneous 95\% intervals for the pairwise differences $\beta_i-\beta_j$ under resampling of the crowd (\emph{participants}; LABEL-5, 2-tuple operator, two-level hierarchy), $i$ ranked above $j$ in the system's full data. $\hat d$ is the full-data difference, se its standard deviation over the redrawn crowds, and the interval $\hat d \mp c\,$se with $c$ the system's max-$t$ critical value: $2.57$ for Gemma 4 E4B, Ministral 3 8B, Ornith 35B and GPT-5.2, and $2.56$ for Gemma 4 31B, Qwen 3.6 35B and the fusion; the stars, the reviewers' own ratings aggregated and resampled in the same way, have $c = 2.55$. No system interval lies above zero, so no system finds an alternative simultaneously better than another; three intervals of the stars do.}
\label{tab:cdm_simultaneous_pairs}
\small
\begin{tabular}{lllrrr}
\toprule
\textbf{System} & \textbf{$i$} & \textbf{$j$} & \textbf{$\hat d$} & \textbf{se} & \textbf{Interval} \\
\midrule
  Gemma 4 E4B & The Ivy & J. Sheekey & +0.0171 & 0.1269 & [$-$0.309, +0.343] \\
   & The Ivy & The Wolseley & +0.1366 & 0.1146 & [$-$0.158, +0.431] \\
   & The Ivy & Oxo Tower & +0.1393 & 0.1477 & [$-$0.240, +0.519] \\
   & J. Sheekey & The Wolseley & +0.1195 & 0.1376 & [$-$0.234, +0.473] \\
   & J. Sheekey & Oxo Tower & +0.1222 & 0.1545 & [$-$0.275, +0.519] \\
   & The Wolseley & Oxo Tower & +0.0027 & 0.1461 & [$-$0.373, +0.378] \\
\addlinespace
  Ministral 3 8B & J. Sheekey & The Ivy & +0.0236 & 0.1033 & [$-$0.242, +0.289] \\
   & J. Sheekey & The Wolseley & +0.1396 & 0.1163 & [$-$0.159, +0.439] \\
   & J. Sheekey & Oxo Tower & +0.1405 & 0.1342 & [$-$0.205, +0.486] \\
   & The Ivy & The Wolseley & +0.1160 & 0.0986 & [$-$0.138, +0.370] \\
   & The Ivy & Oxo Tower & +0.1169 & 0.1139 & [$-$0.176, +0.410] \\
   & The Wolseley & Oxo Tower & +0.0008 & 0.1221 & [$-$0.313, +0.315] \\
\addlinespace
  Gemma 4 31B & J. Sheekey & The Ivy & +0.0773 & 0.1261 & [$-$0.246, +0.400] \\
   & J. Sheekey & Oxo Tower & +0.2714 & 0.1604 & [$-$0.140, +0.683] \\
   & J. Sheekey & The Wolseley & +0.2943 & 0.1404 & [$-$0.066, +0.654] \\
   & The Ivy & Oxo Tower & +0.1941 & 0.1492 & [$-$0.188, +0.576] \\
   & The Ivy & The Wolseley & +0.2170 & 0.1105 & [$-$0.066, +0.500] \\
   & Oxo Tower & The Wolseley & +0.0229 & 0.1557 & [$-$0.376, +0.422] \\
\addlinespace
  Ornith 35B & J. Sheekey & The Ivy & +0.0649 & 0.1125 & [$-$0.224, +0.354] \\
   & J. Sheekey & The Wolseley & +0.2276 & 0.1305 & [$-$0.108, +0.563] \\
   & J. Sheekey & Oxo Tower & +0.2663 & 0.1426 & [$-$0.100, +0.632] \\
   & The Ivy & The Wolseley & +0.1628 & 0.0967 & [$-$0.086, +0.411] \\
   & The Ivy & Oxo Tower & +0.2014 & 0.1310 & [$-$0.135, +0.538] \\
   & The Wolseley & Oxo Tower & +0.0387 & 0.1404 & [$-$0.322, +0.399] \\
\addlinespace
  Qwen 3.6 35B & J. Sheekey & The Ivy & +0.0355 & 0.1171 & [$-$0.264, +0.335] \\
   & J. Sheekey & The Wolseley & +0.2404 & 0.1347 & [$-$0.104, +0.585] \\
   & J. Sheekey & Oxo Tower & +0.2938 & 0.1518 & [$-$0.095, +0.682] \\
   & The Ivy & The Wolseley & +0.2049 & 0.1017 & [$-$0.055, +0.465] \\
   & The Ivy & Oxo Tower & +0.2583 & 0.1386 & [$-$0.096, +0.613] \\
   & The Wolseley & Oxo Tower & +0.0534 & 0.1493 & [$-$0.329, +0.435] \\
\addlinespace
  GPT-5.2 & J. Sheekey & The Ivy & +0.1336 & 0.1295 & [$-$0.199, +0.466] \\
   & J. Sheekey & The Wolseley & +0.2223 & 0.1468 & [$-$0.155, +0.600] \\
   & J. Sheekey & Oxo Tower & +0.3027 & 0.1650 & [$-$0.121, +0.727] \\
   & The Ivy & The Wolseley & +0.0887 & 0.1162 & [$-$0.210, +0.387] \\
   & The Ivy & Oxo Tower & +0.1691 & 0.1503 & [$-$0.217, +0.555] \\
   & The Wolseley & Oxo Tower & +0.0804 & 0.1519 & [$-$0.310, +0.471] \\
\addlinespace
  Fusion (open-weight) & J. Sheekey & The Ivy & +0.0369 & 0.1127 & [$-$0.252, +0.326] \\
   & J. Sheekey & The Wolseley & +0.2043 & 0.1275 & [$-$0.123, +0.531] \\
   & J. Sheekey & Oxo Tower & +0.2188 & 0.1441 & [$-$0.150, +0.588] \\
   & The Ivy & The Wolseley & +0.1675 & 0.1002 & [$-$0.089, +0.424] \\
   & The Ivy & Oxo Tower & +0.1820 & 0.1318 & [$-$0.156, +0.520] \\
   & The Wolseley & Oxo Tower & +0.0145 & 0.1388 & [$-$0.341, +0.370] \\
\addlinespace
  Stars (reference) & The Ivy & J. Sheekey & +0.0104 & 0.0963 & [$-$0.235, +0.256] \\
   & The Ivy & The Wolseley & +0.2352 & 0.0868 & [+0.014, +0.457] \\
   & The Ivy & Oxo Tower & +0.3603 & 0.1086 & [+0.083, +0.637] \\
   & J. Sheekey & The Wolseley & +0.2247 & 0.1093 & [$-$0.054, +0.504] \\
   & J. Sheekey & Oxo Tower & +0.3499 & 0.1273 & [+0.025, +0.675] \\
   & The Wolseley & Oxo Tower & +0.1251 & 0.1261 & [$-$0.197, +0.447] \\
\bottomrule
\end{tabular}
\end{minipage}
\end{adjustbox}
\end{table}

\begin{table}[H]
\centering
\begin{adjustbox}{max totalsize={\textwidth}{0.95\textheight},center}
\begin{minipage}{\textwidth}
\centering
\caption{Madrid problem (TripAdvisor Restaurant Reviews v2.0.0, 12 alternatives). Simultaneous 95\% intervals for the pairwise differences $\beta_i-\beta_j$ under resampling of the crowd (\emph{participants}; LABEL-5, 2-tuple operator, two-level hierarchy), $i$ ranked above $j$ in the system's full data. $\hat d$ is the full-data difference and the interval is $\hat d \mp c\,$se, with se the standard deviation of the difference over the redrawn crowds (not shown) and $c$ the system's max-$t$ critical value. Alternatives are given by their rank in the system's own full-data ranking (Table~\ref{tab:uniovi_cdm_alternatives_label5_two_level}). Only the pairs whose interval lies above zero are listed; the remaining pairs of each system have intervals containing zero. The stars are not listed here: their candidate set, and the two lower limits that decide its boundary, are given in Table~\ref{tab:uniovi_cdm_candidates}.}
\label{tab:uniovi_cdm_simultaneous_pairs}
\small
\begin{tabular}{lrrl}
\toprule
\textbf{System} & \textbf{$c$} & \textbf{Pairs above zero} & \textbf{$(i, j)$: $\hat d$ [interval]} \\
\midrule
  Gemma 4 E4B & 3.27 & 20 of 66 & \parbox[t]{0.56\textwidth}{(1,~6): 0.430 [0.041, 0.819]; (1,~7): 0.480 [0.053, 0.907]; (1,~8): 0.514 [0.082, 0.946]; (1,~9): 0.572 [0.098, 1.046]; (1,~10): 0.762 [0.427, 1.097]; (1,~11): 0.788 [0.358, 1.218]; (1,~12): 1.122 [0.529, 1.715]; (2,~10): 0.728 [0.293, 1.163]; (2,~11): 0.754 [0.240, 1.268]; (2,~12): 1.088 [0.432, 1.744]; (3,~10): 0.610 [0.218, 1.002]; (3,~11): 0.636 [0.163, 1.109]; (3,~12): 0.970 [0.343, 1.597]; (4,~10): 0.552 [0.083, 1.021]; (4,~11): 0.578 [0.036, 1.120]; (4,~12): 0.912 [0.231, 1.593]; (5,~10): 0.548 [0.135, 0.961]; (5,~11): 0.574 [0.084, 1.064]; (5,~12): 0.908 [0.267, 1.549]; (6,~12): 0.692 [0.024, 1.360]} \\
  Ministral 3 8B & 3.26 & 13 of 66 & \parbox[t]{0.56\textwidth}{(1,~10): 0.618 [0.150, 1.086]; (1,~11): 0.650 [0.136, 1.163]; (1,~12): 0.948 [0.358, 1.538]; (2,~9): 0.459 [0.003, 0.915]; (2,~10): 0.588 [0.180, 0.996]; (2,~11): 0.620 [0.163, 1.076]; (2,~12): 0.918 [0.371, 1.464]; (3,~10): 0.490 [0.066, 0.915]; (3,~11): 0.522 [0.051, 0.993]; (3,~12): 0.820 [0.264, 1.377]; (4,~12): 0.785 [0.201, 1.368]; (5,~12): 0.751 [0.149, 1.352]; (6,~12): 0.680 [0.084, 1.277]} \\
  Qwen 3.6 35B & 3.27 & 13 of 66 & \parbox[t]{0.56\textwidth}{(1,~7): 0.460 [0.051, 0.869]; (1,~8): 0.504 [0.064, 0.944]; (1,~9): 0.626 [0.077, 1.175]; (1,~10): 0.664 [0.284, 1.044]; (1,~11): 0.690 [0.237, 1.143]; (1,~12): 1.062 [0.395, 1.729]; (2,~10): 0.536 [0.097, 0.975]; (2,~11): 0.562 [0.063, 1.061]; (2,~12): 0.934 [0.231, 1.637]; (3,~12): 0.892 [0.144, 1.640]; (4,~10): 0.470 [0.005, 0.935]; (4,~12): 0.868 [0.149, 1.587]; (5,~12): 0.768 [0.008, 1.528]} \\
  Fusion (open-weight) & 3.26 & 17 of 66 & \parbox[t]{0.56\textwidth}{(1,~7): 0.457 [0.055, 0.859]; (1,~8): 0.481 [0.065, 0.897]; (1,~9): 0.539 [0.065, 1.013]; (1,~10): 0.671 [0.323, 1.020]; (1,~11): 0.699 [0.277, 1.122]; (1,~12): 1.034 [0.452, 1.616]; (2,~10): 0.613 [0.163, 1.064]; (2,~11): 0.641 [0.130, 1.153]; (2,~12): 0.976 [0.329, 1.623]; (3,~10): 0.545 [0.152, 0.939]; (3,~11): 0.573 [0.116, 1.031]; (3,~12): 0.908 [0.299, 1.517]; (4,~10): 0.491 [0.070, 0.912]; (4,~11): 0.519 [0.038, 1.000]; (4,~12): 0.854 [0.226, 1.481]; (5,~12): 0.799 [0.121, 1.477]; (6,~12): 0.713 [0.058, 1.369]} \\
\bottomrule
\end{tabular}
\end{minipage}
\end{adjustbox}
\end{table}

\section{Representation and processing sensitivity}

\begin{landscape}
\begin{table}[H]
\centering
\begin{adjustbox}{max totalsize={\linewidth}{0.95\textheight},center}
\begin{minipage}{\linewidth}
\centering
\caption{London problem (TripR-2020Large, 4 alternatives). Collective evaluation of each alternative under LABEL-5, aggregated with the 2-tuple operator over the three-level hierarchy. Each cell is $(s_i, \alpha)$; the rank within each column is given in brackets. The fused system aggregates the open-weight models only.}
\label{tab:cdm_alternatives_label5_three_level}
\adjustbox{max width=\linewidth}{%
\begin{tabular}{lccccccc}
\toprule
\textbf{Alternative} & \textbf{Gemma 4 E4B} & \textbf{Ministral 3 8B} & \textbf{Gemma 4 31B} & \textbf{Ornith 35B} & \textbf{Qwen 3.6 35B} & \textbf{GPT-5.2} & \textbf{Fusion (open-weight)} \\
\midrule
  J. Sheekey & (Positive, +0.11) [2] & (Positive, $-$0.36) [2] & (Positive, +0.32) [1] & (Positive, +0.27) [1] & (Positive, +0.31) [1] & (Positive, +0.24) [1] & (Positive, +0.13) [1] \\
  The Ivy & (Positive, +0.13) [1] & (Positive, $-$0.35) [1] & (Positive, +0.25) [2] & (Positive, +0.20) [2] & (Positive, +0.27) [2] & (Positive, +0.10) [2] & (Positive, +0.10) [2] \\
  The Wolseley & (Positive, +0.03) [4] & (Positive, $-$0.44) [3] & (Positive, +0.09) [3] & (Positive, +0.08) [3] & (Positive, +0.11) [3] & (Positive, +0.06) [3] & (Positive, $-$0.03) [3] \\
  Oxo Tower & (Positive, +0.03) [3] & (Positive, $-$0.45) [4] & (Positive, +0.09) [4] & (Positive, +0.04) [4] & (Positive, +0.05) [4] & (Positive, $-$0.03) [4] & (Positive, $-$0.05) [4] \\
\bottomrule
\end{tabular}%
}
\end{minipage}
\end{adjustbox}
\end{table}
\end{landscape}

\begin{table}[H]
\centering
\begin{adjustbox}{max totalsize={\textwidth}{0.95\textheight},center}
\begin{minipage}{\textwidth}
\centering
\caption{London problem (TripR-2020Large, 4 alternatives). Sensitivity of the collective ranking to the linguistic mapping (2-tuple operator, two-level hierarchy). The 2-tuple model places the terms at equally spaced indices; \texttt{compressed} moves the interior terms toward the center and \texttt{stretched} away from it, holding the endpoints and the ordering fixed. On LABEL-5 the five terms sit at 0, 1.6, 2, 2.4 and 4 under \texttt{compressed} and at 0, 0.5, 2, 3.5 and 4 under \texttt{stretched}; on LABEL-7 the seven sit at 0, 2, 2.6, 3, 3.4, 4 and 6 and at 0, 0.4, 1.2, 3, 4.8, 5.6 and 6. $\tau$ is the Kendall correlation of the resulting ranking against the uniform one.}
\label{tab:cdm_mapping}
\small
\adjustbox{max width=\linewidth}{%
\begin{tabular}{lllcc}
\toprule
\textbf{Model} & \textbf{Scheme} & \textbf{Placement} & \textbf{Winner} & \textbf{$\tau$ vs uniform} \\
\midrule
  Gemma 4 E4B & LABEL-7 & \texttt{uniform} & The Ivy & 1.000 \\
  Gemma 4 E4B & LABEL-7 & \texttt{compressed} & The Ivy & 1.000 \\
  Gemma 4 E4B & LABEL-7 & \texttt{stretched} & J. Sheekey & 0.667 \\
  Gemma 4 E4B & LABEL-5 & \texttt{uniform} & The Ivy & 1.000 \\
  Gemma 4 E4B & LABEL-5 & \texttt{compressed} & The Ivy & 1.000 \\
  Gemma 4 E4B & LABEL-5 & \texttt{stretched} & J. Sheekey & 0.333 \\
  Ministral 3 8B & LABEL-7 & \texttt{uniform} & J. Sheekey & 1.000 \\
  Ministral 3 8B & LABEL-7 & \texttt{compressed} & J. Sheekey & 1.000 \\
  Ministral 3 8B & LABEL-7 & \texttt{stretched} & J. Sheekey & 1.000 \\
  Ministral 3 8B & LABEL-5 & \texttt{uniform} & J. Sheekey & 1.000 \\
  Ministral 3 8B & LABEL-5 & \texttt{compressed} & J. Sheekey & 1.000 \\
  Ministral 3 8B & LABEL-5 & \texttt{stretched} & J. Sheekey & 0.667 \\
  Gemma 4 31B & LABEL-7 & \texttt{uniform} & J. Sheekey & 1.000 \\
  Gemma 4 31B & LABEL-7 & \texttt{compressed} & J. Sheekey & 1.000 \\
  Gemma 4 31B & LABEL-7 & \texttt{stretched} & J. Sheekey & 1.000 \\
  Gemma 4 31B & LABEL-5 & \texttt{uniform} & J. Sheekey & 1.000 \\
  Gemma 4 31B & LABEL-5 & \texttt{compressed} & J. Sheekey & 1.000 \\
  Gemma 4 31B & LABEL-5 & \texttt{stretched} & J. Sheekey & 0.667 \\
  Ornith 35B & LABEL-7 & \texttt{uniform} & J. Sheekey & 1.000 \\
  Ornith 35B & LABEL-7 & \texttt{compressed} & J. Sheekey & 1.000 \\
  Ornith 35B & LABEL-7 & \texttt{stretched} & J. Sheekey & 1.000 \\
  Ornith 35B & LABEL-5 & \texttt{uniform} & J. Sheekey & 1.000 \\
  Ornith 35B & LABEL-5 & \texttt{compressed} & J. Sheekey & 1.000 \\
  Ornith 35B & LABEL-5 & \texttt{stretched} & J. Sheekey & 1.000 \\
  Qwen 3.6 35B & LABEL-7 & \texttt{uniform} & J. Sheekey & 1.000 \\
  Qwen 3.6 35B & LABEL-7 & \texttt{compressed} & J. Sheekey & 1.000 \\
  Qwen 3.6 35B & LABEL-7 & \texttt{stretched} & J. Sheekey & 1.000 \\
  Qwen 3.6 35B & LABEL-5 & \texttt{uniform} & J. Sheekey & 1.000 \\
  Qwen 3.6 35B & LABEL-5 & \texttt{compressed} & J. Sheekey & 1.000 \\
  Qwen 3.6 35B & LABEL-5 & \texttt{stretched} & J. Sheekey & 1.000 \\
  GPT-5.2 & LABEL-7 & \texttt{uniform} & J. Sheekey & 1.000 \\
  GPT-5.2 & LABEL-7 & \texttt{compressed} & The Ivy & 0.667 \\
  GPT-5.2 & LABEL-7 & \texttt{stretched} & J. Sheekey & 1.000 \\
  GPT-5.2 & LABEL-5 & \texttt{uniform} & J. Sheekey & 1.000 \\
  GPT-5.2 & LABEL-5 & \texttt{compressed} & J. Sheekey & 1.000 \\
  GPT-5.2 & LABEL-5 & \texttt{stretched} & J. Sheekey & 1.000 \\
  Fusion (open-weight) & LABEL-7 & \texttt{uniform} & J. Sheekey & 1.000 \\
  Fusion (open-weight) & LABEL-7 & \texttt{compressed} & J. Sheekey & 1.000 \\
  Fusion (open-weight) & LABEL-7 & \texttt{stretched} & J. Sheekey & 1.000 \\
  Fusion (open-weight) & LABEL-5 & \texttt{uniform} & J. Sheekey & 1.000 \\
  Fusion (open-weight) & LABEL-5 & \texttt{compressed} & J. Sheekey & 1.000 \\
  Fusion (open-weight) & LABEL-5 & \texttt{stretched} & J. Sheekey & 1.000 \\
\bottomrule
\end{tabular}%
}
\end{minipage}
\end{adjustbox}
\end{table}

\begin{table}[H]
\centering
\begin{adjustbox}{max totalsize={\textwidth}{0.95\textheight},center}
\begin{minipage}{\textwidth}
\centering
\caption{London problem (TripR-2020Large, 4 alternatives). Sensitivity of the collective ranking to the treatment of rejected outputs under LABEL-5 (2-tuple operator, two-level hierarchy). \texttt{available} keeps a review and averages whatever runs parsed; \texttt{listwise} drops the review unless all ten did. Neither is neutral: the first compares reviews aggregated from different numbers of runs, the second removes reviews from the corpus. For the fused system \texttt{listwise} keeps a review only if every member returned all ten runs for it. Only systems with rejections can differ. Min gap is the smallest $\beta$ gap between adjacently ranked alternatives.}
\label{tab:cdm_missing}
\small
\adjustbox{max width=\linewidth}{%
\begin{tabular}{llrlr}
\toprule
\textbf{Model} & \textbf{Policy} & \textbf{Reviews} & \textbf{Winner} & \textbf{Min gap} \\
\midrule
  \multirow{2}{*}{Gemma 4 E4B} & \texttt{available} & 474 & The Ivy & 0.00266 \\
   & \texttt{listwise} & 474 & The Ivy & 0.00266 \\
\midrule
  \multirow{2}{*}{Ministral 3 8B} & \texttt{available} & 474 & J. Sheekey & 0.00085 \\
   & \texttt{listwise} & 413 & J. Sheekey & 0.04607 \\
\midrule
  \multirow{2}{*}{Gemma 4 31B} & \texttt{available} & 474 & J. Sheekey & 0.02288 \\
   & \texttt{listwise} & 474 & J. Sheekey & 0.02288 \\
\midrule
  \multirow{2}{*}{Ornith 35B} & \texttt{available} & 474 & J. Sheekey & 0.03870 \\
   & \texttt{listwise} & 474 & J. Sheekey & 0.03870 \\
\midrule
  \multirow{2}{*}{Qwen 3.6 35B} & \texttt{available} & 474 & J. Sheekey & 0.03555 \\
   & \texttt{listwise} & 472 & J. Sheekey & 0.03555 \\
\midrule
  \multirow{2}{*}{GPT-5.2} & \texttt{available} & 474 & J. Sheekey & 0.08039 \\
   & \texttt{listwise} & 474 & J. Sheekey & 0.08039 \\
\midrule
  \multirow{2}{*}{Fusion (open-weight)} & \texttt{available} & 474 & J. Sheekey & 0.01454 \\
   & \texttt{listwise} & 411 & J. Sheekey & 0.06805 \\
\bottomrule
\end{tabular}%
}
\\[2pt]\footnotesize Rejection is selective: for Ministral 3 8B the 61 affected reviews average $+0.50$ on the normalized scale against $+0.28$ for the rest, and the share of an alternative's reviews affected ranges from 8.4\% to 17.6\%.
\end{minipage}
\end{adjustbox}
\end{table}

\begin{table}[H]
\centering
\begin{adjustbox}{max totalsize={\textwidth}{0.95\textheight},center}
\begin{minipage}{\textwidth}
\centering
\caption{London problem (TripR-2020Large, 4 alternatives). The tie rule of the \texttt{majority} operator (LABEL-5, two-level hierarchy). The reported rule breaks a tie between modal terms toward the lower term; the alternative breaks it toward the upper one. Changed is the number of reviews whose review-level value differs between the rules (for the fusion, the mode is taken over its members' review-level values). Same order says whether the two rules give the same full ranking, and Min gap is the smallest adjacent gap of the collective ranking under each rule.}
\label{tab:cdm_tiebreak}
\small
\adjustbox{max width=\linewidth}{%
\begin{tabular}{lrllcrr}
\toprule
\textbf{System} & \textbf{Changed} & \textbf{Winner (lower)} & \textbf{Winner (upper)} & \textbf{Same order} & \textbf{Min gap (lower)} & \textbf{Min gap (upper)} \\
\midrule
  Gemma 4 E4B & 41 & The Ivy & J. Sheekey & no & 0.0079 & 0.0070 \\
  Ministral 3 8B & 39 & J. Sheekey & J. Sheekey & no & 0.0263 & 0.0087 \\
  Gemma 4 31B & 7 & J. Sheekey & J. Sheekey & yes & 0.0320 & 0.0126 \\
  Ornith 35B & 16 & J. Sheekey & J. Sheekey & yes & 0.0312 & 0.0519 \\
  Qwen 3.6 35B & 13 & J. Sheekey & J. Sheekey & yes & 0.0128 & 0.0216 \\
  GPT-5.2 & 1 & J. Sheekey & J. Sheekey & yes & 0.0764 & 0.0764 \\
  Fusion (open-weight) & 27 & J. Sheekey & J. Sheekey & yes & 0.0509 & 0.0426 \\
\bottomrule
\end{tabular}%
}
\end{minipage}
\end{adjustbox}
\end{table}

\section{Fixed-panel budgets and greedy decoding}

\begin{table}[H]
\centering
\begin{adjustbox}{max totalsize={\textwidth}{0.95\textheight},center}
\begin{minipage}{\textwidth}
\centering
\caption{London problem (TripR-2020Large, 4 alternatives). Sampling temperature: $T = 1.0$ against greedy decoding, under LABEL-5. Same prompt, vocabulary, parser and reviews; for Gemma 4 E4B the two collections also come from different campaigns (the main text). $\mathit{SD}_w$ is the standard deviation of a review's ten runs averaged over reviews; Table~\ref{tab:cdm_discriminance} pools the same spread as the root of the mean variance ($\mathit{SD}_{\mathrm{in}}$), which is always at least as large. Ident.\ is the share of reviews whose ten runs agree; Distinct counts the different review-level aggregates, Terms the linguistic terms the collective evaluations occupy, and Min gap the smallest adjacent gap. Each system is scored on the reviews that survived parsing at both temperatures; Drop is how many that cost. The fusion keeps the eight reviews that Ministral 3 8B loses at $T = 0$ with its other four members at both temperatures, so its $T = 1.0$ Distinct (175) differs from the 171 values of the reported pipeline. With one greedy run per member, the fused aggregate has 23 distinct review-level values. Greedy decoding removes the run-level spread, reduces resolution, and moves the margin in both directions.}
\label{tab:cdm_temperature}
\small
\adjustbox{max width=\linewidth}{%
\begin{tabular}{lrrrrrrrrrrr}
\toprule
 & \multicolumn{4}{c}{\textbf{Run level}} & \multicolumn{2}{c}{\textbf{Resolution}} & \multicolumn{4}{c}{\textbf{Decision}} & \\
\cmidrule(lr){2-5}\cmidrule(lr){6-7}\cmidrule(lr){8-11}
\textbf{Model} & $\mathit{SD}_w^{T1}$ & $\mathit{SD}_w^{T0}$ & Ident.$^{T1}$ & Ident.$^{T0}$ & Dist.$^{T1}$ & Dist.$^{T0}$ & Terms$^{T1}$ & Terms$^{T0}$ & Min gap$^{T1}$ & Min gap$^{T0}$ & Drop \\
\midrule
  Gemma 4 31B & 0.025 & 0.001 & 94\% & 100\% & 21 & 6 & 1 & 1 & 0.0229 & 0.0283 & 0 \\
  Gemma 4 E4B & 0.323 & 0.002 & 26\% & 99\% & 39 & 8 & 1 & 1 & 0.0027 & 0.0534 & 0 \\
  Ministral 3 8B & 0.351 & 0.003 & 23\% & 99\% & 64 & 9 & 1 & 2 & 0.0082 & 0.0120 & 8 \\
  Ornith 35B & 0.200 & 0.000 & 52\% & 100\% & 37 & 5 & 1 & 1 & 0.0387 & 0.0264 & 0 \\
  Qwen 3.6 35B & 0.185 & 0.000 & 55\% & 100\% & 41 & 5 & 1 & 1 & 0.0355 & 0.0427 & 0 \\
  Fusion (open-weight) & -- & -- & -- & -- & 175 & 31 & 1 & 1 & 0.0145 & 0.0152 & 0 \\
\bottomrule
\end{tabular}%
}
\end{minipage}
\end{adjustbox}
\end{table}

\begin{table}[H]
\centering
\begin{adjustbox}{max totalsize={\textwidth}{0.95\textheight},center}
\begin{minipage}{\textwidth}
\centering
\caption{London problem (TripR-2020Large, 4 alternatives). The fused system against each of its members and the hosted reference, on the quantities a decision rests on (LABEL-5, 2-tuple operator, two-level hierarchy). Lead and Min gap are the $\beta$ lead of the winner over the runner-up and the smallest adjacent gap. Top-1 and Exact under runs resample each review's runs with the crowd fixed; Top-1 and Cand.\ under crowd resample the participants with the runs fixed: Cand.\ is the size of the simultaneous candidate set, the alternatives no rival is simultaneously better than (Table~\ref{tab:cdm_candidates}). $r$ is the review-level Pearson correlation with the reviewers' star ratings. Cost is the median request latency times ten runs, in seconds per review (re-measured for Gemma 4 E4B; Table~\ref{tab:latency}); the fusion's is the sum over its members. The hosted reference has no comparable latency.}
\label{tab:cdm_fusion_vs_best}
\small
\adjustbox{max width=\linewidth}{%
\begin{tabular}{llrrrrrrrr}
\toprule
 & & & & \multicolumn{2}{c}{\textbf{Runs}} & \multicolumn{2}{c}{\textbf{Crowd}} & & \\
\cmidrule(lr){5-6}\cmidrule(lr){7-8}
\textbf{System} & \textbf{Winner} & \textbf{Lead} & \textbf{Min gap} & \textbf{Top-1} & \textbf{Exact} & \textbf{Top-1} & \textbf{Cand.} & \textbf{$r$} & \textbf{Cost (s)} \\
\midrule
  Gemma 4 E4B & The Ivy & 0.0171 & 0.0027 & 85.8\% & 48.4\% & 47.4\% & 4 & 0.764 & 3.00 \\
  Ministral 3 8B & J. Sheekey & 0.0236 & 0.0008 & 90.3\% & 46.8\% & 53.8\% & 4 & 0.752 & 1.34 \\
  Gemma 4 31B & J. Sheekey & 0.0773 & 0.0229 & 100.0\% & 100.0\% & 71.1\% & 4 & 0.798 & 3.17 \\
  Ornith 35B & J. Sheekey & 0.0649 & 0.0387 & 100.0\% & 99.9\% & 70.7\% & 4 & 0.816 & 1.61 \\
  Qwen 3.6 35B & J. Sheekey & 0.0355 & 0.0355 & 99.9\% & 99.9\% & 61.3\% & 4 & 0.814 & 1.86 \\
  GPT-5.2 & J. Sheekey & 0.1336 & 0.0804 & 100.0\% & 100.0\% & 81.2\% & 4 & 0.782 & -- \\
\midrule
  Fusion (open-weight) & J. Sheekey & 0.0369 & 0.0145 & 100.0\% & 99.0\% & 60.8\% & 4 & 0.814 & 10.98 \\
\bottomrule
\end{tabular}%
}
\end{minipage}
\end{adjustbox}
\end{table}

\begin{table}[H]
\centering
\begin{adjustbox}{max totalsize={\textwidth}{0.95\textheight},center}
\begin{minipage}{\textwidth}
\centering
\caption{London problem (TripR-2020Large, 4 alternatives). Inference budget at common cost (LABEL-5, 2-tuple operator, two-level hierarchy). A budget of $k$ runs per model is emulated by every $k$-subset of the ten collected runs (Subsets), the same run indices for every review; figures are the median over subsets with the 5th and 95th percentiles in brackets. Top-1 is the share of crowd replicas (participants redrawn as clusters, 20{,}000 replicas shared by every row) in which no alternative strictly beats the subset's own winner; Lead is that winner's $\beta$ lead over the runner-up; Same winner is the share of subsets whose winner is the one the system reports at $T = 1.0$ with ten runs. Calls is the number of model calls per review, $k$ for one model and $m k$ for a fusion of $m$ members, whose $k$-subset is applied by rounds (the same run positions for every member). Cost is estimated from per-request medians: median request latency times $k$, summed over members for the fusion. Greedy rows ($T = 0$) use one run, each of the ten greedy runs taken in turn (Subsets), and the sampled collection's latency.}
\label{tab:cdm_budget}
\small
\adjustbox{max width=\linewidth}{%
\begin{tabular}{lrrrrllr}
\toprule
\textbf{System} & \textbf{$k$} & \textbf{Subsets} & \textbf{Calls} & \textbf{Cost (s)} & \textbf{Top-1} & \textbf{Lead} & \textbf{Same winner} \\
\midrule
  Gemma 4 E4B & 1 & 10 & 1 & 0.30 & 49.5\% \tiny[40.8, 60.0] & 0.0217 \tiny[0.0099, 0.0756] & 70\% \\
  Gemma 4 E4B & 2 & 45 & 2 & 0.60 & 47.5\% \tiny[39.2, 57.2] & 0.0162 \tiny[0.0023, 0.0533] & 76\% \\
  Gemma 4 E4B & 3 & 120 & 3 & 0.90 & 47.1\% \tiny[40.8, 55.9] & 0.0162 \tiny[0.0008, 0.0463] & 82\% \\
  Gemma 4 E4B & 5 & 252 & 5 & 1.50 & 46.8\% \tiny[42.1, 53.1] & 0.0171 \tiny[0.0014, 0.0367] & 94\% \\
  Gemma 4 E4B & 10 & 1 & 10 & 3.00 & 47.4\% \tiny[47.4, 47.4] & 0.0171 \tiny[0.0171, 0.0171] & 100\% \\
  Gemma 4 E4B (T=0) & 1 & 10 & 1 & 0.30 & 68.4\% \tiny[65.9, 68.4] & 0.0685 \tiny[0.0592, 0.0685] & 0\% \\
\midrule
  Ministral 3 8B & 1 & 10 & 1 & 0.13 & 54.4\% \tiny[44.8, 82.7] & 0.0551 \tiny[0.0157, 0.1549] & 50\% \\
  Ministral 3 8B & 2 & 45 & 2 & 0.27 & 57.4\% \tiny[43.3, 82.5] & 0.0384 \tiny[0.0022, 0.1273] & 67\% \\
  Ministral 3 8B & 3 & 120 & 3 & 0.40 & 55.3\% \tiny[44.8, 74.9] & 0.0325 \tiny[0.0044, 0.0985] & 72\% \\
  Ministral 3 8B & 5 & 252 & 5 & 0.67 & 54.3\% \tiny[44.9, 69.5] & 0.0272 \tiny[0.0026, 0.0718] & 79\% \\
  Ministral 3 8B & 10 & 1 & 10 & 1.34 & 53.8\% \tiny[53.8, 53.8] & 0.0236 \tiny[0.0236, 0.0236] & 100\% \\
  Ministral 3 8B (T=0) & 1 & 10 & 1 & 0.13 & 44.4\% \tiny[44.4, 44.8] & 0.0130 \tiny[0.0080, 0.0130] & 10\% \\
\midrule
  Gemma 4 31B & 1 & 10 & 1 & 0.32 & 72.0\% \tiny[67.4, 74.5] & 0.0822 \tiny[0.0648, 0.0895] & 100\% \\
  Gemma 4 31B & 2 & 45 & 2 & 0.63 & 70.7\% \tiny[67.9, 74.3] & 0.0753 \tiny[0.0653, 0.0890] & 100\% \\
  Gemma 4 31B & 3 & 120 & 3 & 0.95 & 71.0\% \tiny[69.2, 73.7] & 0.0770 \tiny[0.0702, 0.0862] & 100\% \\
  Gemma 4 31B & 5 & 252 & 5 & 1.58 & 71.1\% \tiny[69.7, 72.5] & 0.0773 \tiny[0.0722, 0.0824] & 100\% \\
  Gemma 4 31B & 10 & 1 & 10 & 3.17 & 71.1\% \tiny[71.1, 71.1] & 0.0773 \tiny[0.0773, 0.0773] & 100\% \\
  Gemma 4 31B (T=0) & 1 & 10 & 1 & 0.32 & 72.4\% \tiny[72.4, 72.4] & 0.0838 \tiny[0.0838, 0.0838] & 100\% \\
\midrule
  Ornith 35B & 1 & 10 & 1 & 0.16 & 70.8\% \tiny[55.2, 79.2] & 0.0682 \tiny[0.0190, 0.1035] & 100\% \\
  Ornith 35B & 2 & 45 & 2 & 0.32 & 70.2\% \tiny[59.4, 77.9] & 0.0657 \tiny[0.0339, 0.0932] & 100\% \\
  Ornith 35B & 3 & 120 & 3 & 0.48 & 70.3\% \tiny[62.7, 76.5] & 0.0654 \tiny[0.0409, 0.0877] & 100\% \\
  Ornith 35B & 5 & 252 & 5 & 0.81 & 70.6\% \tiny[65.5, 75.1] & 0.0649 \tiny[0.0491, 0.0806] & 100\% \\
  Ornith 35B & 10 & 1 & 10 & 1.61 & 70.7\% \tiny[70.7, 70.7] & 0.0649 \tiny[0.0649, 0.0649] & 100\% \\
  Ornith 35B (T=0) & 1 & 10 & 1 & 0.16 & 78.7\% \tiny[78.7, 78.7] & 0.1005 \tiny[0.1005, 0.1005] & 100\% \\
\midrule
  Qwen 3.6 35B & 1 & 10 & 1 & 0.19 & 62.3\% \tiny[52.6, 76.2] & 0.0409 \tiny[0.0116, 0.0884] & 80\% \\
  Qwen 3.6 35B & 2 & 45 & 2 & 0.37 & 60.9\% \tiny[50.6, 73.2] & 0.0363 \tiny[0.0048, 0.0779] & 89\% \\
  Qwen 3.6 35B & 3 & 120 & 3 & 0.56 & 61.2\% \tiny[51.3, 71.3] & 0.0363 \tiny[0.0053, 0.0682] & 95\% \\
  Qwen 3.6 35B & 5 & 252 & 5 & 0.93 & 61.2\% \tiny[54.2, 68.0] & 0.0355 \tiny[0.0134, 0.0577] & 100\% \\
  Qwen 3.6 35B & 10 & 1 & 10 & 1.86 & 61.3\% \tiny[61.3, 61.3] & 0.0355 \tiny[0.0355, 0.0355] & 100\% \\
  Qwen 3.6 35B (T=0) & 1 & 10 & 1 & 0.19 & 71.1\% \tiny[71.1, 71.1] & 0.0738 \tiny[0.0738, 0.0738] & 100\% \\
\midrule
  Fusion (open-weight) & 1 & 10 & 5 & 1.10 & 60.5\% \tiny[57.2, 65.6] & 0.0357 \tiny[0.0265, 0.0544] & 100\% \\
  Fusion (open-weight) & 2 & 45 & 10 & 2.20 & 60.5\% \tiny[57.4, 65.1] & 0.0360 \tiny[0.0272, 0.0518] & 100\% \\
  Fusion (open-weight) & 3 & 120 & 15 & 3.29 & 60.9\% \tiny[58.2, 63.8] & 0.0373 \tiny[0.0289, 0.0478] & 100\% \\
  Fusion (open-weight) & 5 & 252 & 25 & 5.49 & 60.8\% \tiny[59.0, 62.6] & 0.0370 \tiny[0.0312, 0.0432] & 100\% \\
  Fusion (open-weight) & 10 & 1 & 50 & 10.98 & 60.8\% \tiny[60.8, 60.8] & 0.0369 \tiny[0.0369, 0.0369] & 100\% \\
  Fusion (open-weight) (T=0) & 1 & 10 & 5 & 1.10 & 68.2\% \tiny[68.2, 68.4] & 0.0614 \tiny[0.0614, 0.0618] & 100\% \\
\bottomrule
\end{tabular}%
}
\end{minipage}
\end{adjustbox}
\end{table}

\begin{table}[H]
\centering
\begin{adjustbox}{max totalsize={\textwidth}{0.95\textheight},center}
\begin{minipage}{\textwidth}
\centering
\caption{Madrid problem (TripAdvisor Restaurant Reviews v2.0.0, 12 alternatives). Inference budget at common cost (LABEL-5, 2-tuple operator, two-level hierarchy). A budget of $k$ runs per model is emulated by every $k$-subset of the ten collected runs (Subsets), the same run indices for every review; figures are the median over subsets with the 5th and 95th percentiles in brackets. Top-1 is the share of crowd replicas (participants redrawn as clusters, 20{,}000 replicas shared by every row) in which no alternative strictly beats the subset's own winner; Lead is that winner's $\beta$ lead over the runner-up; Same winner is the share of subsets whose winner is the one the system reports at $T = 1.0$ with ten runs. Calls is the number of model calls per review, $k$ for one model and $m k$ for a fusion of $m$ members, whose $k$-subset is applied by rounds (the same run positions for every member). Cost is the median request latency of Table~\ref{tab:uniovi_latency} times $k$, summed over members for the fusion; those medians come from a sequential re-measurement on the collection's runtime and hardware (Ollama 0.40.4), one request at a time over the first 60 reviews under each scheme, because the collection itself kept four requests in flight per model. Cost is comparable within this table but not with Table~\ref{tab:cdm_budget}. A subset that ties the ten-run winner for first place is credited with that winner, for Top-1 and for Same winner; this happens for Gemma 4 E4B in one subset at each of $k = 1, 2, 3$ and for Ministral 3 8B in one subset at $k = 2$ and five at $k = 3$. Greedy rows ($T = 0$) use one run, each of the ten greedy runs taken in turn (Subsets), at the same latency.}
\label{tab:uniovi_cdm_budget}
\small
\adjustbox{max width=\linewidth}{%
\begin{tabular}{lrrrrllr}
\toprule
\textbf{System} & \textbf{$k$} & \textbf{Subsets} & \textbf{Calls} & \textbf{Cost (s)} & \textbf{Top-1} & \textbf{Lead} & \textbf{Same winner} \\
\midrule
  Gemma 4 E4B & 1 & 10 & 1 & 0.03 & 54.2\% \tiny[46.0, 67.5] & 0.0400 \tiny[0.0090, 0.0710] & 90\% \\
  Gemma 4 E4B & 2 & 45 & 2 & 0.07 & 54.8\% \tiny[45.5, 65.5] & 0.0400 \tiny[0.0100, 0.0600] & 98\% \\
  Gemma 4 E4B & 3 & 120 & 3 & 0.10 & 55.3\% \tiny[46.9, 64.0] & 0.0333 \tiny[0.0130, 0.0533] & 100\% \\
  Gemma 4 E4B & 5 & 252 & 5 & 0.17 & 55.6\% \tiny[49.9, 61.4] & 0.0340 \tiny[0.0200, 0.0480] & 100\% \\
  Gemma 4 E4B & 10 & 1 & 10 & 0.34 & 55.9\% \tiny[55.9, 55.9] & 0.0340 \tiny[0.0340, 0.0340] & 100\% \\
  Gemma 4 E4B (T=0) & 1 & 10 & 1 & 0.03 & 69.2\% \tiny[65.5, 73.0] & 0.0700 \tiny[0.0600, 0.0800] & 100\% \\
\midrule
  Ministral 3 8B & 1 & 10 & 1 & 0.03 & 49.8\% \tiny[33.5, 72.4] & 0.0330 \tiny[0.0054, 0.0977] & 100\% \\
  Ministral 3 8B & 2 & 45 & 2 & 0.05 & 55.5\% \tiny[41.4, 66.2] & 0.0400 \tiny[0.0120, 0.0600] & 84\% \\
  Ministral 3 8B & 3 & 120 & 3 & 0.08 & 54.9\% \tiny[41.7, 64.3] & 0.0367 \tiny[0.0033, 0.0567] & 92\% \\
  Ministral 3 8B & 5 & 252 & 5 & 0.14 & 55.1\% \tiny[48.3, 63.5] & 0.0285 \tiny[0.0090, 0.0497] & 100\% \\
  Ministral 3 8B & 10 & 1 & 10 & 0.27 & 56.8\% \tiny[56.8, 56.8] & 0.0301 \tiny[0.0301, 0.0301] & 100\% \\
  Ministral 3 8B (T=0) & 1 & 10 & 1 & 0.03 & 66.5\% \tiny[66.3, 71.7] & 0.0800 \tiny[0.0800, 0.1000] & 100\% \\
\midrule
  Qwen 3.6 35B & 1 & 10 & 1 & 0.07 & 87.3\% \tiny[68.7, 94.6] & 0.1200 \tiny[0.0600, 0.1710] & 100\% \\
  Qwen 3.6 35B & 2 & 45 & 2 & 0.14 & 87.5\% \tiny[73.8, 94.6] & 0.1200 \tiny[0.0720, 0.1780] & 100\% \\
  Qwen 3.6 35B & 3 & 120 & 3 & 0.21 & 88.9\% \tiny[79.1, 94.2] & 0.1267 \tiny[0.0863, 0.1670] & 100\% \\
  Qwen 3.6 35B & 5 & 252 & 5 & 0.35 & 89.4\% \tiny[83.8, 93.2] & 0.1280 \tiny[0.0982, 0.1578] & 100\% \\
  Qwen 3.6 35B & 10 & 1 & 10 & 0.71 & 90.1\% \tiny[90.1, 90.1] & 0.1280 \tiny[0.1280, 0.1280] & 100\% \\
  Qwen 3.6 35B (T=0) & 1 & 10 & 1 & 0.07 & 79.4\% \tiny[79.4, 79.4] & 0.1000 \tiny[0.1000, 0.1000] & 100\% \\
\midrule
  Fusion (open-weight) & 1 & 10 & 3 & 0.13 & 57.7\% \tiny[52.6, 67.6] & 0.0600 \tiny[0.0348, 0.0703] & 100\% \\
  Fusion (open-weight) & 2 & 45 & 6 & 0.26 & 62.4\% \tiny[55.9, 68.2] & 0.0600 \tiny[0.0440, 0.0667] & 100\% \\
  Fusion (open-weight) & 3 & 120 & 9 & 0.40 & 62.9\% \tiny[58.0, 67.3] & 0.0578 \tiny[0.0478, 0.0667] & 100\% \\
  Fusion (open-weight) & 5 & 252 & 15 & 0.66 & 63.5\% \tiny[60.3, 66.3] & 0.0580 \tiny[0.0513, 0.0644] & 100\% \\
  Fusion (open-weight) & 10 & 1 & 30 & 1.32 & 63.8\% \tiny[63.8, 63.8] & 0.0580 \tiny[0.0580, 0.0580] & 100\% \\
  Fusion (open-weight) (T=0) & 1 & 10 & 3 & 0.13 & 58.6\% \tiny[57.6, 61.5] & 0.0400 \tiny[0.0400, 0.0467] & 100\% \\
\bottomrule
\end{tabular}%
}
\end{minipage}
\end{adjustbox}
\end{table}

\section{Additional graphical diagnostics}

The following figures complement the tabulated results. Individual percentile intervals describe uncertainty in collective evaluations and should be distinguished from simultaneous intervals for pairwise differences.

\begin{figure}[H]
\centering
\includegraphics[width=0.85\textwidth,height=0.65\textheight,keepaspectratio]{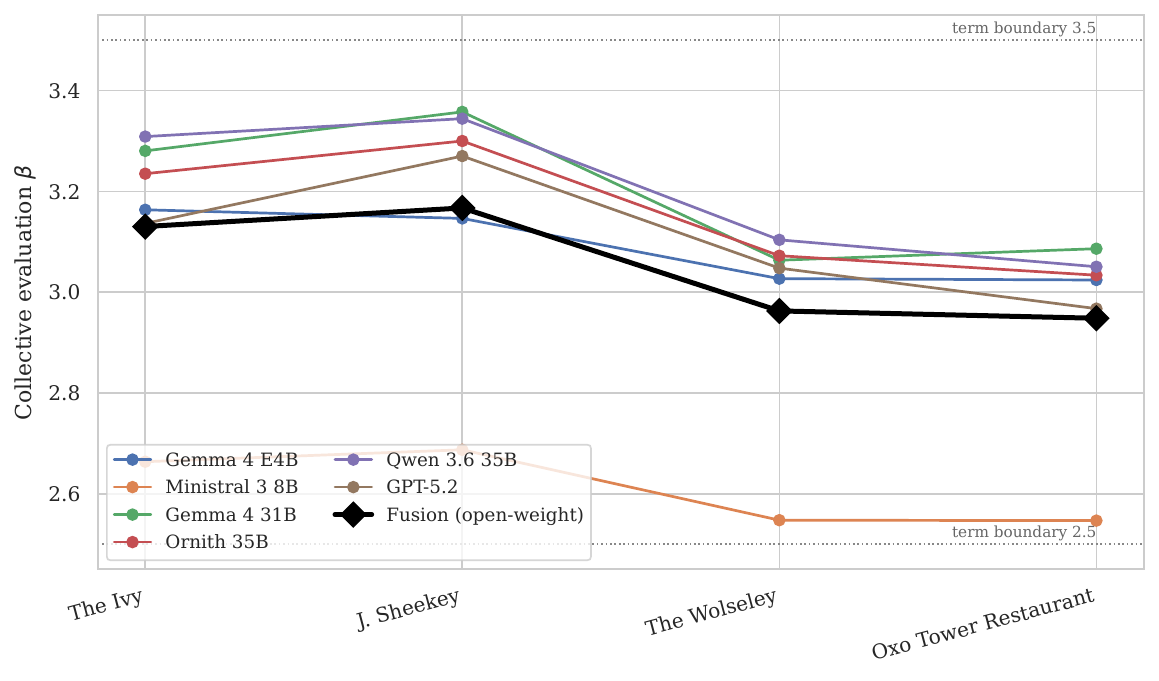}
\caption{London collective profiles on the continuous term-index domain. The linguistic term alone hides the small differences represented by the symbolic translations.}
\end{figure}

\begin{figure}[H]
\centering
\includegraphics[width=0.85\textwidth,height=0.65\textheight,keepaspectratio]{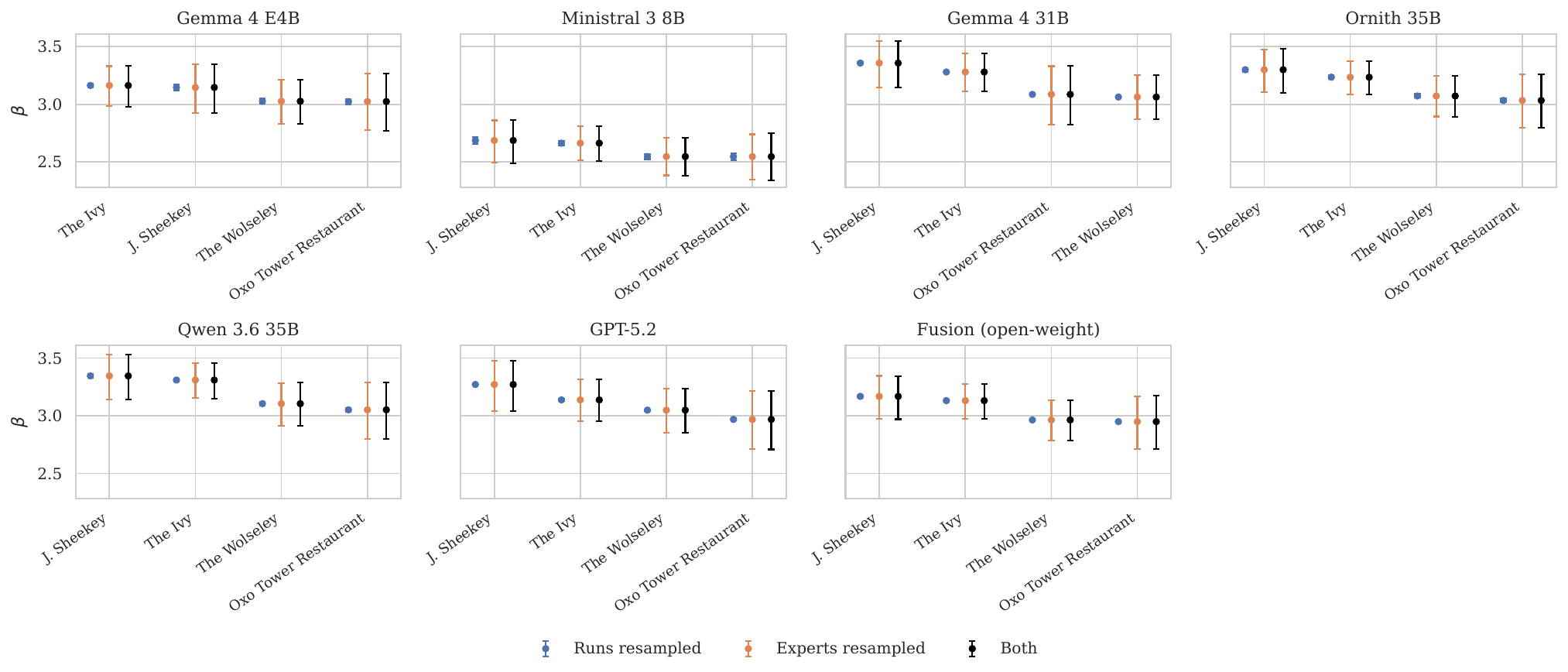}
\caption{London: individual 95\% percentile intervals for collective evaluations under run, participant and joint resampling. These are not simultaneous intervals for pairwise differences.}
\label{fig:london_sensitivity}
\end{figure}

\begin{figure}[H]
\centering
\includegraphics[width=0.85\textwidth,height=0.65\textheight,keepaspectratio]{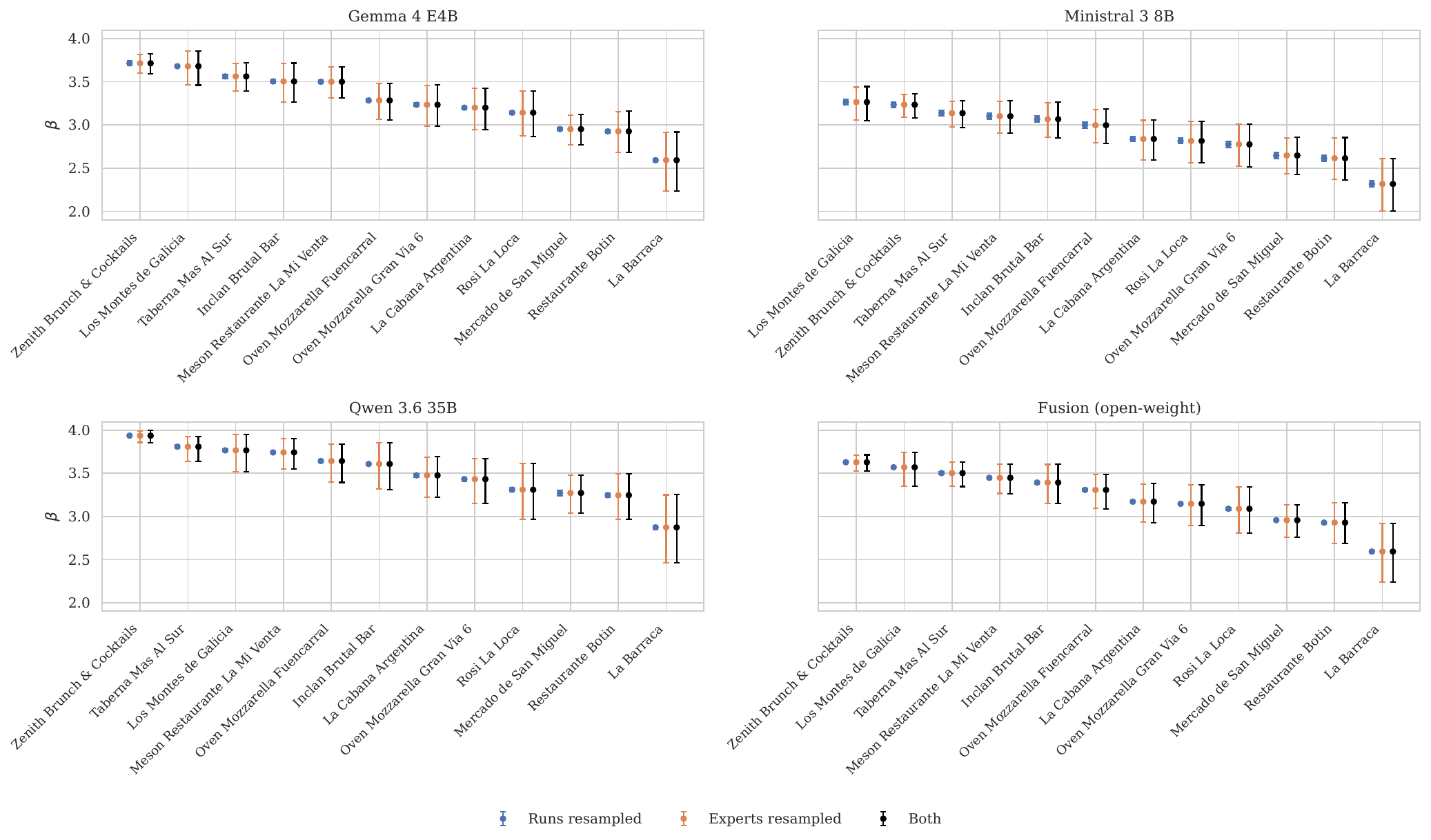}
\caption{Madrid collective evaluations with individual 95\% percentile intervals under run, participant and joint resampling. These intervals describe evaluations, not simultaneous pairwise separation.}
\end{figure}

\begin{figure}[H]
\centering
\includegraphics[width=0.85\textwidth,height=0.65\textheight,keepaspectratio]{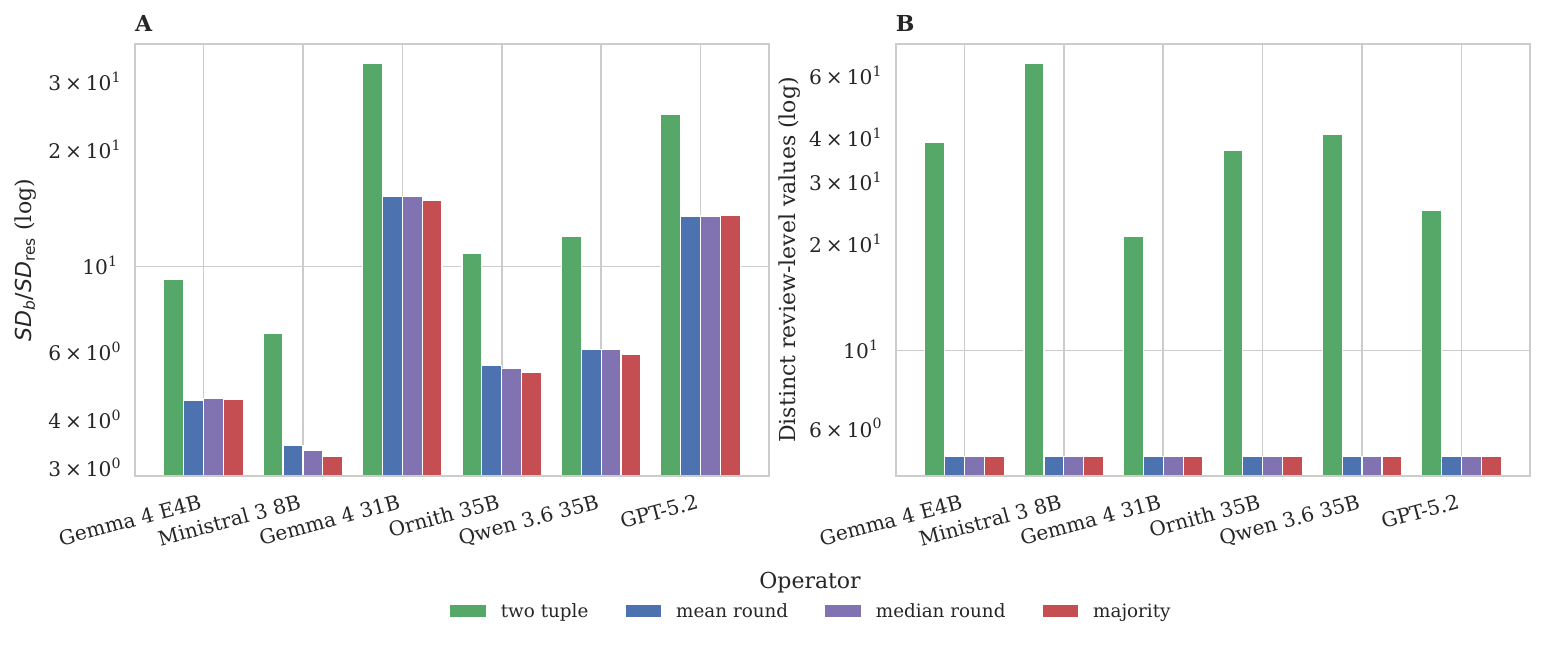}
\caption{London operator comparison: between-review spread relative to residual spread (A), and distinct review-level values (B). These descriptive quantities do not establish decision accuracy.}
\end{figure}

\begin{figure}[H]
\centering
\includegraphics[width=0.85\textwidth,height=0.65\textheight,keepaspectratio]{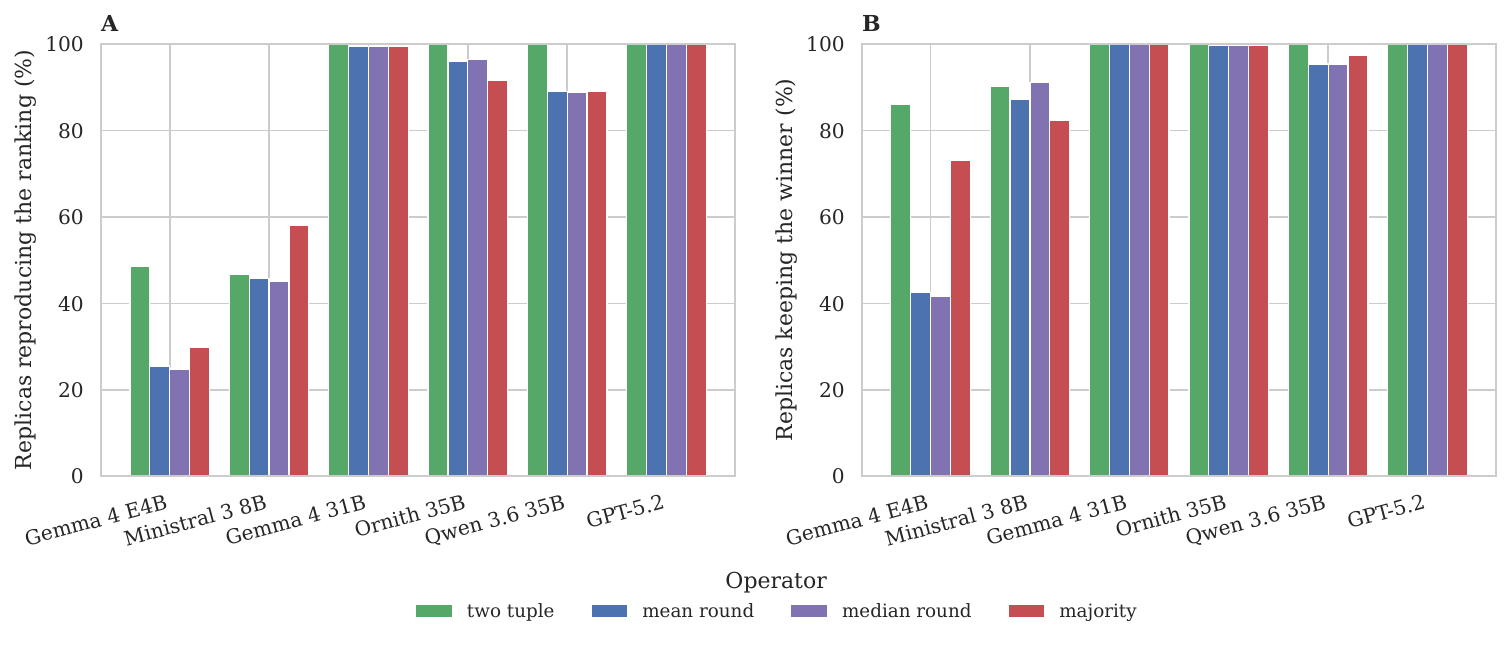}
\caption{London ranking stability under run resampling: complete-order reproduction (A) and winner retention (B). The operator-comparison replica stream differs from the joint-sensitivity stream.}
\end{figure}

\section{Madrid problem: validity, latency, operators and sensitivity}

These tables give, for the Madrid problem, the figures that the main text reports as ranges or single statements: output validity, the latencies behind the budget costs, residual noise and resolution by operator, the operator applied at every level, term placement, the treatment of rejected outputs, greedy decoding and agreement with the reviewers' stars. Every cell comes from the released result files.

\begin{table}[H]
\centering
\begin{adjustbox}{max totalsize={\textwidth}{0.95\textheight},center}
\begin{minipage}{\textwidth}
\centering
\caption{Madrid problem (TripAdvisor Restaurant Reviews v2.0.0, 12 alternatives). Output validity per system under the stated parsing policy: replies collected over the three schemes (600 reviews, ten runs each under SCORE, LABEL-7 and LABEL-5, at $T = 1.0$), replies rejected, and the rejected share over the three schemes (\%) and under LABEL-5 alone. Rejected replies leave the requested vocabulary or name more than one of its labels; they are excluded from every reported metric. All 81 rejections are Ministral 3 8B's: 78 LABEL-5 replies that name no label and 3 LABEL-7 replies that name more than one. Every review keeps at least six accepted LABEL-5 runs for every system. No hosted system was run on this problem.}
\label{tab:uniovi_cdm_validity}
\small
\begin{tabular}{lrrrr}
\toprule
\textbf{Model} & \textbf{Collected} & \textbf{Rejected} & \textbf{\%} & \textbf{LABEL-5 \%} \\
\midrule
  Gemma 4 E4B & 18{,}000 & 0 & 0.00\% & 0.00\% \\
  Ministral 3 8B & 18{,}000 & 81 & 0.45\% & 1.30\% \\
  Qwen 3.6 35B & 18{,}000 & 0 & 0.00\% & 0.00\% \\
\bottomrule
\end{tabular}
\end{minipage}
\end{adjustbox}
\end{table}

\begin{table}[H]
\centering
\begin{adjustbox}{max totalsize={\textwidth}{0.95\textheight},center}
\begin{minipage}{\textwidth}
\centering
\caption{Madrid problem (TripAdvisor Restaurant Reviews v2.0.0, 12 alternatives). Wall-clock latency per request on a single consumer GPU, measured around the backend call. The figures come from a sequential re-measurement, not from the collection: the collection kept four requests in flight per model, so its timings include queueing at the server and are not used. The re-measurement, made on the collection day, sent the first 60 reviews of the corpus once under each of the three schemes (SCORE, LABEL-7 and LABEL-5, $T = 1.0$), one request at a time, through the same experiment configuration, runtime (Ollama 0.40.4) and hardware as the collection (Table~\ref{tab:uniovi_provenance}), with the speculative-decoding draft of Gemma 4 E4B disabled as in the collection. $n$ is the number of requests. No request is discarded as warm-up: the first request of each model and scheme takes 1.07--3.08\,s, which lifts the mean above the 95th percentile but leaves the median unchanged at three decimals. The costs of Table~\ref{tab:uniovi_cdm_budget} are these medians times the number of calls; they are comparable with each other but not with the London latencies of Table~\ref{tab:latency}.}
\label{tab:uniovi_latency}
\adjustbox{max width=\linewidth}{%
\begin{tabular}{lrrrr}
\toprule
\textbf{Model} & \textbf{$n$} & \textbf{Mean (s)} & \textbf{Median (s)} & \textbf{p95 (s)} \\
\midrule
  Gemma 4 E4B & 180 & 0.071 & 0.034 & 0.043 \\
  Ministral 3 8B & 180 & 0.044 & 0.027 & 0.038 \\
  Qwen 3.6 35B & 180 & 0.120 & 0.071 & 0.084 \\
\bottomrule
\end{tabular}%
}
\end{minipage}
\end{adjustbox}
\end{table}

\begin{table}[H]
\centering
\begin{adjustbox}{max totalsize={\textwidth}{0.95\textheight},center}
\begin{minipage}{\textwidth}
\centering
\caption{Madrid problem (TripAdvisor Restaurant Reviews v2.0.0, 12 alternatives). Between-review spread against the noise each operator leaves, in $\beta$ units (LABEL-5, two-level hierarchy), as Table~\ref{tab:cdm_discriminance} gives it for London. $\mathit{SD}_b$ is the spread of the aggregated per-review evaluation across reviews; $\mathit{SD}_{\mathrm{in}}$ the spread of the raw runs within a review, identical across operators by construction; $\mathit{SD}_{\mathrm{res}}$ the spread of the operator's own output under resampling of the runs (5{,}000 replicas), the noise that survives aggregation. Both within-review spreads are the square root of the mean of the per-review variances. Ratio is $\mathit{SD}_b/\mathit{SD}_{\mathrm{res}}$ and Share $\mathit{SD}^2_b/(\mathit{SD}^2_b+\mathit{SD}^2_{\mathrm{res}})$, descriptive ratios, not an ICC. Distinct counts the different review-level aggregates and the different collective values of the twelve alternatives; fewer than 12 means that some alternatives tie.}
\label{tab:uniovi_cdm_discriminance}
\small
\begin{tabular}{llcccccrr}
\toprule
\textbf{Model} & \textbf{Operator} & \textbf{$\mathit{SD}_b$} & \textbf{$\mathit{SD}_{\mathrm{in}}$} & \textbf{$\mathit{SD}_{\mathrm{res}}$} & \textbf{Ratio} & \textbf{Share} & \textbf{Distinct (review)} & \textbf{Distinct (collective)} \\
\midrule
  \multirow{4}{*}{Gemma 4 E4B} & \texttt{two\_tuple} & 0.850 & 0.237 & 0.075 & 11.3 & 0.992 & 31 & 12 \\
   & \texttt{mean\_round} & 0.882 & 0.237 & 0.149 & 5.9 & 0.972 & 5 & 11 \\
   & \texttt{median\_round} & 0.882 & 0.237 & 0.149 & 5.9 & 0.972 & 5 & 11 \\
   & \texttt{majority} & 0.884 & 0.237 & 0.153 & 5.8 & 0.971 & 5 & 10 \\
\midrule
  \multirow{4}{*}{Ministral 3 8B} & \texttt{two\_tuple} & 0.809 & 0.379 & 0.121 & 6.7 & 0.978 & 58 & 12 \\
   & \texttt{mean\_round} & 0.856 & 0.379 & 0.244 & 3.5 & 0.925 & 5 & 11 \\
   & \texttt{median\_round} & 0.856 & 0.379 & 0.250 & 3.4 & 0.921 & 5 & 11 \\
   & \texttt{majority} & 0.835 & 0.379 & 0.249 & 3.4 & 0.918 & 5 & 11 \\
\midrule
  \multirow{4}{*}{Qwen 3.6 35B} & \texttt{two\_tuple} & 0.922 & 0.241 & 0.076 & 12.1 & 0.993 & 36 & 12 \\
   & \texttt{mean\_round} & 0.944 & 0.241 & 0.148 & 6.4 & 0.976 & 5 & 11 \\
   & \texttt{median\_round} & 0.944 & 0.241 & 0.149 & 6.4 & 0.976 & 5 & 11 \\
   & \texttt{majority} & 0.947 & 0.241 & 0.162 & 5.8 & 0.972 & 5 & 12 \\
\bottomrule
\end{tabular}
\end{minipage}
\end{adjustbox}
\end{table}

\begin{table}[H]
\centering
\begin{adjustbox}{max totalsize={\textwidth}{0.95\textheight},center}
\begin{minipage}{\textwidth}
\centering
\caption{Madrid problem (TripAdvisor Restaurant Reviews v2.0.0, 12 alternatives). Where the operator is allowed to act, under LABEL-5 and the two-level hierarchy, as Table~\ref{tab:cdm_scope} gives it for London. \texttt{review} is the reported pipeline: the operator aggregates the ten runs of a review, and the level above carries $\beta$ forward as a continuous value. \texttt{all} applies the same operator again where the reviews of an alternative are pooled, so that the collective evaluation itself lands on a term. Distinct is how many different collective values the twelve alternatives receive, and Min gap the smallest $\beta$ gap between adjacently ranked alternatives (-- where some alternatives tie). Unlike in London, rounding at the review stage already ties some alternatives in eight of the nine cases; at every level no rounded operator leaves more than two distinct values.}
\label{tab:uniovi_cdm_scope}
\small
\adjustbox{max width=\linewidth}{%
\begin{tabular}{llccr}
\toprule
\textbf{Model} & \textbf{Operator} & \textbf{Scope} & \textbf{Distinct} & \textbf{Min gap} \\
\midrule
  \multirow{8}{*}{Gemma 4 E4B} & \texttt{two\_tuple} & \texttt{review} & 12 & 0.0040 \\
   & \texttt{two\_tuple} & \texttt{all} & 12 & 0.0040 \\
   & \texttt{mean\_round} & \texttt{review} & 11 & -- \\
   & \texttt{mean\_round} & \texttt{all} & 2 & -- \\
   & \texttt{median\_round} & \texttt{review} & 11 & -- \\
   & \texttt{median\_round} & \texttt{all} & 2 & -- \\
   & \texttt{majority} & \texttt{review} & 10 & -- \\
   & \texttt{majority} & \texttt{all} & 2 & -- \\
\midrule
  \multirow{8}{*}{Ministral 3 8B} & \texttt{two\_tuple} & \texttt{review} & 12 & 0.0226 \\
   & \texttt{two\_tuple} & \texttt{all} & 12 & 0.0226 \\
   & \texttt{mean\_round} & \texttt{review} & 11 & -- \\
   & \texttt{mean\_round} & \texttt{all} & 2 & -- \\
   & \texttt{median\_round} & \texttt{review} & 11 & -- \\
   & \texttt{median\_round} & \texttt{all} & 1 & -- \\
   & \texttt{majority} & \texttt{review} & 11 & -- \\
   & \texttt{majority} & \texttt{all} & 1 & -- \\
\midrule
  \multirow{8}{*}{Qwen 3.6 35B} & \texttt{two\_tuple} & \texttt{review} & 12 & 0.0240 \\
   & \texttt{two\_tuple} & \texttt{all} & 12 & 0.0240 \\
   & \texttt{mean\_round} & \texttt{review} & 11 & -- \\
   & \texttt{mean\_round} & \texttt{all} & 2 & -- \\
   & \texttt{median\_round} & \texttt{review} & 11 & -- \\
   & \texttt{median\_round} & \texttt{all} & 1 & -- \\
   & \texttt{majority} & \texttt{review} & 12 & 0.0200 \\
   & \texttt{majority} & \texttt{all} & 1 & -- \\
\bottomrule
\end{tabular}%
}
\end{minipage}
\end{adjustbox}
\end{table}

\begin{table}[H]
\centering
\begin{adjustbox}{max totalsize={\textwidth}{0.95\textheight},center}
\begin{minipage}{\textwidth}
\centering
\caption{Madrid problem (TripAdvisor Restaurant Reviews v2.0.0, 12 alternatives). Sensitivity of the collective ranking to the linguistic mapping (2-tuple operator, two-level hierarchy), as Table~\ref{tab:cdm_mapping} gives it for London. The 2-tuple model places the terms at equally spaced indices; \texttt{compressed} moves the interior terms toward the center and \texttt{stretched} away from it, holding the endpoints and the ordering fixed. On LABEL-5 the five terms sit at 0, 1.6, 2, 2.4 and 4 under \texttt{compressed} and at 0, 0.5, 2, 3.5 and 4 under \texttt{stretched}; on LABEL-7 the seven sit at 0, 2, 2.6, 3, 3.4, 4 and 6 and at 0, 0.4, 1.2, 3, 4.8, 5.6 and 6. $\tau$ is the Kendall correlation of the resulting twelve-alternative ranking against the uniform one of the same scheme, and a change of winner is read against that uniform row: under uniform placement Ministral 3 8B ranks Zenith Brunch \& Cocktails--Valverde first on LABEL-7 and Los Montes de Galicia first on LABEL-5.}
\label{tab:uniovi_cdm_mapping}
\small
\adjustbox{max width=\linewidth}{%
\begin{tabular}{lllcc}
\toprule
\textbf{Model} & \textbf{Scheme} & \textbf{Placement} & \textbf{Winner} & \textbf{$\tau$ vs uniform} \\
\midrule
  Gemma 4 E4B & LABEL-7 & \texttt{uniform} & Zenith Brunch \& Cocktails--Valverde & 1.000 \\
  Gemma 4 E4B & LABEL-7 & \texttt{compressed} & Los Montes de Galicia & 0.909 \\
  Gemma 4 E4B & LABEL-7 & \texttt{stretched} & Zenith Brunch \& Cocktails--Valverde & 0.939 \\
  Gemma 4 E4B & LABEL-5 & \texttt{uniform} & Zenith Brunch \& Cocktails--Valverde & 1.000 \\
  Gemma 4 E4B & LABEL-5 & \texttt{compressed} & Los Montes de Galicia & 0.848 \\
  Gemma 4 E4B & LABEL-5 & \texttt{stretched} & Zenith Brunch \& Cocktails--Valverde & 0.939 \\
  Ministral 3 8B & LABEL-7 & \texttt{uniform} & Zenith Brunch \& Cocktails--Valverde & 1.000 \\
  Ministral 3 8B & LABEL-7 & \texttt{compressed} & Los Montes de Galicia & 0.909 \\
  Ministral 3 8B & LABEL-7 & \texttt{stretched} & Zenith Brunch \& Cocktails--Valverde & 1.000 \\
  Ministral 3 8B & LABEL-5 & \texttt{uniform} & Los Montes de Galicia & 1.000 \\
  Ministral 3 8B & LABEL-5 & \texttt{compressed} & Los Montes de Galicia & 0.879 \\
  Ministral 3 8B & LABEL-5 & \texttt{stretched} & Zenith Brunch \& Cocktails--Valverde & 0.970 \\
  Qwen 3.6 35B & LABEL-7 & \texttt{uniform} & Zenith Brunch \& Cocktails--Valverde & 1.000 \\
  Qwen 3.6 35B & LABEL-7 & \texttt{compressed} & Zenith Brunch \& Cocktails--Valverde & 0.970 \\
  Qwen 3.6 35B & LABEL-7 & \texttt{stretched} & Zenith Brunch \& Cocktails--Valverde & 0.909 \\
  Qwen 3.6 35B & LABEL-5 & \texttt{uniform} & Zenith Brunch \& Cocktails--Valverde & 1.000 \\
  Qwen 3.6 35B & LABEL-5 & \texttt{compressed} & Zenith Brunch \& Cocktails--Valverde & 0.939 \\
  Qwen 3.6 35B & LABEL-5 & \texttt{stretched} & Zenith Brunch \& Cocktails--Valverde & 0.939 \\
  Fusion (open-weight) & LABEL-7 & \texttt{uniform} & Zenith Brunch \& Cocktails--Valverde & 1.000 \\
  Fusion (open-weight) & LABEL-7 & \texttt{compressed} & Los Montes de Galicia & 0.909 \\
  Fusion (open-weight) & LABEL-7 & \texttt{stretched} & Zenith Brunch \& Cocktails--Valverde & 1.000 \\
  Fusion (open-weight) & LABEL-5 & \texttt{uniform} & Zenith Brunch \& Cocktails--Valverde & 1.000 \\
  Fusion (open-weight) & LABEL-5 & \texttt{compressed} & Zenith Brunch \& Cocktails--Valverde & 0.939 \\
  Fusion (open-weight) & LABEL-5 & \texttt{stretched} & Zenith Brunch \& Cocktails--Valverde & 1.000 \\
\bottomrule
\end{tabular}%
}
\end{minipage}
\end{adjustbox}
\end{table}

\begin{table}[H]
\centering
\begin{adjustbox}{max totalsize={\textwidth}{0.95\textheight},center}
\begin{minipage}{\textwidth}
\centering
\caption{Madrid problem (TripAdvisor Restaurant Reviews v2.0.0, 12 alternatives). Sensitivity of the collective ranking to the treatment of rejected outputs under LABEL-5 (2-tuple operator, two-level hierarchy), as Table~\ref{tab:cdm_missing} gives it for London. \texttt{available} keeps a review and averages whatever runs parsed; \texttt{listwise} drops the review unless all ten did. Neither is neutral: the first compares reviews aggregated from different numbers of runs, the second removes reviews from the corpus. For the fused system \texttt{listwise} keeps a review only if every member returned all ten runs for it. Only Ministral 3 8B has rejections, so only it and the fusion can differ. Min gap is the smallest $\beta$ gap between adjacently ranked alternatives.}
\label{tab:uniovi_cdm_missing}
\small
\adjustbox{max width=\linewidth}{%
\begin{tabular}{llrlr}
\toprule
\textbf{Model} & \textbf{Policy} & \textbf{Reviews} & \textbf{Winner} & \textbf{Min gap} \\
\midrule
  \multirow{2}{*}{Gemma 4 E4B} & \texttt{available} & 600 & Zenith Brunch \& Cocktails--Valverde & 0.00400 \\
   & \texttt{listwise} & 600 & Zenith Brunch \& Cocktails--Valverde & 0.00400 \\
\midrule
  \multirow{2}{*}{Ministral 3 8B} & \texttt{available} & 600 & Los Montes de Galicia & 0.02263 \\
   & \texttt{listwise} & 557 & Los Montes de Galicia & 0.02645 \\
\midrule
  \multirow{2}{*}{Qwen 3.6 35B} & \texttt{available} & 600 & Zenith Brunch \& Cocktails--Valverde & 0.02400 \\
   & \texttt{listwise} & 600 & Zenith Brunch \& Cocktails--Valverde & 0.02400 \\
\midrule
  \multirow{2}{*}{Fusion (open-weight)} & \texttt{available} & 600 & Zenith Brunch \& Cocktails--Valverde & 0.02420 \\
   & \texttt{listwise} & 557 & Los Montes de Galicia & 0.00331 \\
\bottomrule
\end{tabular}%
}
\\[2pt]\footnotesize Rejection is selective: for Ministral 3 8B the 43 affected reviews average $+0.51$ on the normalized scale against $+0.45$ for the rest, and the share of an alternative's reviews affected ranges from 2.0\% to 12.0\%.
\end{minipage}
\end{adjustbox}
\end{table}

\begin{table}[H]
\centering
\begin{adjustbox}{max totalsize={\textwidth}{0.95\textheight},center}
\begin{minipage}{\textwidth}
\centering
\caption{Madrid problem (TripAdvisor Restaurant Reviews v2.0.0, 12 alternatives). Sampling temperature: $T = 1.0$ against greedy decoding, under LABEL-5, ten runs per review at each temperature. Same prompt, vocabulary, parser and reviews; both collections were made on 14 October 2026 with the same runtime and weights (Table~\ref{tab:uniovi_provenance}). $\mathit{SD}_w$ is the standard deviation of a review's ten runs averaged over reviews; Table~\ref{tab:uniovi_cdm_discriminance} pools the same spread as the root of the mean variance ($\mathit{SD}_{\mathrm{in}}$), which is always at least as large. Ident.\ is the share of reviews whose ten runs agree; Distinct counts the different review-level aggregates, Terms the linguistic terms the collective evaluations occupy, and Min gap the smallest adjacent gap. Each system is scored on the reviews that survived parsing at both temperatures; Drop is how many that cost. Ministral 3 8B loses one review, whose ten greedy replies were all rejected; the fusion keeps that review with its other two members, so its $T = 1.0$ Min gap differs slightly from that of Table~\ref{tab:uniovi_cdm_missing}. Every system keeps its $T = 1.0$ winner at $T = 0$. Greedy decoding removes the run-level spread, reduces resolution, and moves the margin in both directions; the zero Min gap of Qwen 3.6 35B at $T = 0$ is a tie between two adjacently ranked alternatives.}
\label{tab:uniovi_cdm_temperature}
\small
\adjustbox{max width=\linewidth}{%
\begin{tabular}{lrrrrrrrrrrr}
\toprule
 & \multicolumn{4}{c}{\textbf{Run level}} & \multicolumn{2}{c}{\textbf{Resolution}} & \multicolumn{4}{c}{\textbf{Decision}} & \\
\cmidrule(lr){2-5}\cmidrule(lr){6-7}\cmidrule(lr){8-11}
\textbf{Model} & $\mathit{SD}_w^{T1}$ & $\mathit{SD}_w^{T0}$ & Ident.$^{T1}$ & Ident.$^{T0}$ & Dist.$^{T1}$ & Dist.$^{T0}$ & Terms$^{T1}$ & Terms$^{T0}$ & Min gap$^{T1}$ & Min gap$^{T0}$ & Drop \\
\midrule
  Gemma 4 E4B & 0.137 & 0.011 & 66\% & 97\% & 31 & 15 & 2 & 2 & 0.0040 & 0.0060 & 0 \\
  Ministral 3 8B & 0.318 & 0.008 & 24\% & 98\% & 57 & 15 & 2 & 2 & 0.0226 & 0.0040 & 1 \\
  Qwen 3.6 35B & 0.138 & 0.000 & 66\% & 100\% & 36 & 5 & 2 & 2 & 0.0240 & 0.0000 & 0 \\
  Fusion (open-weight) & -- & -- & -- & -- & 127 & 32 & 2 & 2 & 0.0236 & 0.0220 & 0 \\
\bottomrule
\end{tabular}%
}
\end{minipage}
\end{adjustbox}
\end{table}

\begin{table}[H]
\centering
\begin{adjustbox}{max totalsize={\textwidth}{0.95\textheight},center}
\begin{minipage}{\textwidth}
\centering
\caption{Madrid problem (TripAdvisor Restaurant Reviews v2.0.0, 12 alternatives). Agreement with the 1--5 star rating the author of each review attached to it, over all 600 reviews under LABEL-5, as Table~\ref{tab:cdm_stars} gives it for London. $r$ and $\rho$ are the Pearson and Spearman correlations over reviews; a rating of $h$ stars is $\beta^{\star} = h-1$, so Diff.\ (the mean signed difference from the stars, negative when the pipeline scores a review below its author) and MAD are in $\beta$ units. Same term is the share of reviews whose aggregate rounds onto the author's own term. $\tau$ is Kendall's $\tau_b$ between each collective ranking and the star ranking, which ties two pairs of restaurants exactly and keeps both ties: Zenith Brunch \& Cocktails--Valverde (3.94 mean stars) $>$ Los Montes de Galicia (3.82) $>$ Taberna Mas Al Sur $=$ Meson Restaurante La Mi Venta (3.76) $>$ Inclan Brutal Bar (3.66) $>$ Oven Mozzarella Fuencarral $=$ La Cabana Argentina (3.54) $>$ Oven Mozzarella Gran Via 6 (3.52) $>$ Rosi La Loca (3.32) $>$ Restaurante Botin (3.30) $>$ Mercado de San Miguel (3.24) $>$ La Barraca (2.78). The stars rate a visit, the answer rates the text: related constructs, not one quantity measured twice.}
\label{tab:uniovi_cdm_stars}
\small
\begin{tabular}{lccrrcr}
\toprule
\textbf{Model} & \textbf{$r$} & \textbf{$\rho$} & \textbf{Diff.} & \textbf{MAD} & \textbf{Same term} & \textbf{$\tau_b$} \\
\midrule
  Gemma 4 E4B & 0.815 & 0.644 & $-$0.241 & 0.361 & 67.8\% & +0.89 \\
  Ministral 3 8B & 0.780 & 0.609 & $-$0.615 & 0.682 & 36.5\% & +0.89 \\
  Qwen 3.6 35B & 0.871 & 0.705 & $-$0.006 & 0.242 & 78.5\% & +0.89 \\
  Fusion (open-weight) & 0.853 & 0.660 & $-$0.287 & 0.422 & 68.0\% & +0.95 \\
\bottomrule
\end{tabular}
\end{minipage}
\end{adjustbox}
\end{table}

\section{Paired comparisons and variance of the winner's lead}

These tables give the uncertainty of the comparisons between systems that the main text makes, and the ratio of the inference to the participant variance of each system's lead. All use the 20{,}000 participant replicas of seed 42 shared by every system.

\begin{table}[H]
\centering
\begin{adjustbox}{max totalsize={\textwidth}{0.95\textheight},center}
\begin{minipage}{\textwidth}
\centering
\caption{Paired comparisons between systems (LABEL-5, 2-tuple operator, two-level hierarchy). Lead A$-$B is the difference between the two systems' leads of the winner over the runner-up of the full data, recomputed on the same redrawn crowds, with its 2.5th and 97.5th percentiles; when the two systems do not share that pair, the gap of one named pair is evaluated in both (notes a--c). Crowd Top-1 is as in the main text. Its difference A$-$B is the mean of paired retention indicators, and its standard error (SE, percentage points) reflects the sampling variability of the crowd, not Monte Carlo error: it comes from a nested bootstrap, in which the first 2{,}000 of the 20{,}000 crowds are outer replicas and each receives 1{,}000 inner redraws of its participants, with every configuration's Top-1 recomputed against its own winner in the outer crowd; the interval is the normal approximation, the difference $\pm 1.96$ SE. The Monte Carlo standard errors of the same differences are 0.16--0.61 points. Budget rows compare configurations at the stated number of calls per review, with Top-1 the median over run subsets. (a) The two systems rank the leading pair in opposite order; the gap J.~Sheekey$-$The Ivy is compared. (b) Different runner-up; the gap between Zenith Brunch \& Cocktails--Valverde and Los Montes de Galicia is compared (with the gap between Zenith Brunch \& Cocktails--Valverde and Taberna Mas Al Sur instead, Qwen 3.6 35B differs from Gemma 4 E4B by $-0.024$ $[-0.153, +0.108]$ and from the fusion by $+0.002$ $[-0.073, +0.078]$). (c) Different winners; the gap between Zenith Brunch \& Cocktails--Valverde and Los Montes de Galicia is compared (with the gap to Taberna Mas Al Sur, Qwen 3.6 35B differs from Ministral 3 8B by $+0.031$ $[-0.098, +0.160]$). (d) In Madrid, review-level \texttt{majority} already ties alternatives for two of the three members (Table~\ref{tab:uniovi_cdm_scope}), and the number of distinct fused values under \texttt{majority} is not reported; since ties count as retention, this row is not an operator comparison under the rule of the main text.}
\label{tab:paired_comparisons}
\footnotesize
\setlength{\tabcolsep}{4pt}
\adjustbox{max width=\linewidth}{%
\begin{tabular}{llcccc}
\toprule
\textbf{Problem} & \textbf{A against B} & \textbf{Lead A$-$B [95\%]} & \textbf{Crowd Top-1 A, B} & \textbf{Top-1 A$-$B [95\%]} & \textbf{SE} \\
\midrule
London & GPT-5.2 / Gemma 4 E4B & $+0.151$ $[+0.043, +0.253]^{a}$ & 81.2\%, 47.4\% & $+33.8$ $[-11.3, +78.8]$ & 23.0 \\
 & GPT-5.2 / Ministral 3 8B & $+0.110$ $[-0.015, +0.234]$ & 81.2\%, 53.8\% & $+27.5$ $[-8.6, +63.5]$ & 18.4 \\
 & GPT-5.2 / Gemma 4 31B & $+0.056$ $[-0.032, +0.143]$ & 81.2\%, 71.1\% & $+10.1$ $[-14.0, +34.2]$ & 12.3 \\
 & GPT-5.2 / Ornith 35B & $+0.069$ $[-0.032, +0.163]$ & 81.2\%, 70.7\% & $+10.6$ $[-14.2, +35.3]$ & 12.6 \\
 & GPT-5.2 / Qwen 3.6 35B & $+0.098$ $[+0.008, +0.184]$ & 81.2\%, 61.3\% & $+19.9$ $[-13.0, +52.8]$ & 16.8 \\
 & GPT-5.2 / Fusion & $+0.097$ $[+0.017, +0.173]$ & 81.2\%, 60.8\% & $+20.5$ $[-10.4, +51.3]$ & 15.8 \\
 & Gemma 4 31B / Fusion & $+0.041$ $[-0.020, +0.099]$ & 71.1\%, 60.8\% & $+10.4$ $[-7.7, +28.5]$ & 9.2 \\
 & Ornith 35B / Fusion & $+0.028$ $[-0.019, +0.079]$ & 70.7\%, 60.8\% & $+9.9$ $[-8.3, +28.1]$ & 9.3 \\
 & Qwen 3.6 35B / Fusion & $-0.001$ $[-0.038, +0.036]$ & 61.3\%, 60.8\% & $+0.6$ $[-11.7, +12.8]$ & 6.3 \\
 & Fusion / Gemma 4 E4B & $+0.054$ $[-0.022, +0.131]^{a}$ & 60.8\%, 47.4\% & $+13.3$ $[-16.4, +43.0]$ & 15.1 \\
 & Fusion / Ministral 3 8B & $+0.013$ $[-0.078, +0.104]$ & 60.8\%, 53.8\% & $+7.0$ $[-17.5, +31.5]$ & 12.5 \\
 & Fusion, \texttt{majority} / continuous & $+0.049$ $[-0.016, +0.114]$ & 74.2\%, 60.8\% & $+13.4$ $[-8.1, +34.9]$ & 11.0 \\
 & Gemma 4 31B $k{=}5$ / Fusion $k{=}1$ (5 calls) & -- & 71.1\%, 60.5\% & $+10.6$ $[-7.2, +28.4]$ & 9.1 \\
 & Gemma 4 31B $k{=}10$ / Fusion $k{=}2$ (10 calls) & -- & 71.1\%, 60.5\% & $+10.6$ $[-7.1, +28.3]$ & 9.0 \\
 & Fusion $k{=}10$ / Fusion $k{=}1$ (50 and 5 calls) & -- & 60.8\%, 60.5\% & $+0.3$ $[-3.5, +4.0]$ & 1.9 \\
\midrule
Madrid & Qwen 3.6 35B / Gemma 4 E4B & $+0.136$ $[+0.022, +0.250]^{b}$ & 90.1\%, 55.9\% & $+34.2$ $[-10.8, +79.1]$ & 22.9 \\
 & Qwen 3.6 35B / Ministral 3 8B & $+0.200$ $[+0.075, +0.328]^{c}$ & 90.1\%, 56.8\% & $+33.3$ $[-20.5, +87.2]$ & 27.5 \\
 & Qwen 3.6 35B / Fusion & $+0.112$ $[+0.044, +0.180]^{b}$ & 90.1\%, 63.8\% & $+26.3$ $[-13.3, +65.8]$ & 20.2 \\
 & Fusion / Gemma 4 E4B & $+0.024$ $[-0.045, +0.092]$ & 63.8\%, 55.9\% & $+7.9$ $[-15.2, +31.0]$ & 11.8 \\
 & Fusion / Ministral 3 8B & $+0.088$ $[+0.015, +0.165]^{c}$ & 63.8\%, 56.8\% & $+7.0$ $[-31.6, +45.7]$ & 19.7 \\
 & Fusion, \texttt{majority} / continuous$^{d}$ & $+0.042$ $[-0.067, +0.149]$ & 73.7\%, 63.8\% & $+9.9$ $[-20.6, +40.3]$ & 15.5 \\
 & Qwen 3.6 35B $k{=}1$ / Fusion $k{=}10$ (1 and 30 calls) & -- & 87.3\%, 63.8\% & $+23.5$ $[-13.4, +60.3]$ & 18.8 \\
 & Qwen 3.6 35B $k{=}3$ / Fusion $k{=}1$ (3 calls) & -- & 88.9\%, 57.7\% & $+31.3$ $[-4.7, +67.2]$ & 18.3 \\
 & Fusion $k{=}10$ / Fusion $k{=}1$ (30 and 3 calls) & -- & 63.8\%, 57.7\% & $+6.1$ $[-5.8, +18.0]$ & 6.1 \\
 & Gemma 4 E4B, greedy / sampled & -- & 69.2\%, 55.9\% & $+13.3$ $[-12.4, +38.9]$ & 13.1 \\
 & Ministral 3 8B, greedy / sampled & -- & 66.5\%, 56.8\% & $+9.8$ $[-19.9, +39.4]$ & 15.1 \\
 & Qwen 3.6 35B, greedy / sampled & -- & 79.4\%, 90.1\% & $-10.7$ $[-35.3, +13.9]$ & 12.6 \\
 & Fusion, greedy / sampled & -- & 58.6\%, 63.8\% & $-5.3$ $[-23.5, +13.0]$ & 9.3 \\
\bottomrule
\end{tabular}%
}
\end{minipage}
\end{adjustbox}
\end{table}

\begin{table}[H]
\centering
\begin{adjustbox}{max totalsize={\textwidth}{0.95\textheight},center}
\begin{minipage}{\textwidth}
\centering
\caption{Variance of the winner's lead under participant and inference resampling (LABEL-5, 2-tuple operator, two-level hierarchy). The lead is taken between the full-data winner and runner-up of each system. $\mathrm{se}$ is its standard deviation over the 20{,}000 participant replicas with the model outputs fixed ($\mathrm{var}_{\mathrm{part}}=\mathrm{se}^2$). $\mathrm{var}_{\mathrm{inf}}(k)$ is its variance with the participants fixed when every review receives $k$ runs drawn with replacement from its collected runs (rejected runs discarded; fusion members drawn independently), computed exactly, so that the columns scale exactly as $1/k$, and checked by Monte Carlo; the columns give $\mathrm{var}_{\mathrm{inf}}(k)/\mathrm{var}_{\mathrm{part}}$ in percent, and the Monte Carlo estimates lie within 0.98--1.03 times the exact values except for Ministral 3 8B (up to 1.06 in London and 1.13 in Madrid), whose rejected runs leave some reviews without an accepted response at small $k$. $k^{*}$ is the smallest $k$ with $\mathrm{var}_{\mathrm{inf}}(k)\le\theta\,\mathrm{var}_{\mathrm{part}}$, that is $k\ge 10\,\mathrm{var}_{\mathrm{inf}}(10)/(\theta\,\mathrm{var}_{\mathrm{part}})$, for $\theta = 10\%$ and $5\%$, computed from the exact values; for the fusions it counts runs per member. The last three columns compare the observed crowd Top-1 with the normal approximation $F_{\mathcal{N}}(\mathrm{lead}/\mathrm{se})$, where $F_{\mathcal{N}}$ is the standard normal distribution function, which ignores rivals other than the runner-up, and with a multivariate normal approximation over all rivals of the winner.}
\label{tab:variance_lead}
\footnotesize
\setlength{\tabcolsep}{4pt}
\begin{tabular}{llrrrrrrrrrr}
\toprule
\textbf{Problem} & \textbf{System} & \textbf{Lead} & \textbf{se} & \textbf{$k{=}1$} & \textbf{$k{=}3$} & \textbf{$k{=}10$} & \textbf{$k^{*}_{10\%}$} & \textbf{$k^{*}_{5\%}$} & \textbf{$F_{\mathcal{N}}(\mathrm{lead}/\mathrm{se})$} & \textbf{Multivariate} & \textbf{Top-1} \\
\midrule
London & Gemma 4 E4B & 0.0171 & 0.127 & 15.8 & 5.3 & 1.58 & 2 & 4 & 55.4\% & 47.0\% & 47.4\% \\
 & Ministral 3 8B & 0.0236 & 0.103 & 30.8 & 10.3 & 3.08 & 4 & 7 & 59.0\% & 54.2\% & 53.8\% \\
 & Gemma 4 31B & 0.0773 & 0.126 & 1.0 & 0.3 & 0.10 & 1 & 1 & 73.0\% & 71.4\% & 71.1\% \\
 & Ornith 35B & 0.0649 & 0.112 & 10.9 & 3.6 & 1.09 & 2 & 3 & 71.8\% & 70.5\% & 70.7\% \\
 & Qwen 3.6 35B & 0.0355 & 0.117 & 9.4 & 3.1 & 0.94 & 1 & 2 & 61.9\% & 61.3\% & 61.3\% \\
 & GPT-5.2 & 0.1336 & 0.130 & 1.9 & 0.6 & 0.19 & 1 & 1 & 84.9\% & 81.1\% & 81.2\% \\
 & Fusion & 0.0369 & 0.113 & 2.7 & 0.9 & 0.27 & 1 & 1 & 62.8\% & 60.8\% & 60.8\% \\
\midrule
Madrid & Gemma 4 E4B & 0.0340 & 0.115 & 18.9 & 6.3 & 1.89 & 2 & 4 & 61.6\% & 56.6\% & 55.9\% \\
 & Ministral 3 8B & 0.0301 & 0.119 & 43.1 & 14.4 & 4.31 & 5 & 9 & 60.0\% & 54.5\% & 56.8\% \\
 & Qwen 3.6 35B & 0.1280 & 0.082 & 20.9 & 7.0 & 2.09 & 3 & 5 & 94.0\% & 84.5\% & 90.1\% \\
 & Fusion & 0.0580 & 0.111 & 8.4 & 2.8 & 0.84 & 1 & 2 & 69.9\% & 63.2\% & 63.8\% \\
\bottomrule
\end{tabular}
\end{minipage}
\end{adjustbox}
\end{table}

\begin{table}[H]
\centering
\begin{adjustbox}{max totalsize={\textwidth}{0.95\textheight},center}
\begin{minipage}{\textwidth}
\centering
\caption{Paired comparisons involving the reviewers' stars, on the same 20{,}000 redrawn crowds: London problem (TripR-2020Large, 4 alternatives), except for the Madrid comparison noted in (A). (A) Difference in review-level Pearson correlation with the stars, each correlation weighted by the multiplicities of the redrawn crowd, with the 2.5th and 97.5th percentiles of the paired difference; in Madrid, Qwen 3.6 35B exceeds the fusion by $+0.017$ $[+0.007, +0.028]$. (B) The stars treated as a system: each pairwise gap of the star evaluation minus the same gap of the fusion, in $\beta$ units, pairs in the order of the star ranking.}
\label{tab:stars_paired}
\small
\begin{tabular}{lccc}
\toprule
\multicolumn{4}{l}{\textbf{(A) Correlation with the stars}} \\
\textbf{A$-$B} & \textbf{$r_A$} & \textbf{$r_B$} & \textbf{Difference [95\%]} \\
\midrule
Gemma 4 31B $-$ GPT-5.2 & 0.798 & 0.782 & $+0.015$ $[+0.0003, +0.031]$ \\
Ornith 35B $-$ GPT-5.2 & 0.816 & 0.782 & $+0.034$ $[+0.015, +0.054]$ \\
Qwen 3.6 35B $-$ GPT-5.2 & 0.814 & 0.782 & $+0.032$ $[+0.014, +0.051]$ \\
Fusion $-$ GPT-5.2 & 0.814 & 0.782 & $+0.032$ $[+0.017, +0.047]$ \\
Fusion $-$ Ornith 35B & 0.814 & 0.816 & $-0.002$ $[-0.013, +0.009]$ \\
\midrule
\multicolumn{4}{l}{\textbf{(B) Stars minus fusion, pairwise gaps}} \\
\textbf{Pair} & \textbf{Stars} & \textbf{Fusion} & \textbf{Difference [95\%]} \\
\midrule
The Ivy $-$ J.~Sheekey & $+0.010$ & $-0.037$ & $+0.047$ $[-0.087, +0.183]$ \\
The Ivy $-$ The Wolseley & $+0.235$ & $+0.168$ & $+0.068$ $[-0.052, +0.189]$ \\
The Ivy $-$ Oxo Tower & $+0.360$ & $+0.182$ & $+0.178$ $[+0.030, +0.330]$ \\
J.~Sheekey $-$ The Wolseley & $+0.225$ & $+0.204$ & $+0.020$ $[-0.115, +0.154]$ \\
J.~Sheekey $-$ Oxo Tower & $+0.350$ & $+0.219$ & $+0.131$ $[-0.040, +0.298]$ \\
The Wolseley $-$ Oxo Tower & $+0.125$ & $+0.015$ & $+0.111$ $[-0.041, +0.259]$ \\
\bottomrule
\end{tabular}
\end{minipage}
\end{adjustbox}
\end{table}

\end{document}